\PassOptionsToPackage{table}{xcolor}
\makeatletter
\def\input@path{{EITNLP/}{./}}
\makeatother
\documentclass{eitnlp}
\graphicspath{{EITNLP/}{./}}
\usepackage{tabularx,multirow,makecell,enumitem,array,wrapfig}
\usepackage{algorithm,algpseudocode}

\usepackage{amsmath,amsfonts,bm}

\def\eqref#1{equation~\ref{#1}}

\def\1{\bm{1}}

\DeclareMathAlphabet{\mathsfit}{\encodingdefault}{\sfdefault}{m}{sl}
\SetMathAlphabet{\mathsfit}{bold}{\encodingdefault}{\sfdefault}{bx}{n}

\newcommand{\VtwoV}{\textsc{V2V}}
\newcommand{\TtwoV}{\textsc{T2V}}
\newcommand{\TtwoT}{\textsc{T2T}}

\newcommand{\cmark}{\textcolor{green!45!black}{\checkmark}}
\newcommand{\graycell}[1]{\cellcolor{gray!20}#1}
\definecolor{lightblue}{RGB}{222,237,247}

\title{MWOP: Modality-aware Width-wise Operation Pruning for Efficient MLLMs}
\runningtitle{MWOP}
\author{\textbf{Xudong Wang}\textsuperscript{1,*}\qquad
\textbf{Hao Wu}\textsuperscript{1,2,*}\qquad
\textbf{Haozhe Hu}\textsuperscript{1,2}\qquad
\textbf{Peiran Yin}\textsuperscript{1,2}\\
\textbf{Xinghao Chen}\textsuperscript{1,3}\qquad
\textbf{Yunpu Ma}\textsuperscript{4}\qquad
\textbf{Wei Zhang}\textsuperscript{1}\qquad
\textbf{Xiaoyu Shen}\textsuperscript{1,\textdagger}}
\affiliation{%
\textsuperscript{1}EIT-NLP Lab, Eastern Institute of Technology, Ningbo\\
\textsuperscript{2}Shanghai Jiao Tong University\qquad
\textsuperscript{3}The Hong Kong Polytechnic University\\
\textsuperscript{4}Munich Center for Machine Learning, LMU Munich}
\authornotes{\textsuperscript{*}Equal contribution.\qquad
\textsuperscript{\textdagger}Corresponding author.}
\contact{\texttt{haowu.83@sjtu.edu.cn}\qquad\texttt{xyshen@eitech.edu.cn}}

\begin{document}
\maketitle
\begin{abstract}
Multimodal large language models (MLLMs) incur substantial inference costs when processing long visual-textual sequences. While existing operation compression methods exploit modality-level redundancy, they largely treat computation within attention heads and shared feed-forward network (FFN) channels as unified units, leaving finer-grained redundancy underexplored. We find that redundancy varies both across modality-interaction paths within the same attention head and across visual and textual executions of the same FFN channel. Based on these findings, we propose \emph{Modality-aware Width-wise Operation Pruning (MWOP)}, which independently prunes visual-to-visual (V2V), text-to-visual (T2V), and text-to-text (T2T) attention paths within each layer, and separately selects FFN channels for visual and textual inputs. A first-order Taylor criterion guides the pruning process, with FFN importance re-evaluated after attention pruning and LoRA-based recovery training. To translate the resulting fine-grained sparsity into practical acceleration, we further develop path-sparse Triton attention kernels and compact visual-side FFN execution. MWOP preserves the token sequence while reducing attention and FFN computation, making it complementary to token compression and enabling simultaneous reduction of sequence length and per-token computation. On LLaVA-OneVision-7B, MWOP alone achieves a $1.6\times$ prefill speedup with 99.7\% average performance retention across 12 benchmarks. Combined with two representative token compression methods, it further increases their prefill speedups from $2.0\times$ and $1.9\times$ to $2.9\times$ and $2.7\times$, respectively. Results on Qwen2.5-VL-7B further demonstrate its applicability across architectures.
The code is available at \url{https://github.com/EIT-NLP/MWOP}.
\end{abstract}

\section{Introduction}
\label{sec:introduction}
Multimodal large language models (MLLMs) process hundreds or thousands of visual tokens together with textual instructions through a shared Transformer decoder~\citep{li2024llava,lin2024preserve,bai2025qwen25vltechnicalreport}. The resulting long sequences incur substantial inference overhead, which can be reduced in two complementary ways: \emph{reducing the number of tokens being processed} or \emph{reducing the computation performed on each token}. The former, known as \emph{token compression}, prunes or merges redundant visual tokens~\citep{zhang2024sparsevlm, wen2025token}, and has been extensively studied with diverse token importance criteria and pruning strategies~\citep{xing2024pyramiddrop, yang2025visionzip,fan2025visipruner,wu2026hidrop}. The latter corresponds to \emph{operation pruning}, which reduces the computation applied to the retained tokens by skipping or sparsifying attention and FFN operations~\citep{liu2025fine, li2025redundancylens, zhang2024treat, yuan2025shortv}.

\begin{figure}[htbp]
    \centering
    \includegraphics[width=\linewidth]{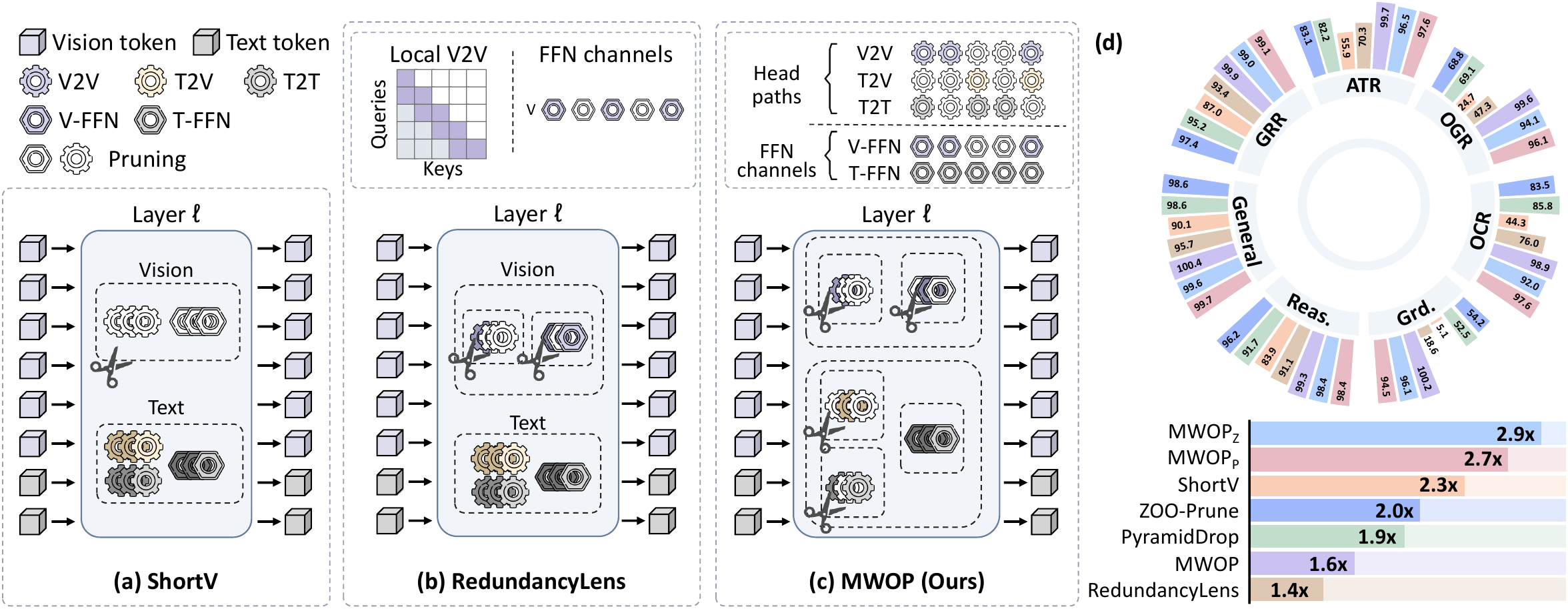}
    \caption{\small \textbf{Comparison of operation compression methods.} (a) ShortV skips visual token updates in selected layers. (b) RedundancyLens sparsifies visual-side attention and FFN computation. (c) MWOP performs finer-grained width compression by independently pruning V2V, T2V, and T2T attention paths and modality-conditioned FFN channel executions while preserving the token sequence. (d) MWOP achieves a 1.6$\times$ prefill speedup with negligible performance loss. Combining it with token compression increases the speedup to 2.9$\times$.}
    \label{fig:first}
\end{figure}

Compared with token compression, operation pruning has received considerably less attention in MLLMs. Existing operation pruning methods are still relatively \emph{coarse-grained}. As illustrated in Fig.~\ref{fig:first}(a--c), they typically remove computation by skipping the updates of visual tokens in selected layers~\citep{yuan2025shortv,liu2026vica}, restricting the attention range of visual tokens, or removing attention heads with limited visual activity~\citep{zhang2024treat,li2025redundancylens}. These methods demonstrate substantial redundancy in MLLM computation, but largely treat the computation within each attention layer or FFN as a unified unit. Such a treatment overlooks the fact that different computations within and across layers can play different roles and exhibit different levels of redundancy.

In fact, the computation in an MLLM layer is inherently heterogeneous. Under causal attention with visual tokens preceding textual tokens, each attention head contains three modality-interaction paths: visual-to-visual (\VtwoV{}), text-to-vision (\TtwoV{}), and text-to-text (\TtwoT{}). We find that these paths exhibit distinct redundancy and pruning sensitivity: \VtwoV{} spans a broader range of layers and heads and is more important for grounding, whereas \TtwoV{} is concentrated in fewer heads and tolerates more pruning, while \TtwoT{} requires more conservative compression (Section~\ref{sec:attn}). A head can also be redundant along one path while remaining important along another. Similarly, despite shared channel parameters, FFNs show modality- and depth-dependent redundancy, with visual-side executions substantially more redundant than text-side ones and stronger redundancy in early and late layers than in middle layers (Section~\ref{sec:ffn}). These findings motivate \emph{fine-grained, nonuniform width-wise operation pruning}, with independent \VtwoV{}, \TtwoV{}, and \TtwoT{} pruning across attention heads and separate FFN channel selection for visual and textual inputs.

Building on these insights, we propose \textbf{Modality-aware Width-wise Operation Pruning (MWOP)}. Comparisons of pruning granularity, importance criteria, and task sensitivity inform its staged design, which uses a common first-order Taylor criterion for both attention paths and modality-specific FFN channel executions~\citep{molchanov2019importance}. MWOP first selects head--path pairs sequentially for \VtwoV{}, \TtwoV{}, and \TtwoT{}, evaluating pruning budgets across task families to expose degradation hidden by aggregate scores. Crucially, attention pruning alters FFN channel importance. MWOP therefore recomputes modality-specific FFN Taylor scores under the selected attention masks before allocating channel budgets nonuniformly across layers. Candidate configurations undergo LoRA-based recovery to assess their post-training performance. Fixed offline masks, path-sparse attention, and compact visual-side FFNs translate these selections into practical acceleration without online selection overhead.

To translate these pruning decisions into practical speedup, we develop a path-sparse Triton attention kernel that skips masked query--key and attention--value products while maintaining a shared online softmax over retained key intervals~\cite{hu2026beyond}. For FFNs, we pack selected projection slices into compact visual-side weights while preserving full-width text execution. The optimized Qwen implementation further uses operator fusion to reduce data movement during modality splitting, activation, and output merging. These adaptations turn pruning decisions into measured latency reductions without online selection overhead.

On LLaVA-OneVision-7B, MWOP prunes 40\% of \VtwoV{}, 60\% of \TtwoV{}, and 10\% of \TtwoT{} head--path blocks, together with 50\% of visual-side FFN channel executions, while preserving all text-side FFN executions. Operation pruning alone reduces LLM-side FLOPs by \textbf{43.2\%}, retains \textbf{99.7\%} of dense-model performance across 12 benchmarks, and achieves a \textbf{1.6$\times$ decoder-prefill speedup}, \emph{approaching the idealized compute-only ceiling of 1.76$\times$ under linear FLOPs-to-latency scaling}. By preserving the token sequence, MWOP also complements token compression: combining it with ZOO-Prune or PyramidDrop provides an additional \textbf{1.4--1.5$\times$ speedup}, reaching up to \textbf{2.9$\times$} over the dense model. Further evaluations support its applicability to Qwen2.5-VL-7B and transfer to multi-image and video inputs.

Our contributions are threefold:
\textbf{(1)} We show that redundancy differs across paths within the same attention head and across modalities within the same FFN channel, motivating finer-grained pruning decisions.
\textbf{(2)} We develop MWOP, a fine-grained operation pruning framework that integrates head-wise path-level attention pruning and modality-conditioned FFN channel pruning through Taylor-guided staged selection and recovery training.
\textbf{(3)} We demonstrate substantial standalone acceleration with near-dense performance, additional gains beyond token compression, applicability across two MLLM architectures and multiple input formats.
\section{Related Work}
\label{sec:related_work}

Visual \emph{token compression} methods can be categorized by the stage at which token reduction occurs: (1) Encoder-level methods prune or merge patch tokens during visual encoding~\citep{bolya2022token, wu2026utptrack}. (2) Projector-level methods reduce visual token redundancy between the visual encoder and the LLM. PruMerge~\citep{shang2025llava} and VisionZip~\citep{yang2025visionzip} shorten visual token sequences by merging highly similar tokens. (3) LLM-level methods compress visual tokens within the LLM. FastV~\citep{chen2024image} and PyramidDrop~\citep{xing2024pyramiddrop} prune visual tokens using importance scores derived from text-to-visual attention.

\emph{Operation compression} skips or sparsifies attention and FFN operations while preserving the token sequence. Methods are broadly categorized into depth and width compression based on the targeted computational units. Depth compression skips visual token processing in selected LLM layers. ShortV~\citep{yuan2025shortv} freezes visual token attention and FFN updates in selected layers, while ViCA~\citep{liu2026vica} removes these updates throughout the LLM and retains \TtwoV{} interactions only at selected layers. HiDrop~\citep{wu2026hidrop} combines late injection and early exit with progressive token pruning. Width compression targets computation within layers. RedundancyLens~\citep{li2025redundancylens} uses local \VtwoV{} attention and dynamic FFN channel selection for visual tokens. In contrast, MWOP performs finer-grained width compression through head-wise path-level attention pruning and modality-conditioned selection of FFN channel executions, while remaining compatible with token compression.

\section{MWOP: Formulation and Overview}
\label{sec:preliminaries}

\paragraph{Modality-Specific Paths in Self-Attention.}
\label{sec:attention-paths}

Excluding system tokens from the formulation, we consider visual and user-text tokens in a unified sequence. At layer $l$, let $V^{(l)}$ and $T^{(l)}$ denote their respective hidden states. For head $h$, the output $O^{(l,h)}$ before output projection takes the blockwise form
\begingroup
\setlength{\abovedisplayskip}{3pt}
\setlength{\belowdisplayskip}{3pt}
\setlength{\abovedisplayshortskip}{3pt}
\setlength{\belowdisplayshortskip}{3pt}
\begin{equation*}
O^{(l,h)}
=
\begin{bmatrix}
A_{VV}^{(l,h)} & A_{VT}^{(l,h)}\\
A_{TV}^{(l,h)} & A_{TT}^{(l,h)}
\end{bmatrix}
\begin{bmatrix}
V^{(l)}W_V^{(l,h)}\\
T^{(l)}W_V^{(l,h)}
\end{bmatrix}
\end{equation*}
\endgroup
where $W_V^{(l,h)}$ is the value projection shared across modalities, and the first and second subscripts of each attention block indicate query and key/value modalities, respectively. Under causal attention with visual tokens preceding user-text tokens, $A_{VT}^{(l,h)}=0$, leaving three paths: \VtwoV{}, \TtwoV{}, and \TtwoT{}. These paths support visual aggregation, cross-modal transfer, and textual propagation, respectively. Together with the layer-wise importance patterns in Section~\ref{sec:attn} and Appendix~\ref{app:attn_layerwise}, this functional progression motivates the staged pruning order \VtwoV{} $\rightarrow$ \TtwoV{} $\rightarrow$ \TtwoT{}.

For path-level masking analysis, we independently gate each path's additive contribution. Although \TtwoV{} and \TtwoT{} share softmax normalization, we zero either contribution without renormalizing the retained attention weights. This enables evaluation of each path's impact, treating each head--path pair as a pruning unit.

\paragraph{Modality-Conditioned Computation in the FFN.}
\label{sec:ffn-formulation}

At layer $l$, let $x^{(l)}$ denote a token state, $\Phi^{(l)}(x^{(l)})\in\mathbb{R}^{d_{\mathrm{ffn}}^{(l)}}$ its intermediate FFN activation, and $W_{\mathrm{out}}^{(l)}$ the output projection. Here, $d_{\mathrm{ffn}}^{(l)}$ is the intermediate width, and $\Phi^{(l)}$ encompasses the input projections and channel-wise nonlinearities of standard or gated FFNs. For modality $\mu\in\{V,T\}$, let $\mathcal{I}_\mu^{(l)}\subseteq\{1,\ldots,d_{\mathrm{ffn}}^{(l)}\}$ index the retained channels. Omitting bias terms for brevity, the pruned FFN is evaluated as
\begingroup
\setlength{\abovedisplayskip}{3pt}
\setlength{\belowdisplayskip}{3pt}
\setlength{\abovedisplayshortskip}{3pt}
\setlength{\belowdisplayshortskip}{3pt}
\begin{equation*}
\widehat{\operatorname{FFN}}_\mu^{(l)}(x^{(l)})
=
\Phi_{\mathcal{I}_\mu^{(l)}}^{(l)}(x^{(l)})
\,W_{\mathrm{out}}^{(l)}[\mathcal{I}_\mu^{(l)},:]
\end{equation*}
\endgroup
where $\Phi_{\mathcal{I}_\mu^{(l)}}^{(l)}$ computes only the retained channels using the corresponding parameter slices, and $[\mathcal{I}_\mu^{(l)},:]$ selects the matching output-projection rows. Since FFNs process attention-updated token states, we re-evaluate channel importance after attention pruning. Parameters remain shared across modalities, but $\mathcal{I}_V^{(l)}$ and $\mathcal{I}_T^{(l)}$ may differ. We therefore treat the \emph{visual-side} and \emph{text-side} executions of each shared FFN channel as separate pruning units.

\paragraph{Overview of MWOP.}
\label{sec:overview}

\begin{figure}[t]
    \centering
    \includegraphics[width=\linewidth]{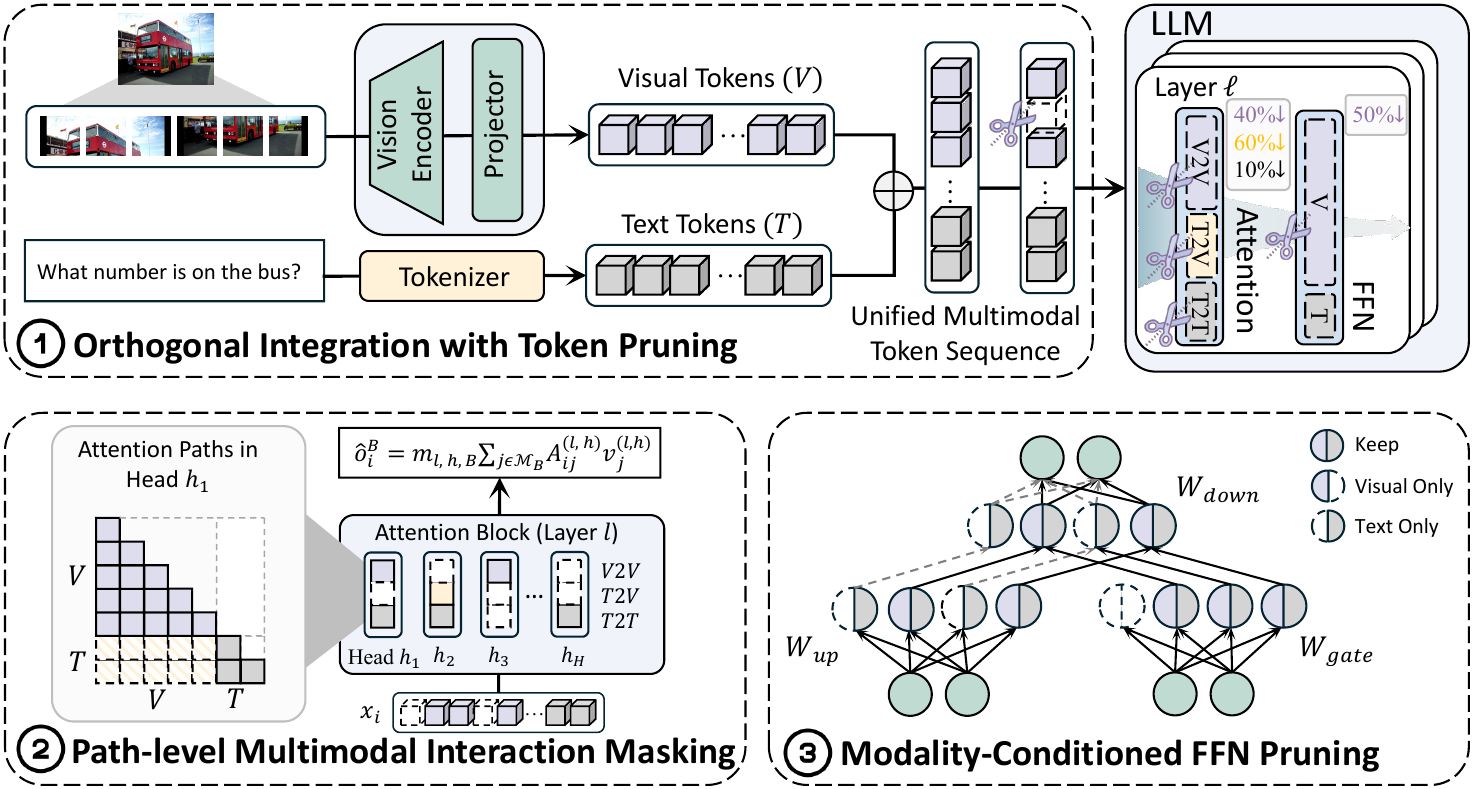}
    \caption{\small \textbf{Overview of MWOP.} (1) MWOP performs width pruning while preserving the token sequence and remaining compatible with token pruning. (2) \VtwoV{}, \TtwoV{}, and \TtwoT{} paths are masked independently within each attention head. (3) Shared FFN channels are selectively executed for visual tokens, textual tokens, or both.}
    \label{fig:mwop-overview}
\end{figure}

As illustrated in Figure~\ref{fig:mwop-overview}, MWOP performs head-wise path-level attention pruning in the order \VtwoV{} $\rightarrow$ \TtwoV{} $\rightarrow$ \TtwoT{}, guided by first-order Taylor importance. With the attention masks fixed, it recomputes FFN channel importance using the same Taylor criterion and selects channels separately for visual and textual tokens. LoRA-based post-training recovery then mitigates the performance loss from pruning. MWOP preserves the token sequence and is therefore orthogonal to token pruning. Appendices~\ref{app:taylor} and~\ref{app:ffn_3_metric} detail Taylor importance estimation and present comparisons with other importance criteria. Algorithm~\ref{alg:mwop_algorithm} in the appendix summarizes staged pruning and configuration selection. Following this staged procedure, we first analyze attention-path redundancy in Section~\ref{sec:attn} and then modality-conditioned FFN redundancy under the selected attention mask in Section~\ref{sec:ffn}.
\begin{figure}[!b]
    \centering
    \includegraphics[width=\textwidth]{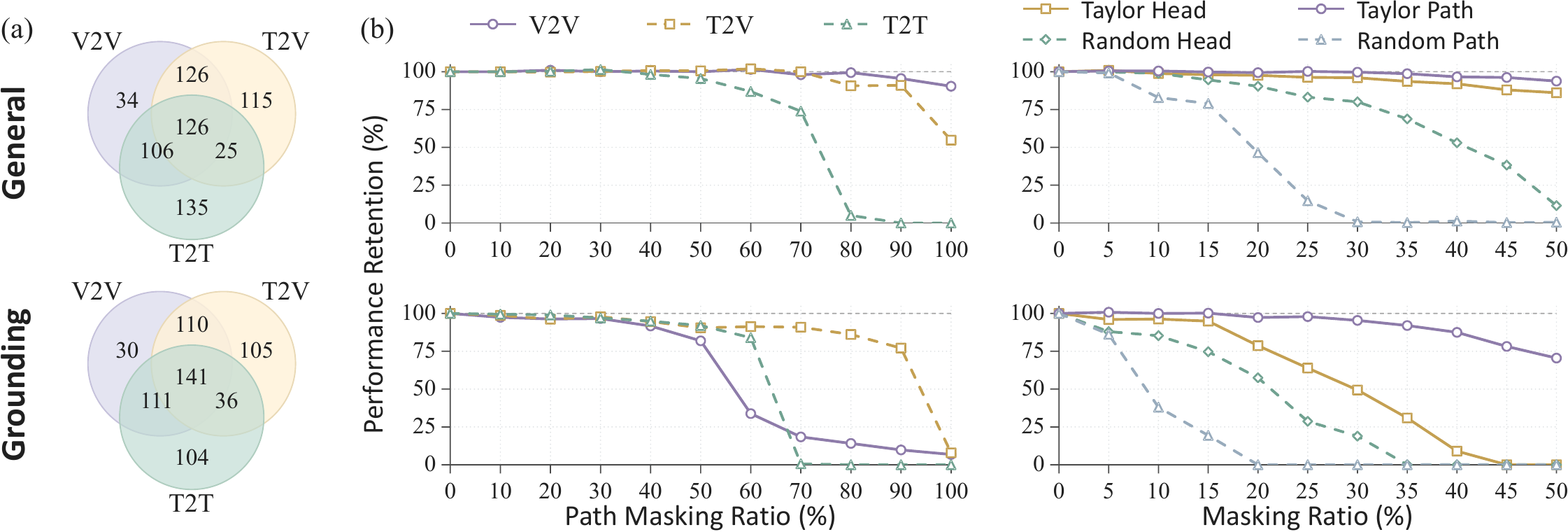}
    \caption{\small \textbf{Path-wise attention redundancy.} Rows show General (top) and Grounding (bottom). (a) Overlap of the bottom-50\% Taylor-ranked head sets across paths. (b) Single-path masking with the other paths kept dense (left) and whole-head versus head-wise path-level masking using Taylor or random selection at matched numbers of masked path blocks (right). All analyses are training-free, with performance retention measured relative to the original dense LLaVA-OneVision-7B model.}
    \label{app:taylor}
    \label{fig:path-overview}
\end{figure}

\vspace{-5pt}
\section{Path-Wise Redundancy in Self-Attention}
\label{sec:attn}

\begin{table}[!t]
\centering
\caption{\small \textbf{Attention-path pruning with post-training recovery.} Setting indicates the path whose masking ratio is varied at each stage. Scores are normalized to the dense reference. ATR reports overall retention. GRR covers General and Reasoning, while OGR covers OCR and Grounding. Superscript $\mathrm{R}$ denotes uniformly random block selection for the marked path, while unmarked paths use Taylor selection. Taylor and random counterparts share the same masking ratios and recovery protocol. \textbf{Bold} and \underline{underlined} values indicate the \textbf{best} and \underline{second-best} results within each stage, respectively. Gray shading marks the selected configurations. Per-benchmark results are provided in Appendix~\ref{app:detail_result_ov}.}
\label{tab:path-mask-ratio}
\resizebox{\textwidth}{!}{
\begin{tabular}{
    c|
    *{3}{>{\centering\arraybackslash}p{1.1cm}}|
    *{3}{>{\centering\arraybackslash}p{1.1cm}}|
    *{4}{>{\centering\arraybackslash}p{1.1cm}}
}
\toprule
\multirow{2}{*}{\textbf{Setting}}
& \multicolumn{3}{c|}{\textbf{Mask Ratio (\%)}}
& \multicolumn{3}{c|}{\textbf{Aggregate (\%)}}
& \multicolumn{4}{c}{\textbf{Category (\%)}} \\
\cmidrule(lr){2-4} \cmidrule(lr){5-7} \cmidrule(lr){8-11}
& \textbf{\VtwoV} & \textbf{\TtwoV} & \textbf{\TtwoT} 
& \textbf{ATR} & \textbf{GRR} & \textbf{OGR}
& \textbf{Gen.} & \textbf{Reas.} & \textbf{OCR} & \textbf{Grd.} \\
\midrule
Dense & 0 & 0 & 0 & 100.0 & 100.0 & 100.0 & 100.0 & 100.0 & 100.0 & 100.0 \\
\midrule
\multirow{5}{*}{\VtwoV} 
& 30 & 0 & 0 & \textbf{100.1} & \underline{100.1} & \textbf{100.1} & 100.3 & 99.9 & \textbf{100.1} & \textbf{100.1} \\
& \graycell{40} & \graycell{0} & \graycell{0} & \graycell{\underline{100.0}} & \graycell{\textbf{100.3}} & \graycell{\underline{99.7}} & \graycell{\textbf{100.6}} & \graycell{\textbf{100.1}} & \graycell{\underline{99.4}} & \graycell{\underline{99.9}} \\
& $40^{\mathrm{R}}$ & 0 & 0 & 96.0 & 98.7 & 93.4 & 99.8 & 97.6 & 95.3 & 91.5 \\
& 50 & 0 & 0 & 99.7 & \textbf{100.3} & 99.2 & \underline{100.5} & \underline{100.0} & 98.9 & 99.5 \\
& 60 & 0 & 0 & 98.8 & 99.8 & 97.8 & 100.2 & 99.5 & 97.1 & 98.5 \\
\midrule
\multirow{6}{*}{\TtwoV} 
& 40 & 50 & 0 & \textbf{100.4} & \underline{100.0} & \textbf{100.8} & \underline{100.1} & \underline{99.9} & \textbf{100.4} & \textbf{101.2} \\ 
& \graycell{40} & \graycell{60} & \graycell{0} & \graycell{\underline{100.3}} & \graycell{\textbf{100.1}} & \graycell{\underline{100.6}} & \graycell{\textbf{100.3}} & \graycell{\textbf{100.0}} & \graycell{\underline{99.9}} & \graycell{\textbf{101.2}} \\
& 40 & $60^{\mathrm{R}}$ & 0 & 95.0 & 95.6 & 94.5 & 97.7 & 93.4 & 96.0 & 92.9 \\
& $40^{\mathrm{R}}$ & $60^{\mathrm{R}}$ & 0 & 91.3 & 92.6 & 90.0 & 97.6 & 87.5 & 96.0 & 83.9 \\
& 40 & 70 & 0 & 99.5 & 99.6 & 99.4 & 99.5 & 99.7 & 99.0 & \underline{99.8} \\
& 40 & 80 & 0 & 98.5 & 98.7 & 98.3 & 98.5 & 98.9 & 97.4 & 99.2 \\
\midrule
\multirow{6}{*}{\TtwoT} 
& \graycell{40} & \graycell{60} & \graycell{10} & \graycell{\textbf{100.0}} & \graycell{\textbf{99.9}} & \graycell{100.1} & \graycell{\textbf{100.3}} & \graycell{\underline{99.5}} & \graycell{99.9} & \graycell{100.3} \\
& 40 & 60 & $10^{\mathrm{R}}$ & 99.7 & 99.0 & 100.3 & 99.3 & 98.7 & \underline{100.0} & 100.5 \\
& $40^{\mathrm{R}}$ & $60^{\mathrm{R}}$ & $10^{\mathrm{R}}$ & 89.4 & 92.7 & 86.1 & 96.0 & 89.3 & 90.5 & 81.8 \\
& 40 & 60 & 20 & \underline{99.9} & \textbf{99.9} & 99.9 & \underline{99.9} & \textbf{99.8} & \textbf{100.1} & 99.6 \\
& 40 & 60 & 30 & 99.7 & \underline{99.1} & \underline{100.2} & \underline{99.9} & 98.4 & 99.4 & \underline{101.1} \\
& 40 & 60 & 40 & 99.4 & 98.5 & \underline{100.4} & 99.8 & 97.1 & 99.3 & \textbf{101.4} \\
\bottomrule
\end{tabular}
}
\vspace{-10pt}
\end{table}

\paragraph{Redundancy Is Path-Conditional and Task-Dependent.}
\label{sec:path-asymmetry}
We conduct training-free analyses on LLaVA-OneVision-7B using representative General and Grounding benchmarks (Section~\ref{sec:exp_setup}). Within each task family, we rank heads separately for each path by aggregated first-order Taylor importance. Each bottom-50\% set contains 392 of 784 heads, yet only 126 heads on General and 141 on Grounding are shared across all three sets (Figure~\ref{fig:path-overview}(a)). Low importance along one path therefore does not imply low importance along others, motivating separate pruning decisions within each head. We then evaluate single-path masking on 1,000 examples from these benchmarks, keeping the other paths dense (Figure~\ref{fig:path-overview}(b), left). General tolerates substantial \VtwoV{} and \TtwoV{} masking, whereas Grounding degrades sharply beyond 40\% \VtwoV{} masking but retains over 90\% of baseline performance at 70\% \TtwoV{} masking. \TtwoT{} masking causes performance collapse near 80\% on General and 70\% on Grounding. These differences motivate path-specific budgets evaluated across task families, with more conservative \VtwoV{} pruning to preserve Grounding performance. Analyses for all four task families are provided in Appendix~\ref{app:path_venn_4}.

\vspace{-5pt}
\paragraph{Granularity and Importance-Guided Selection Both Matter.}
\label{sec:path-vs-head}
At masking ratio $q$, whole-head masking removes a fraction $q$ of heads, while head-wise path-level masking independently removes the same fraction from each path, matching masked path-block counts up to rounding (Figure~\ref{fig:path-overview}(b), right). At 30\% masking, Taylor-guided path-level masking retains 95.3\% performance versus 49.3\% for whole-head masking on Grounding, and 99.6\% versus 96.1\% on General. Taylor also outperforms random selection at both granularities at this ratio, supporting the combined benefits of finer pruning granularity and importance-guided selection.

\paragraph{Recovering a General-Purpose Path Mask.}
\label{sec:general-path-mask}
The training-free analysis motivates asymmetric budgets, particularly more conservative \VtwoV{} than \TtwoV{} masking for Grounding. We evaluate candidates sequentially for \VtwoV{}, \TtwoV{}, and \TtwoT{} after LoRA-based recovery, fixing earlier masks and keeping later paths dense. Hereafter, dense references are post-trained on the same data unless otherwise specified. We adopt $(40\%,60\%,10\%)$, achieving 100.0\% ATR and 99.5\%--100.3\% retention across task families (Table~\ref{tab:path-mask-ratio}). Raising \TtwoT{} masking to 40\% lowers Reasoning retention to 97.1\%, despite 99.4\% ATR. Randomizing all masks at these ratios yields 89.4\% ATR under identical recovery, showing that Taylor's benefit persists.

\section{Modality-Aware Redundancy in the FFN}
\label{sec:ffn}

\begin{figure}[!t]
    \centering
    \includegraphics[width=\textwidth]{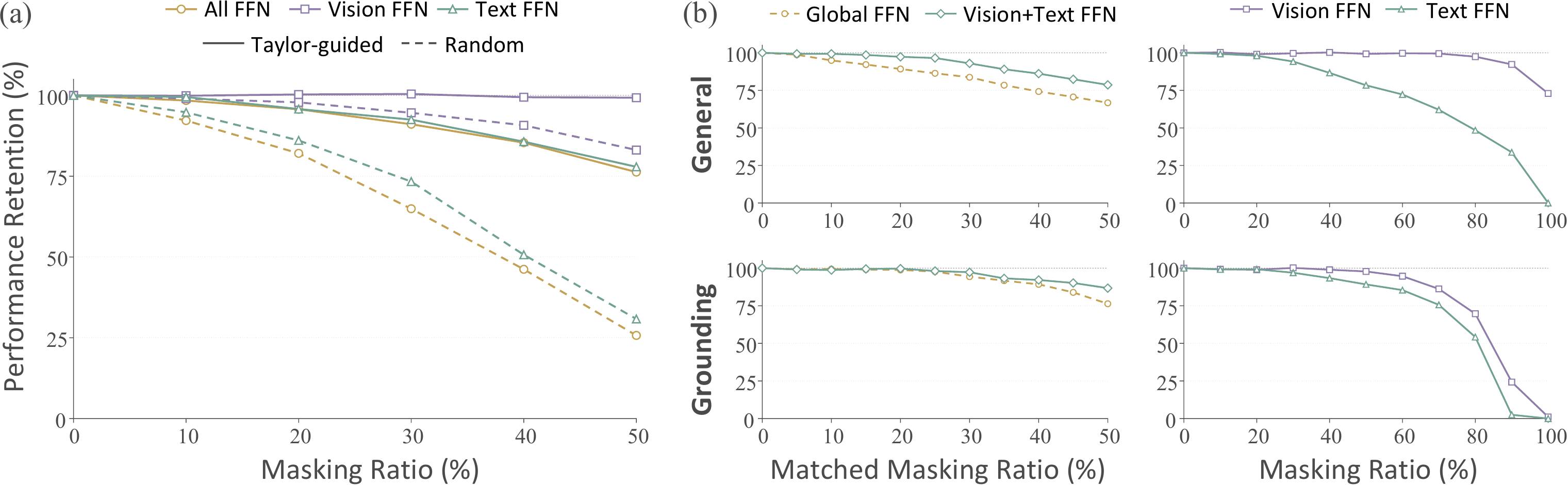}
    \caption{\small \textbf{Modality-conditioned FFN redundancy.} (a) Taylor (solid) vs. random (dashed) selection for All, Vision, and Text masking, with retention averaged equally across four task families. (b) Shared (Global) vs. modality-specific (Vision+Text) channel masking at matched per-modality ratios (left) and sensitivity to visual-only vs. text-only masking (right). All analyses are training-free, with performance measured relative to the attention-pruned model $\mathrm{MWOP}_{\mathrm{Attn}}$ with dense FFNs.}
    \label{fig:ffn-overview}
\end{figure}

\begin{table}[!t]
\centering
\caption{\small \textbf{Modality-conditioned FFN pruning with post-training recovery.} Visual and Text denote FFN masking ratios. Scores are normalized to the dense reference. Superscript $\mathrm{R}$ denotes uniformly random channel selection for the marked modality, while unmarked masks use Taylor selection. Taylor and random counterparts share the same masking ratios and recovery protocol. Per-benchmark results are provided in Appendix~\ref{app:detail_result_ov}.}
\label{tab:ffn-progressive-search}
\resizebox{\textwidth}{!}{
\begin{tabular}{
    c|
    *{2}{>{\centering\arraybackslash}p{1.4cm}}|
    *{3}{>{\centering\arraybackslash}p{1.1cm}}|
    *{4}{>{\centering\arraybackslash}p{1.1cm}}
}
\toprule
\multirow{2}{*}{\textbf{Setting}}
& \multicolumn{2}{c|}{\textbf{Mask Ratio (\%)}}
& \multicolumn{3}{c|}{\textbf{Aggregate (\%)}}
& \multicolumn{4}{c}{\textbf{Category (\%)}} \\
\cmidrule(lr){2-3} \cmidrule(lr){4-6} \cmidrule(lr){7-10}
& \textbf{Visual} & \textbf{Text}
& \textbf{ATR} & \textbf{GRR} & \textbf{OGR}
& \textbf{Gen.} & \textbf{Reas.} & \textbf{OCR} & \textbf{Grd.} \\
\midrule
MWOP\textsubscript{Attn} & 0 & 0 & 100.0 & 99.9 & 100.1 & 100.3 & 99.5 & 99.9 & 100.3 \\
\midrule
\multirow{6}{*}{\shortstack{Visual\\Neuron}} 
& 40 & 0 & \textbf{100.0} & \textbf{99.9} & \textbf{100.0} & \underline{100.1} & \textbf{99.7} & \textbf{99.7} & \textbf{100.3} \\
& $40^{\mathrm{R}}$ & 0 & 96.4 & 97.3 & 95.6 & 99.0 & 95.6 & 94.3 & 96.9 \\
& \graycell{50} & \graycell{0} & \graycell{\underline{99.7}} & \graycell{\textbf{99.9}} & \graycell{\underline{99.6}} & \graycell{\textbf{100.4}} & \graycell{\underline{99.3}} & \graycell{\underline{98.9}} & \graycell{\underline{100.2}} \\
& $50^{\mathrm{R}}$ & 0 & 94.9 & 96.3 & 93.6 & 98.5 & 94.2 & 92.7 & 94.4 \\
& 60 & 0 & 99.3 & \underline{99.7} & 98.9 & 100.0 & 99.4 & 98.2 & 99.6 \\
& 70 & 0 & 97.0 & 96.3 & 97.8 & 94.4 & 98.2 & 97.2 & 98.4 \\
\midrule
\multirow{5}{*}{\shortstack{Text\\Neuron}}
& \graycell{40} & \graycell{10}
& \graycell{\textbf{99.5}} & \graycell{\textbf{99.4}} & \graycell{\textbf{99.6}}
& \graycell{\textbf{99.8}} & \graycell{\textbf{99.0}} & \graycell{\textbf{99.0}} & \graycell{\textbf{100.2}} \\
& 40 & $10^{\mathrm{R}}$ & 98.1 & 97.3 & 98.9 & 97.6 & \underline{97.1} & 97.6 & \underline{100.1} \\
& $40^{\mathrm{R}}$ & $10^{\mathrm{R}}$ & 94.2 & 94.5 & 93.9 & 96.4 & 92.7 & 93.6 & 94.1 \\
& 40 & 20 & \underline{98.4} & \underline{97.4} & \underline{99.4} & \underline{98.3} & 96.5 & \underline{98.8} & 99.9 \\
& 40 & 30 & 97.0 & 95.0 & 99.1 & 95.9 & 94.0 & 98.2 & 99.9 \\
\bottomrule
\end{tabular}
}
\vspace{-5pt}
\end{table}

\paragraph{FFN Redundancy Is Modality-Conditioned.}
\label{sec:ffn-modality-redundancy}
Under the $(\VtwoV{},\TtwoV{},\TtwoT{})=(40\%,60\%,\\10\%)$ attention mask from Section~\ref{sec:attn}, we recompute first-order Taylor importance separately for visual- and text-side FFN channel executions, termed \emph{visual neurons} and \emph{text neurons}, respectively. At masking ratio $r$, Global masks the same bottom-$r$ channels for both modalities, whereas Vision+Text selects separate sets at matched per-modality ratios. Vision+Text retains more performance on General and Grounding, with a widening advantage at higher ratios (Figure~\ref{fig:ffn-overview}(b), left). Thus, parameter sharing does not imply identical redundancy across modalities. Training-free single-modality masking reveals asymmetric sensitivity (Figure~\ref{fig:ffn-overview}(b), right). Visual-side masking preserves General performance over a broad range, whereas text-side masking degrades earlier on both task families. Grounding is less tolerant than General of aggressive visual-side masking. These findings motivate modality-conditioned channel selection and asymmetric FFN pruning budgets. Analyses for all four task families appear in Appendix~\ref{app:ffn_4tasks}.

\paragraph{Granularity and Importance-Guided Selection Both Matter.}
\label{sec:ffn-granularity}
We compare Taylor and random selection at matched ratios within each FFN masking setting (Figure~\ref{fig:ffn-overview}(a)). Taylor outperforms random selection at every tested nonzero ratio. At 50\% masking, the gains are 50.5\%, 16.2\%, and 47.1\% for All, Vision, and Text masking, respectively. Visual-only masking remains nearly lossless with Taylor, compared with 83.1\% retention under random selection. Visual-only masking also retains more performance than text-only masking under random selection, indicating that modality asymmetry persists without Taylor ranking. Together with the shared-mask comparison in Section~\ref{sec:ffn-modality-redundancy}, these results support both modality-conditioned pruning granularity and importance-guided selection.

\paragraph{Recovering the Joint Attention--FFN Operating Point.}
\label{sec:attention-ffn-composition}
Guided by the training-free analysis, we evaluate FFN masking ratios after LoRA-based post-training recovery, keeping the attention mask fixed. With text channels dense, we select 50\% visual masking, retaining 99.7\% ATR and 98.9\%--100.4\% across task families (Table~\ref{tab:ffn-progressive-search}). Increasing visual masking to 60\% reduces OCR and Grounding retention, while 70\% lowers General retention to 94.4\%. A separate diagnostic comparison fixes visual masking at 40\% and varies text masking. Adding 10\% text masking reduces ATR from 100.0\% to 99.5\%, with larger ratios further degrading General and Reasoning. Random selection at 50\% visual masking yields 94.9\% ATR under identical recovery, showing that importance-guided selection remains beneficial after post-training.

\vspace{-5pt}
\section{Experiments}
\label{sec:experiments}

\subsection{Experimental Settings}
\label{sec:exp_setup}
\paragraph{Models.}
We conduct the main experiments on LLaVA-OneVision-7B~\citep{li2024llava} and extend the evaluation to Qwen2.5-VL-7B~\citep{bai2025qwen25vltechnicalreport} to assess the effectiveness of MWOP across different MLLM architectures. Architectural details are provided in Appendix~\ref{app:lmm}.

\vspace{-5pt}
\paragraph{Benchmarks.}
We evaluate MWOP on 12 benchmarks spanning 4 capability categories. General includes GQA~\citep{hudson2019gqa}, VQAv2~\citep{goyal2017making}, OK-VQA~\citep{marino2019ok}; Reasoning includes ScienceQA-IMG~\citep{lu2022learn}, AI2D~\citep{kembhavi2016diagram}, MMStar~\citep{chen2024we}; OCR includes TextVQA~\citep{singh2019towards}, DocVQA~\citep{mathew2021docvqa}, OCRBench~\citep{liu2024ocrbench}; Grounding includes RefCOCO, RefCOCO+, RefCOCOg~\citep{kazemzadeh2014referitgame, yu2016modeling, mao2016generation}. See Appendix~\ref{app:benchmarks} for details.

\vspace{-5pt}
\paragraph{Efficiency Evaluation.}
Following HiDrop~\citep{wu2026hidrop}, we report theoretical LLM FLOPs and measured prefill latency. Let $V_{\ell}$ and $T_{\ell}$ count visual and subsequent text tokens at layer $\ell$, excluding the system prefix, and $N_{\ell}=V_{\ell}+T_{\ell}$. Counting two FLOPs per multiply--accumulate,
\begingroup
\small
\setlength{\abovedisplayskip}{3pt}
\setlength{\belowdisplayskip}{3pt}
\setlength{\abovedisplayshortskip}{3pt}
\setlength{\belowdisplayshortskip}{3pt}
\begin{equation*}
\mathrm{FLOPs}
=
\sum_{\ell=1}^{L}
\Big[
4N_{\ell}d(d+d_{\mathrm{KV}})
+
4d\big(
\rho_{\mathrm{V2V},\ell}C(V_{\ell})
+
\rho_{\mathrm{T2V},\ell}T_{\ell}V_{\ell}
+
\rho_{\mathrm{T2T},\ell}C(T_{\ell})
\big)
+
6dm\big(
\rho_{V,\ell}V_{\ell}
+
\rho_{T,\ell}T_{\ell}
\big)
\Big]
\end{equation*}
\endgroup
where $d$, $d_{\mathrm{KV}}$, and $m$ denote hidden, key/value projection, and original FFN intermediate dimensions. $C(x)=x(x+1)/2$ counts causal pairs; $\rho$ denotes retained attention-path and modality-specific FFN channel ratios. FLOPs use average evaluation token counts $(V,T)$ of $(3215,19)$ for LLaVA-OneVision and $(485,38)$ for Qwen2.5-VL. Prefill latency is measured on an NVIDIA A100 40GB using a unified stack optimized for token split--merge, modality-specific FFNs, and attention-path skipping. Appendices~\ref{app:flops} and~\ref{app:efficient-inference} detail derivation and inference implementation.

\vspace{-5pt}
\paragraph{Implementation Details.}
We perform post-training recovery of the compressed model via one epoch of LoRA fine-tuning (rank 128, scaling factor 256, dropout 0.05) on a multimodal instruction mixture of 282K samples, detailed in Appendix~\ref{app:post-train-data}. Training runs on 8 NVIDIA A100 40GB GPUs with a global batch size of 256, a learning rate of $1\times10^{-5}$, 3\% warmup, and cosine decay. For a fair comparison, all methods evaluated with post-training recovery share the same recovery dataset. Method and training configurations are detailed in Appendices~\ref{app:method_config} and~\ref{app:train_config}, respectively.

\begin{table}[htbp]
\centering
\caption{\small \textbf{Comparison with operation compression methods on LLaVA-OneVision-7B at approximately 60\% retained LLM-side FLOPs.} Dep., Wid., and Tok. denote depth, width, and token compression; some compared methods also compress tokens. ATR, GRR, and OGR measure performance retention relative to the dense model.
Unshaded rows apply compression after post-training; red and blue rows use post-training recovery with compression applied, with blue denoting ours. \textsuperscript{$\dagger$} marks methods originally proposed without post-compression recovery but evaluated with it here.
MWOP achieves the best performance with the fewest FLOPs among compared methods, retaining 99.7\% overall (ATR) and 99.6\% fine-grained performance (OGR).
}
\label{tab:main-comparison-67}
\setlength{\tabcolsep}{2.0pt}
\resizebox{\textwidth}{!}{
\begin{tabular}{c|ccc|c|ccc|ccc|ccc|ccc|ccc}
\toprule
\multirow{2}{*}{\textbf{Method}} &
\multicolumn{3}{c|}{\textbf{Sparsity}} &
\multirow{2}{*}{\makecell{\textbf{Ret.}\\\textbf{(\%)}}} &
\multicolumn{3}{c|}{\textbf{General}} &
\multicolumn{3}{c|}{\textbf{Reasoning}} &
\multicolumn{3}{c|}{\textbf{OCR}} &
\multicolumn{3}{c|}{\textbf{Grounding}} &
\multirow{2}{*}{\makecell{\textbf{ATR}\\\textbf{(\%)}}} &
\multirow{2}{*}{\makecell{\textbf{GRR}\\\textbf{(\%)}}} &
\multirow{2}{*}{\makecell{\textbf{OGR}\\\textbf{(\%)}}} \\
& \textbf{Dep.} & \textbf{Wid.} & \textbf{Tok.}
&
& \textbf{GQA} & \textbf{VQA$^{\text{v2}}$} & \textbf{VQA$^{\text{ok}}$}
& \textbf{SQA} & \textbf{AI2D} & \textbf{MMStar}
& \textbf{VQA$^{\text{T}}$} & \textbf{VQA$^{\text{D}}$} & \textbf{OCRB}
& \textbf{Ref} & \textbf{Ref+} & \textbf{RefG}
& & & \\
\midrule
LLaVA-OneVision-7B
& - & - & - & 100.0
& 63.1 & 82.2 & 59.9
& 96.0 & 81.9 & 61.4
& 75.3 & 87.4 & 63.5
& 77.9 & 70.5 & 76.9
& 100.0 & 100.0 & 100.0 \\
\midrule

ShortV
& \cmark & & & 65.6
& 62.5 & 81.0 & 58.9
& 94.7 & 79.8 & 60.1
& 63.6 & 70.7 & 49.5
& 46.3 & 42.1 & 42.8
& 84.0 & 98.3 & 69.7 \\

RedundancyLens
& & \cmark & & 58.4
& 62.5 & 81.3 & 58.9
& 92.9 & 80.2 & 58.1
& 71.1 & 81.9 & 59.9
& 62.6 & 54.8 & 60.8
& 92.1 & 97.6 & 86.6 \\

HalfV
& \cmark & & \cmark & 60.4
& 62.6 & 81.4 & 59.7
& 95.6 & 81.4 & 59.9
& 73.1 & 85.1 & 55.0
& 52.7 & 46.4 & 47.8
& 89.2 & 99.1 & 79.4 \\

DOP\textsubscript{V}
& \cmark & \cmark & \cmark & 58.3
& 62.6 & 80.7 & 58.7
& 95.3 & 81.4 & 61.2
& 67.0 & 73.2 & 54.1
& 54.8 & 46.9 & 52.2
& 88.0 & 98.9 & 77.1 \\

\rowcolor{red!8}
ShortV\textsuperscript{$\dagger$}
& \cmark & & & 65.6
& 63.2 & \underline{81.8} & 60.0
& \underline{95.9} & 80.9 & 60.5
& 72.1 & 82.8 & 59.6
& \underline{76.3} & \underline{68.7} & \underline{74.7}
& 97.8 & 99.5 & 96.1 \\

\rowcolor{red!8}
RedundancyLens\textsuperscript{$\dagger$}
& & \cmark & & 58.4
& \underline{63.3} & 81.6 & 59.8
& 93.8 & \underline{81.3} & 59.2
& 73.5 & \underline{84.8} & \underline{62.6}
& 75.5 & 67.7 & \underline{74.7}
& \underline{98.0} & 98.8 & \underline{97.2} \\

\rowcolor{red!8}
YOPO
& \cmark & \cmark & & 70.0
& 61.2 & 77.9 & 57.1
& 85.2 & 73.2 & 50.1
& 40.0 & 29.0 & 24.4
& 34.2 & 27.2 & 28.9
& 66.0 & 91.1 & 40.8 \\

\rowcolor{red!8}
HalfV\textsuperscript{$\dagger$}
& \cmark & & \cmark & 60.4
& 63.1 & 81.5 & 59.7
& 95.5 & \textbf{81.7} & \textbf{60.8}
& 72.3 & 82.1 & 54.0
& 50.4 & 42.3 & 43.6
& 87.8 & 99.5 & 76.0 \\

\rowcolor{red!8}
Skip-Vision
& & \cmark & \cmark & 59.3
& 63.1 & 81.5 & \textbf{60.7}
& 95.3 & \underline{81.3} & \underline{60.7}
& \textbf{74.0} & 84.5 & 58.7
& 66.6 & 58.0 & 63.3
& 94.6 & \underline{99.6} & 89.6 \\

\rowcolor{red!8}
DOP\textsubscript{V}\textsuperscript{$\dagger$}
& \cmark & \cmark & \cmark & 58.3
& 63.1 & 81.5 & 59.9
& 94.9 & \underline{81.3} & 59.9
& 72.1 & 82.6 & 59.1
& 74.4 & 66.9 & 72.8
& 96.9 & 99.1 & 94.7 \\

\rowcolor{lightblue}
MWOP (ours)
& & \cmark & & \textbf{56.8}
& \textbf{63.4} & \textbf{82.0} & \underline{60.5}
& \textbf{96.2} & \underline{81.3} & 60.5
& \underline{73.8} & \textbf{85.0} & \textbf{64.5}
& \textbf{77.9} & \textbf{70.7} & \textbf{77.1}
& \textbf{99.7} & \textbf{99.9} & \textbf{99.6} \\

\bottomrule
\end{tabular}
}
\vspace{-5pt}
\end{table}

\begin{table}[!b]
\centering
\caption{\small
\textbf{Comparison of token and operation compression methods on LLaVA-OneVision-7B at approximately 30\% retained LLM-side FLOPs.}
MWOP\textsubscript{P} and MWOP\textsubscript{Z} combine MWOP with PyramidDrop and ZOO-Prune.
\textsuperscript{$\dagger$} denotes originally training-free methods evaluated with post-training recovery, while \textsuperscript{$\ddagger$} denotes originally training-based methods evaluated without post-training recovery. By composing operation and token compression, MWOP\textsubscript{Z} and MWOP\textsubscript{P} outperform all baselines overall while using fewer FLOPs.
}
\label{tab:main-comparison-33}
\setlength{\tabcolsep}{2.0pt}
\resizebox{\textwidth}{!}{
\begin{tabular}{c|ccc|c|ccc|ccc|ccc|ccc|ccc}
\toprule
\multirow{2}{*}{\textbf{Method}} &
\multicolumn{3}{c|}{\textbf{Sparsity}} &
\multirow{2}{*}{\makecell{\textbf{Ret.}\\\textbf{(\%)}}} &
\multicolumn{3}{c|}{\textbf{General}} &
\multicolumn{3}{c|}{\textbf{Reasoning}} &
\multicolumn{3}{c|}{\textbf{OCR}} &
\multicolumn{3}{c|}{\textbf{Grounding}} &
\multirow{2}{*}{\makecell{\textbf{ATR}\\\textbf{(\%)}}} &
\multirow{2}{*}{\makecell{\textbf{GRR}\\\textbf{(\%)}}} &
\multirow{2}{*}{\makecell{\textbf{OGR}\\\textbf{(\%)}}} \\
& \textbf{Dep.} & \textbf{Wid.} & \textbf{Tok.}
&
& \textbf{GQA} & \textbf{VQA$^{\text{v2}}$} & \textbf{VQA$^{\text{ok}}$}
& \textbf{SQA} & \textbf{AI2D} & \textbf{MMStar}
& \textbf{VQA$^{\text{T}}$} & \textbf{VQA$^{\text{D}}$} & \textbf{OCRB}
& \textbf{Ref} & \textbf{Ref+} & \textbf{RefG}
& & & \\
\midrule
LLaVA-OneVision-7B
& - & - & - & 100.0
& 63.1 & 82.2 & 59.9
& 96.0 & 81.9 & 61.4
& 75.3 & 87.4 & 63.5
& 77.9 & 70.5 & 76.9
& 100.0 & 100.0 & 100.0 \\
\midrule

FastV
& & & \cmark & 32.8
& 59.9 & 77.6 & 58.6
& 89.7 & 77.0 & 54.6
& 61.1 & 60.4 & 44.6
& 47.5 & 39.5 & 46.5
& 80.1 & 93.9 & 66.3 \\

PyramidDrop\textsuperscript{$\ddagger$}
& & & \cmark & 33.4
& 60.5 & 80.1 & 59.8
& 90.2 & 76.2 & 52.1
& 70.2 & 75.2 & 40.5
& 16.0 & 13.0 & 13.7
& 72.1 & 94.2 & 50.0 \\

VisionZip
& & & \cmark & 32.7
& 61.8 & 80.3 & 59.8
& 91.5 & 76.8 & 55.8
& 69.7 & 55.2 & 50.1
& 8.5 & 7.4 & 9.4
& 70.3 & 95.9 & 44.7 \\

DART
& & & \cmark & 32.8
& 62.2 & 80.9 & 59.3
& 92.0 & 78.0 & 57.1
& 71.7 & 75.6 & 54.9
& 22.3 & 19.4 & 20.8
& 77.6 & 96.7 & 58.6 \\

VisPruner
& & & \cmark & 32.7
& 62.6 & 79.6 & 58.9
& 92.8 & 79.0 & 57.5
& 68.9 & 61.8 & 52.7
& 35.1 & 28.8 & 36.5
& 80.0 & 96.8 & 63.1 \\

PruMerge+
& & & \cmark & 32.7
& 61.5 & 79.5 & 57.1
& 89.5 & 77.0 & 54.4
& 67.3 & 52.3 & 49.7
& 37.4 & 31.8 & 34.7
& 77.6 & 94.2 & 60.9 \\

ZOO-Prune
& & & \cmark & 32.7
& 63.1 & 80.1 & 58.9
& 93.9 & 79.1 & 57.8
& 69.0 & 64.4 & 54.2
& 42.7 & 36.8 & 42.7
& 83.1 & 97.4 & 68.8 \\

ShortV
& \cmark & & & 37.5
& 57.5 & 74.8 & 52.9
& 83.2 & 71.0 & 48.1
& 45.3 & 26.3 & 27.2
& 3.8 & 3.5 & 4.2
& 55.9 & 87.0 & 24.7 \\

RedundancyLens
& & \cmark & & 32.0
& 60.5 & 78.6 & 57.3
& 88.7 & 76.2 & 53.9
& 60.6 & 58.2 & 51.4
& 15.3 & 12.6 & 14.2
& 70.3 & 93.4 & 47.3 \\

HalfV
& \cmark & & \cmark & 33.2
& 48.3 & 60.4 & 44.5
& 82.1 & 69.3 & 41.3
& 35.0 & 22.3 & 11.9
& 1.9 & 1.4 & 1.8
& 46.6 & 77.0 & 16.2 \\

DOP\textsubscript{V}
& \cmark & \cmark & \cmark & 32.0
& 63.3 & 79.6 & 58.8
& 94.2 & 80.2 & 60.0
& 65.5 & 75.6 & 51.3
& 46.3 & 39.2 & 45.7
& 84.8 & 98.1 & 71.5 \\

\rowcolor{red!8}
FastV\textsuperscript{$\dagger$}
& & & \cmark & 32.8
& 61.2 & 78.6 & 59.4
& 90.6 & 78.9 & 55.1
& 63.7 & 64.6 & 45.5
& 59.0 & 49.8 & 56.6
& 85.2 & 95.4 & 75.0 \\

\rowcolor{red!8}
PyramidDrop
& & & \cmark & 33.4
& 61.9 & 80.7 & 59.7
& 90.8 & 76.5 & 53.5
& 70.1 & 76.4 & 48.8
& 44.0 & 35.9 & 38.7
& 82.2 & 95.2 & 69.1 \\

\rowcolor{red!8}
VisionZip\textsuperscript{$\dagger$}
& & & \cmark & 32.7
& 62.6 & 81.0 & 60.3
& 92.3 & 78.9 & 56.4
& 71.7 & 63.9 & 53.7
& 40.1 & 32.8 & 35.9
& 81.7 & 97.1 & 66.2 \\

\rowcolor{red!8}
DART\textsuperscript{$\dagger$}
& & & \cmark & 32.8
& 62.7 & 81.1 & 60.0
& 93.0 & 79.3 & 58.0
& \underline{72.5} & 79.8 & 58.0
& 31.9 & 27.5 & 29.6
& 82.0 & 97.7 & 66.2 \\

\rowcolor{red!8}
VisPruner\textsuperscript{$\dagger$}
& & & \cmark & 32.7
& 62.9 & 81.1 & 59.6
& 94.0 & 80.3 & 58.2
& 72.2 & 69.0 & 58.1
& 57.0 & 49.3 & 54.8
& 89.1 & 98.1 & 80.1 \\

\rowcolor{red!8}
PruMerge+\textsuperscript{$\dagger$}
& & & \cmark & 32.7
& 62.6 & 81.0 & \textbf{60.5}
& 91.2 & 78.5 & 56.3
& 70.4 & 60.5 & 53.2
& 54.8 & 46.2 & 51.5
& 85.9 & 96.9 & 74.9 \\

\rowcolor{red!8}
ZOO-Prune\textsuperscript{$\dagger$}
& & & \cmark & 32.7
& \textbf{63.4} & 81.4 & \underline{60.1}
& 94.8 & 80.3 & 58.3
& 72.0 & 72.2 & 58.6
& 64.0 & 55.6 & 60.7
& 91.8 & 98.6 & 85.1 \\

\rowcolor{red!8}
ShortV\textsuperscript{$\dagger$}
& \cmark & & & 37.5
& 62.3 & 80.0 & 57.3
& 88.5 & 77.9 & 55.2
& 63.8 & 63.7 & 51.4
& 55.1 & 47.1 & 50.2
& 84.2 & 94.8 & 73.5 \\

\rowcolor{red!8}
RedundancyLens\textsuperscript{$\dagger$}
& & \cmark & & 32.0
& 62.7 & 80.7 & 58.6
& 90.4 & 78.9 & 56.4
& 69.2 & 75.6 & 59.5
& 65.6 & 56.9 & 62.7
& 91.3 & 96.3 & 86.4 \\

\rowcolor{red!8}
YOPO
& \cmark & \cmark & & 34.7
& 58.9 & 74.1 & 53.5
& 84.0 & 71.9 & 48.6
& 22.6 & 17.2 & 14.2
& 29.3 & 21.4 & 23.3
& 58.1 & 87.8 & 28.4 \\

\rowcolor{red!8}
HalfV\textsuperscript{$\dagger$}
& \cmark & & \cmark & 33.2
& 55.9 & 72.0 & 55.3
& 85.6 & 71.9 & 47.1
& 45.9 & 34.8 & 26.7
& 22.8 & 16.2 & 18.5
& 61.8 & 87.0 & 36.5 \\

\rowcolor{red!8}
Skip-Vision
& & \cmark & \cmark & 32.0
& 63.0 & 80.9 & 59.3
& 94.6 & \underline{81.1} & \underline{60.1}
& 66.1 & 74.9 & 54.7
& 62.9 & 55.2 & 59.9
& 90.8 & 98.8 & 82.7 \\

\rowcolor{red!8}
DOP\textsubscript{V}\textsuperscript{$\dagger$}
& \cmark & \cmark & \cmark & 32.0
& \underline{63.3} & 81.4 & 60.0
& 94.7 & 80.6 & \textbf{60.5}
& 71.2 & \underline{82.7} & 57.9
& 65.7 & 57.4 & 64.1
& 93.7 & \textbf{99.1} & 88.2 \\

\rowcolor{lightblue}
MWOP\textsubscript{Z} (ours)
& & \cmark & \cmark & 27.9
& 62.9 & \underline{81.6} & 59.8
& \textbf{95.8} & \textbf{81.4} & 58.9
& 71.9 & 73.4 & \underline{61.4}
& \textbf{75.2} & \textbf{67.6} & \textbf{73.9}
& \underline{96.5} & \underline{99.0} & \underline{94.1} \\

\rowcolor{lightblue}
MWOP\textsubscript{P} (ours)
& & \cmark & \cmark & 30.8
& 62.9 & \textbf{81.7} & 60.0
& \underline{95.6} & 80.3 & 59.9
& \textbf{73.6} & \textbf{83.8} & \textbf{63.0}
& \underline{73.8} & \underline{66.3} & \underline{73.0}
& \textbf{97.6} & \textbf{99.1} & \textbf{96.1} \\

\bottomrule
\end{tabular}
}
\end{table}

\subsection{Main Result}
\label{sec:main-result}

\paragraph{Comparison with Operation Compression Methods.}
Table~\ref{tab:main-comparison-67} compares MWOP with operation compression methods at approximately 60\% retained LLM-side FLOPs. MWOP achieves the highest ATR, GRR, and OGR of 99.7\%, 99.9\%, and 99.6\%, respectively, while retaining the fewest FLOPs (56.8\%). Its advantage is particularly pronounced in fine-grained tasks: MWOP improves OGR by 2.4 points over RedundancyLens with recovery, the strongest baseline, with corresponding ATR and GRR gains of 1.7 and 1.1 points. These results show that MWOP maintains near-dense performance across general visual understanding, reasoning, OCR, and grounding.

\vspace{-5pt}
\paragraph{Composition with Token Compression.} 
Token and operation compression target complementary dimensions of LLM-side computation: token count and computation per retained token. Combining MWOP with PyramidDrop and ZOO-Prune yields MWOP\textsubscript{P} and MWOP\textsubscript{Z}. Table~\ref{tab:main-comparison-33} compares these variants with token and operation compression methods at approximately 30\% retained LLM-side FLOPs.
MWOP\textsubscript{P} and MWOP\textsubscript{Z} rank first and second
in ATR (97.6\%, 96.5\%), GRR (99.1\%, 99.0\%), and OGR (96.1\%, 94.1\%), while using fewer FLOPs (30.8\%, 27.9\%) than all baselines. Under the same recovery protocol, it improves ATR, GRR, and OGR over PyramidDrop by 15.4, 3.9, and 27.0 points, while MWOP\textsubscript{Z} improves over ZOO-Prune by 4.7, 0.4, and 9.0 points. 
MWOP\textsubscript{P} also surpasses DOP\textsubscript{V} with recovery, the strongest baseline, by 3.9, 0.0, and 7.9 points, respectively. These results show that combining operation and token compression preserves more performance than token compression alone, particularly on fine-grained tasks, at a comparable budget.

\begin{table}[!b]
\centering
\caption{\small
\textbf{Efficiency and cross-model generalization.}
\textbf{Left}: Theoretical FLOPs and practical prefill latency with breakdowns on two models. The component-removal configurations serve as references for potential speedup. Combining MWOP with token compression further reduces FLOPs and prefill latency on both architectures.
\textbf{Right}: Cross-model generalization on Qwen2.5-VL-7B at approximately 40\% retained LLM-side FLOPs, with standalone MWOP at 67.4\% included as a reference. Composition improves overall and fine-grained retention over token compression alone. Per-benchmark results are reported in Appendix Table~\ref{tab:qwen-main-comparison}.
}
\label{tab:effi}
\begingroup
\setlength{\tabcolsep}{2pt}
\begin{minipage}[t]{0.6\textwidth}
\vspace{0pt}
\centering
\resizebox{\linewidth}{!}{%
\begin{tabular}{l|ccc|ccc}
\toprule
\multirow{2}{*}{\textbf{Method}}
& \multicolumn{3}{c|}{\textbf{Theoretical (TFLOPs)}}
& \multicolumn{3}{c}{\textbf{Practical (Prefill Latency, ms)}} \\
\cmidrule(lr){2-4}
\cmidrule(lr){5-7}
& \textbf{Attn}
& \textbf{FFN}
& \textbf{Total}
& \textbf{Attn}
& \textbf{FFN}
& \textbf{Total} \\
\midrule

\rowcolor{gray!14}
\multicolumn{7}{c}{\textbf{LLaVA-OneVision-7B}} \\
\midrule
Dense
& 7.42 & 36.89 & 44.31
& 45.8 & 188.3 & 237.0 \\
\midrule

\makecell[l]{w/o Vis. Token}
& 0.03 & 0.22 & 0.25 
& 2.5~($18.1\times$)
& 9.7~($19.4\times$)
& 13.2~($18.0\times$) \\

w/o V2V\&T2V
& 5.32 & 36.89 & 42.21
& 32.7~($1.4\times$)
& 188.3~($1.0\times$)
& 221.5~($1.1\times$) \\

w/o FFN\textsubscript{V}
& 7.42 & 0.22 & 7.63 
& 45.8~($1.0\times$)
& 10.8~($17.4\times$)
& 56.9~($4.2\times$) \\

\midrule

ShortV
& 3.35 & 14.62 & 17.97
& 23.2~($2.0\times$)
& 80.5~($2.3\times$)
& 104.5~($2.3\times$) \\

RedundancyLens
& 6.60 & 22.22 & 28.82 
& 42.9~($1.1\times$)
& 130.4~($1.4\times$)
& 174.8~($1.4\times$) \\
\rowcolor{lightblue}
MWOP (ours)
& 6.57 & 18.60 & 25.17 
& 42.5~($1.1\times$)
& 107.4~($1.8\times$)
& 152.1~($1.6\times$) \\
\midrule

ZOO-Prune
& 3.20 & 18.55 & 21.75 
& 21.3~($2.2\times$)
& 96.3~($2.0\times$)
& 119.0~($2.0\times$) \\
\rowcolor{lightblue}
MWOP\textsubscript{Z} (ours)
& \underline{2.99}
& \textbf{9.38}
& \textbf{12.37}
& \textbf{20.4}~($\mathbf{2.2}\times$)
& \textbf{59.5}~($\mathbf{3.2}\times$)
& \textbf{81.2}~($\mathbf{2.9}\times$) \\
\midrule
PyramidDrop
& 3.42 & 18.55 & 21.96
& 23.1~($2.0\times$)
& 100.1~($1.9\times$)
& 124.9~($1.9\times$) \\
\rowcolor{lightblue}
MWOP\textsubscript{P} (ours)
& 3.16
& \underline{10.49}
& \underline{13.65} 
& \underline{21.9}~($\underline{2.1}\times$)
& 64.8~($2.9\times$)
& 88.2~($2.7\times$) \\
\midrule

\rowcolor{gray!14}
\multicolumn{7}{c}{\textbf{Qwen2.5-VL-7B}} \\
\midrule
Dense
& 0.91 & 5.97 & 6.88
& 8.2 & 40.7 & 49.8 \\
\midrule

\makecell[l]{w/o Vis. Token}
& 0.06 & 0.43 & 0.50 
& 2.5~($3.2\times$)
& 10.0~($4.1\times$)
& 13.5~($3.7\times$) \\

w/o V2V\&T2V
& 0.86 & 5.97 & 6.83
& 7.7~($1.1\times$)
& 38.0~($1.1\times$)
& 46.7~($1.1\times$) \\

w/o FFN\textsubscript{V}
& 0.91 & 0.43 & 1.35 
& 8.0~($1.0\times$)
& 10.4~($3.9\times$)
& 19.4~($2.6\times$) \\
\midrule

ShortV
& 0.46 & 2.61 & 3.07
& 5.8~($1.4\times$)
& \underline{22.3}~($\underline{1.8}\times$)
& \underline{28.7}~($\underline{1.7}\times$) \\

RedundancyLens
& 0.91 & 3.75 & 4.66 
& 8.8~($0.9\times$)
& 32.3~($1.3\times$)
& 41.8~($1.2\times$) \\
\rowcolor{lightblue}
MWOP (ours)
& 0.88 & 3.76 & 4.64 
& 7.8~($1.1\times$)
& 29.9~($1.4\times$)
& 38.8~($1.3\times$) \\
\midrule
ZOO-Prune
& 0.48 & 3.19 & 3.67
& 5.2~($1.6\times$)
& 30.2~($1.3\times$)
& 36.5~($1.4\times$) \\
\rowcolor{lightblue}
MWOP\textsubscript{Z} (ours)
& 0.47
& \textbf{2.09}
& \textbf{2.56} 
& \textbf{4.4}~($\mathbf{1.9}\times$)
& \textbf{19.7}~($\mathbf{2.1}\times$)
& \textbf{25.1}~($\mathbf{2.0}\times$) \\
\midrule
PyramidDrop
& 0.48 & 3.19 & 3.68 
& 5.1~($1.6\times$)
& 25.0~($1.6\times$)
& 31.3~($1.6\times$) \\
\rowcolor{lightblue}
MWOP\textsubscript{P} (ours)
& 0.47 & 2.44 & 2.91 
& \underline{4.6}~($\underline{1.8}\times$)
& 23.5~($1.7\times$)
& 29.6~($1.7\times$) \\
\bottomrule
\end{tabular}%
}
\end{minipage}\hfill
\begin{minipage}[t]{0.38\textwidth}
\vspace{0pt}
\centering
\includegraphics[
  width=\linewidth,
  height=10.3cm 
]{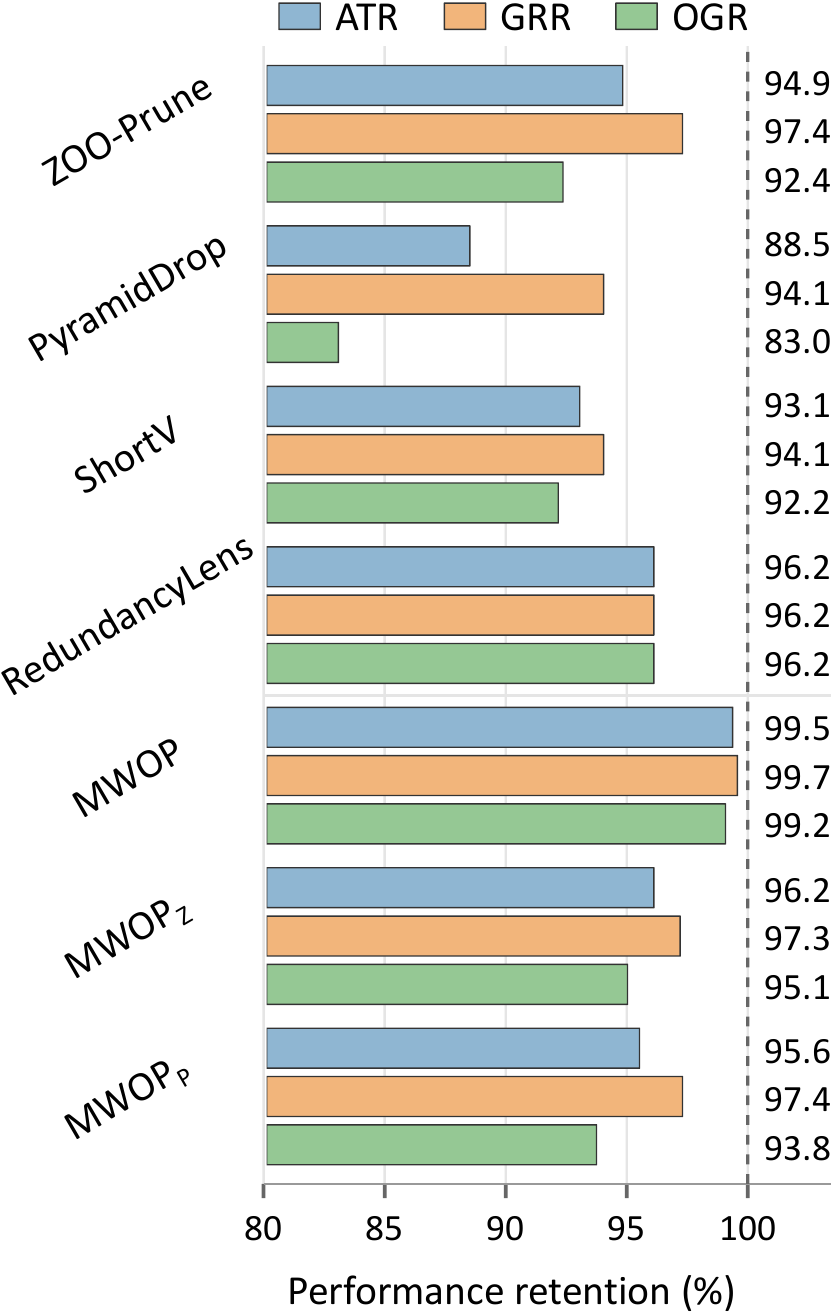}
\end{minipage}
\endgroup
\end{table}

\vspace{-5pt}
\paragraph{Efficiency and Cross-Model Generalization.}
Table~\ref{tab:effi} (left) reports efficiency evaluation on two MLLMs. Removing V2V and T2V gives only a 1.1$\times$ prefill speedup on both models, while removing visual-side FFNs yields 4.2$\times$ on LLaVA-OneVision and 2.6$\times$ on Qwen2.5-VL, indicating greater acceleration potential in visual-side FFNs at the evaluated input lengths.
On LLaVA-OneVision, MWOP reduces FLOPs by 43.2\% with a 1.6$\times$ prefill speedup, mainly from 1.8$\times$ faster FFNs. Composing MWOP with PyramidDrop and ZOO-Prune raises speedups from 1.9$\times$/2.0$\times$ to 2.7$\times$/2.9$\times$, respectively, with FLOP reductions of 69.2\%/72.1\%.
On Qwen2.5-VL, MWOP reduces FLOPs by 32.6\% with a 1.3$\times$ prefill speedup. Composition raises ZOO-Prune's speedup from 1.4$\times$ to 2.0$\times$ and PyramidDrop's from 1.6$\times$ to 1.7$\times$. Shorter visual sequences may limit gains through lower arithmetic intensity and greater relative memory-access and fixed overheads (Appendix~\ref{app:sequence-length}).
Table~\ref{tab:effi} (right) evaluates cross-model generalization on Qwen2.5-VL. For reference, standalone MWOP retains 67.4\% of LLM-side FLOPs and achieves 99.5\% ATR, 99.7\% GRR, and 99.2\% OGR. At approximately 40\% retained FLOPs, MWOP\textsubscript{Z} and MWOP\textsubscript{P} outperform their token-only counterparts by 1.3 and 7.1 points in ATR, 2.7 and 10.8 points in OGR, respectively. These results support MWOP's effectiveness across architectures and its complementary benefits to token compression, particularly on fine-grained tasks.
\section{Discussion}
\label{sec:discussion}
\paragraph{Task-Specialized Path Masks.}

\begin{wrapfigure}{r}{0.5\textwidth}
    \centering
    \vspace{-10pt}
    \includegraphics[width=\linewidth]{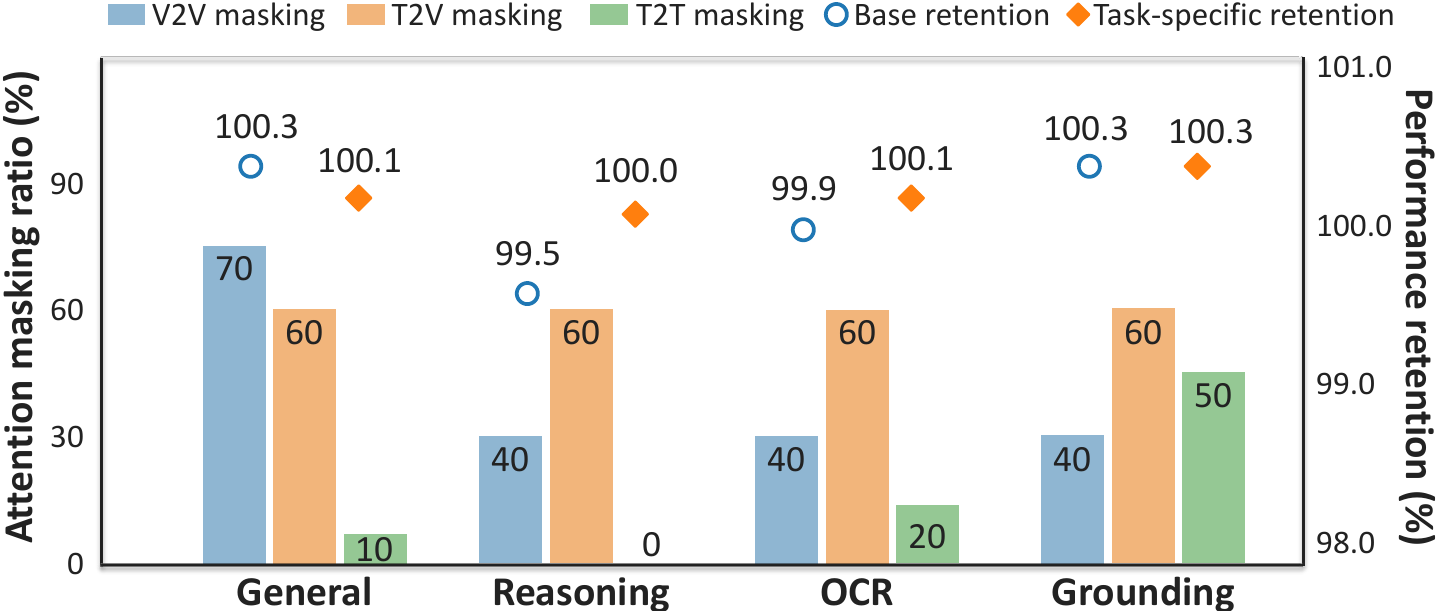}
    \caption{\small \textbf{Task-specialized attention-path compression.} Bars show path compression ratios, and markers show post-recovery performance retention.}
    \label{fig:task-specific-attention}
\end{wrapfigure}

Starting from the shared configuration (40\% \VtwoV{}, 60\% \TtwoV{}, 10\% \TtwoT{}), we adjust path compression ratios using task-specific Taylor profiles (Figure~\ref{fig:task-specific-attention}). After post-training recovery, increasing \VtwoV{} compression to 70\% for General retains 100.1\% performance, versus 100.3\% with shared masks. For Reasoning, retaining all \TtwoT{} paths raises retention from 99.5\% to 100.0\%. Increasing \TtwoT{} compression to 20\% for OCR and 50\% for Grounding yields 100.1\% and 100.3\% retention, versus 99.9\% and 100.3\%, respectively. Task-specialized masks thus allow stronger compression on selected paths while preserving target-category performance.

\paragraph{Multi-Image and Video Understanding.}
Table~\ref{tab:multiimage-video-compression} reports multi-image and video results after recovery solely on single-image data. Without additional multi-image or video fine-tuning, MWOP retains 100.4\% and 99.2\% of dense performance, respectively, achieving the highest video retention among the compared compression methods. These results demonstrate that MWOP preserves the dense model's capabilities across input formats. Incorporating multi-image or video data into recovery training may further improve performance.

\begin{table}[htbp]
\centering
\caption{\small 
\textbf{Multi-image and video evaluation on LLaVA-OneVision-7B.}  All methods are evaluated directly after post-training recovery on single-image data. Sub. denotes subtitles.
}
\setlength{\tabcolsep}{2pt}
\resizebox{\textwidth}{!}{
\begin{tabular}{l cccc ccccc}
\toprule
\multirow{3}{*}{\textbf{Settings}}
& \multicolumn{4}{c}{\textbf{Multi-Image}}
& \multicolumn{5}{c}{\textbf{Video}} \\
\cmidrule(lr){2-5}
\cmidrule(lr){6-10}

& \multirow{2}{*}{\textbf{Avg (\%)}}
& \multirow{2}{*}{\textbf{Q-Bench2}}
& \multirow{2}{*}{\textbf{Mantis-Eval}}
& \multirow{2}{*}{\textbf{BLINK}}
& \multirow{2}{*}{\textbf{Avg (\%)}}
& \multirow{2}{*}{\textbf{MVBench}}
& \multicolumn{2}{c}{\textbf{Video-MME}}
& \multirow{2}{*}{\textbf{NExT-QA}} \\
& & & & & &
& \textbf{w/o Sub.}
& \textbf{w/ Sub.}
& \\
\midrule
LLaVA-OneVision-7B 
& 100.0 & 72.6 & 54.8 & 48.8
& 100.0 & 57.9 & 58.7 & 62.0 & 78.7 \\
\midrule
\rowcolor{red!8}
ZOO-Prune\textsuperscript{$\dagger$}
& \textbf{100.8} & \textbf{75.0} & \underline{54.4} & \underline{48.7}
& \underline{98.9} & \underline{56.3} & \underline{58.0} & \textbf{61.4} & \underline{79.2} \\
\rowcolor{red!8}
PyramidDrop
& 90.6 & 70.9 & 46.5 & 43.6
& 96.9 & 54.9 & 56.7 & 60.0 & 78.2 \\
\rowcolor{red!8}
ShortV\textsuperscript{$\dagger$}
& 85.3 & 60.4 & 46.1 & 43.2
& 92.1 & 52.5 & 52.4 & 57.1 & 75.9 \\
\rowcolor{red!8}
RedundancyLens\textsuperscript{$\dagger$}
& 85.1 & 62.6 & 43.8 & 43.5
& 91.2 & 52.6 & 51.2 & 55.9 & 76.1  \\
\rowcolor{lightblue}
MWOP
& \underline{100.4} & 72.7 & \textbf{55.3} & \textbf{48.9}
& \textbf{99.2} & \textbf{56.8} & \textbf{58.1} & \underline{61.2} & \textbf{79.6} \\
\rowcolor{lightblue}
MWOP\textsubscript{P}
& 97.9 & \underline{73.2} & 53.5 & 46.5
& 93.6 & 53.0 & 53.9 & 57.9 & 77.0 \\
\rowcolor{lightblue}
MWOP\textsubscript{Z}
& 96.9 & 72.8 & 52.5 & 46.2
& 93.9 & 52.9 & 54.6 & 58.0 & 76.9 \\
\bottomrule
\end{tabular}
}
\label{tab:multiimage-video-compression}
\vspace{-5pt}
\end{table}
\section{Conclusion}
\label{sec:conclusion}

We identify fine-grained redundancy across modality-interaction paths within attention heads and modality-specific executions of shared FFN channels. Building on these findings, MWOP combines head-wise path-level attention pruning and modality-conditioned FFN channel pruning through Taylor-guided selection and post-training recovery. Experiments on two MLLM architectures show that MWOP reduces computation while preserving near-dense performance, including on fine-grained OCR and grounding tasks. Its composition with token compression further accelerates prefill while maintaining high performance retention, demonstrating the complementary benefits of reducing token count and computation per retained token. Future work will explore learnable gating to jointly optimize pruning decisions and compression budgets.

\bibliographystyle{plainnat}
\IfFileExists{EITNLP/references.bib}{\bibliography{EITNLP/references}}{\bibliography{references}}

\clearpage
\appendix
This appendix supplements the main text with further analyses of attention and FFN redundancy, detailed results across tasks and backbones, and experimental settings for pruning, recovery, and efficiency evaluation. It is organized as follows:
\begin{itemize}[leftmargin=1.35em, labelsep=0.45em]
    \item Section~\ref{app:attention_analysis}: attention-path importance, task-dependent redundancy, and calibration-task effects.
    \item Section~\ref{app:ffn_analysis}: modality-conditioned FFN importance, rescoring after attention pruning, and modality-specific masking sensitivity.
    \item Section~\ref{app:add_exp}: detailed pruning results across benchmarks and backbones, and input-length effects on computational efficiency.
    \item Section~\ref{app:implement_details}: model and pruning configurations, sequential search and recovery, training settings, and efficiency evaluation protocols.
    \item Section~\ref{app:methods_details}: descriptions of the comparison methods.
    \item Section~\ref{app:benchmarks}: descriptions of the evaluation benchmarks.
\end{itemize}

\section{More Attention Analysis}
\label{app:attention_analysis}
We extend the analysis in Section~\ref{sec:attn} with a detailed investigation of attention-path importance and redundancy across General, Reasoning, OCR, and Grounding. We first compare importance estimation criteria (Section~\ref{app:taylor}) and examine the layer-wise distribution of path importance (Section~\ref{app:attn_layerwise}). We then analyze path-conditional redundancy across task categories (Section~\ref{app:path_venn_4}), assess the cross-task similarity of importance rankings (Section~\ref{app:attn_cross_task}), and evaluate how the calibration task affects pruning performance (Section~\ref{app:attn_calibration}). Together, these analyses provide empirical support for path-wise pruning and inform the choice of importance criteria and calibration data.

\subsection{Attention Importance Estimation}
\label{app:taylor}
Each attention head is decomposed into three pruning units corresponding to \VtwoV{}, \TtwoV{}, and \TtwoT{}. We distinguish attention-weight statistics, output-contribution statistics, and loss-based importance, since these quantities characterize different aspects of a path.

\paragraph{Attention Mass.}
Attention mass measures the normalized attention weight assigned to a path. For a query token $i$ and a path with key positions $\mathcal K_b$, its path-specific mass is $\sum_{j\in\mathcal K_b} A_{ij}$, where $A$ is the attention matrix of the corresponding head. This statistic describes the distribution of attention across key modalities, without accounting for value vectors or sensitivity of the prediction loss.

\paragraph{Attention Contribution.}
Attention contribution characterizes the output produced by a path rather than its attention weights alone. For a path $b$ whose key/value modality is $\nu$, this contribution is $O_b=A_b U_\nu$, where $A_b$ is the corresponding attention block and $U_\nu$ contains the projected value vectors. We use the sample-averaged score $\|O_b\|_{\mathrm F}/|\mathcal Q_b|$, where $\|\cdot\|_{\mathrm F}$ denotes the Frobenius norm and $|\mathcal Q_b|$ is the number of query tokens. Contributions are measured before the output projection.

\paragraph{First-Order Taylor.}
For a pruning unit $u$, the Taylor score approximates the magnitude of the loss change induced by gating its output to zero~\citep{molchanov2019importance}:
\begin{equation*}
S(u)
=
\frac{1}{|\mathcal D_{\mathrm{cal}}|}
\sum_{x\in\mathcal D_{\mathrm{cal}}}
\left|
\sum_k
o_{u,k}(x)
\frac{\partial \mathcal L_x}{\partial o_{u,k}(x)}
\right|,
\end{equation*}
where $\mathcal D_{\mathrm{cal}}$ is the calibration set, $\mathcal L_x$ is the sample loss, and $o_{u,k}(x)$ is the $k$-th element of the unit's output. Within each sample, activation-gradient products are summed before taking the absolute value. The resulting magnitudes are then averaged across samples.

For attention, each unit is a path $b\in\{\mathrm{VV},\mathrm{TV},\mathrm{TT}\}$ of a head at a given layer. We compute the additive contribution $O_b=A_bU_\nu$ before the attention output projection. We use the following per-query-normalized Taylor score as a \emph{path contribution sensitivity proxy}:
\begin{equation*}
S_{\ell,h,b}^{\mathrm{attn}}
=\frac{1}{|\mathcal D_{\mathrm{cal}}|}
\sum_{x\in\mathcal D_{\mathrm{cal}}}
\frac{\left|\left\langle O_{\ell,h,b}(x),
\nabla_{O_{\ell,h}(x)}\mathcal L_x\right\rangle\right|}
{|\mathcal Q_b(x)|},
\end{equation*}
where $\mathcal Q_b(x)$ denotes the query set, $O_{\ell,h}(x)$ is the full head output before projection, and the inner product sums over queries and head dimensions. We use teacher-forced answer cross-entropy for $\mathcal L_x$. With attention weights held fixed, the score measures the loss sensitivity to gating $O_{\ell,h,b}$ and excludes the softmax renormalization induced by inference masking. At inference, we apply path masks before softmax and normalize over the retained causal keys. The sparse kernel skips the masked QK and attention-value products to reduce FLOPs (Section~\ref{app:efficient-inference}).

\begin{figure}[htbp]
    \centering
    \includegraphics[width=\linewidth]{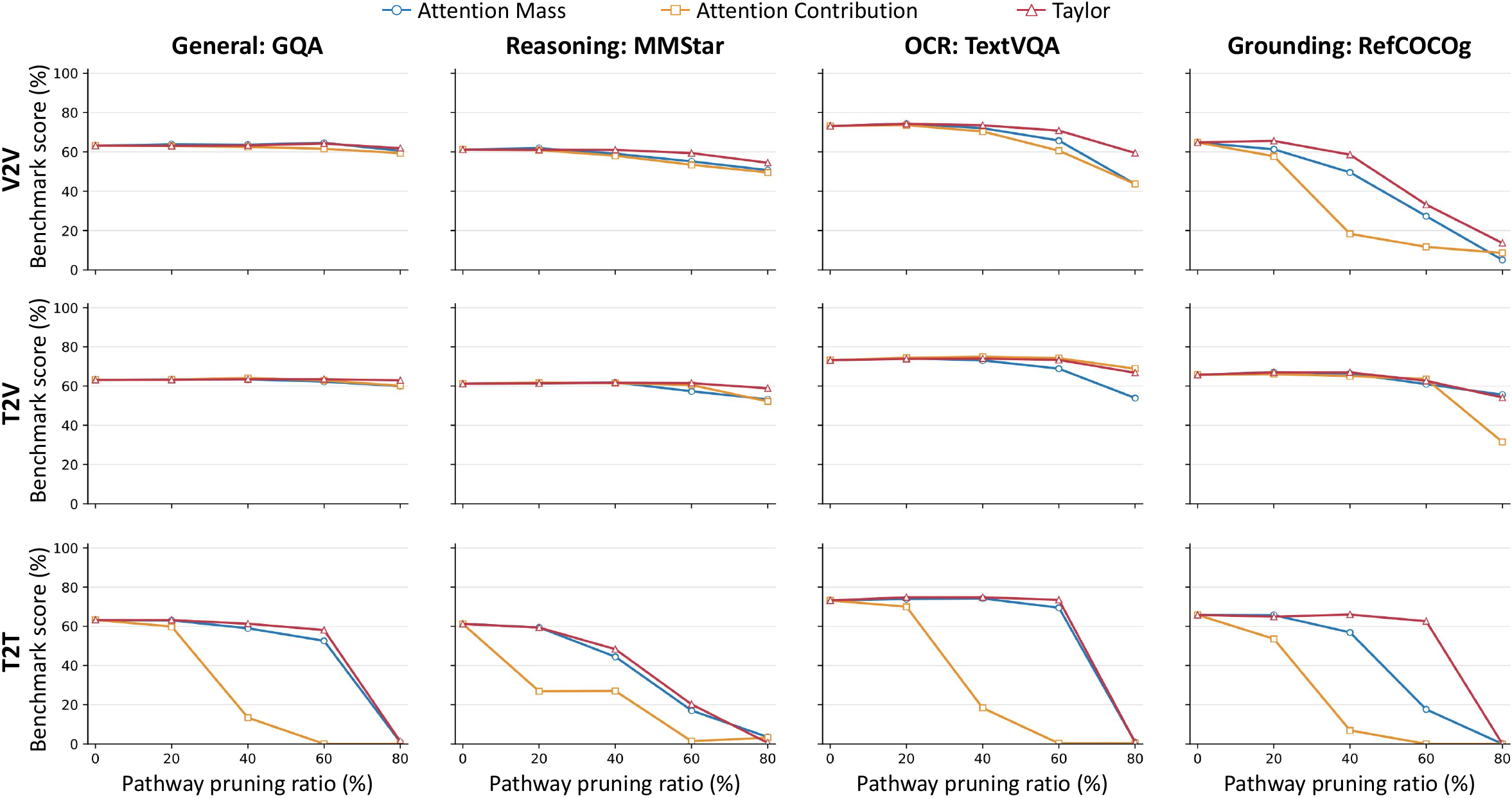}
    \caption{\small \textbf{Comparison of attention-path importance criteria on LLaVA-OneVision-7B.} Rows correspond to \VtwoV{}, \TtwoV{}, and \TtwoT{} masking. Columns respectively show GQA, MMStar, TextVQA, and RefCOCOg, representing General, Reasoning, OCR, and Grounding. Curves use Attention Mass, Attention Contribution, or first-order Taylor importance for training-free path selection.}
    \label{fig:appendix_attn_metric}
\end{figure}

\paragraph{Comparison of Importance Criteria.}
Figure~\ref{fig:appendix_attn_metric} compares the criteria across four representative benchmarks. Taylor generally preserves performance more effectively as masking increases, with clear advantages for \VtwoV{} on TextVQA and RefCOCOg and for \TtwoT{} across the evaluated benchmarks. Attention Contribution exhibits sharp \TtwoT{} performance drops at moderate masking ratios, while Attention Mass remains more competitive but degrades earlier than Taylor on RefCOCOg. Differences are smaller for \TtwoV{}, where Attention Contribution is competitive on TextVQA.

\subsection{Layer-Wise Distribution of Path Importance}
\label{app:attn_layerwise}

\begin{figure}[htbp]
    \centering
    \includegraphics[width=\linewidth]{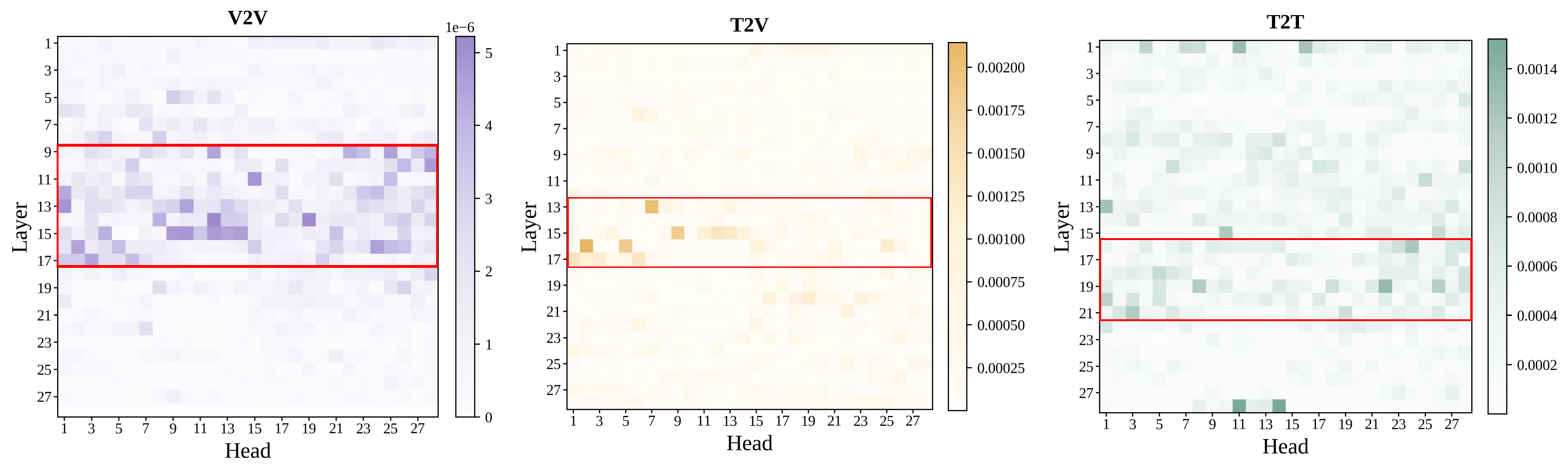}
    \caption{\small \textbf{Layer-head distribution of attention-path Taylor importance on LLaVA-OneVision-7B.} Panels show \VtwoV{}, \TtwoV{}, and \TtwoT{}. Rows and columns index decoder layers and query heads, using one-based labels. Darker cells indicate larger importance scores, and red boxes highlight the depth ranges discussed in the text.}
    \label{fig:appendix_attn_taylor_heatmaps}
\end{figure}

Figure~\ref{fig:appendix_attn_taylor_heatmaps} shows that the three paths have different depth profiles. High \VtwoV{} importance is concentrated approximately in layers 9 to 17 and becomes much weaker beyond layer 19. \TtwoV{} importance is more localized, with prominent bands around layers 13 to 17 and a small number of high-scoring heads. \TtwoT{} importance is distributed more broadly, with several prominent heads around layers 16 to 21 as well as isolated peaks in other decoder layers.

The overlapping high-importance regions are consistent with a gradual shift from visual processing to cross-modal integration and text-side computation. This pattern also motivates our sequential search order: first \VtwoV{}, then \TtwoV{}, and finally \TtwoT{}. The nonuniform profiles also suggest that uniform layer-wise allocation can overlook important paths. Each panel uses a separate color scale, with intensities indicating relative importance within that path.

\subsection{Path-Conditional Redundancy across Task Categories}
\label{app:path_venn_4}

\begin{figure}[htbp]
    \centering
    \includegraphics[width=\linewidth]{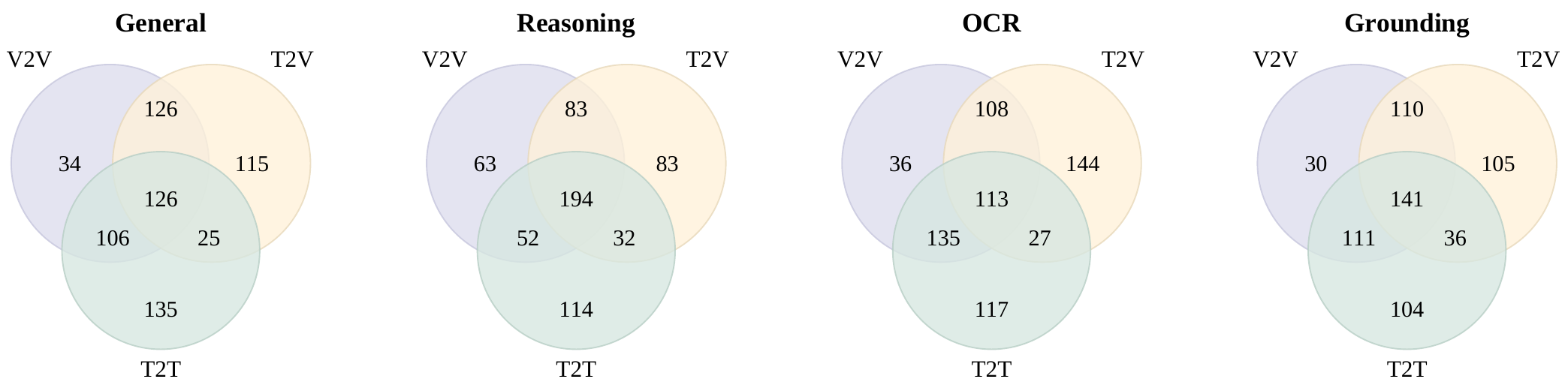}
    \caption{\small \textbf{Overlap of low-importance attention paths across four task categories.} Each circle contains the layer-head identities in the bottom 50\% of Taylor scores for one path: 392 of 784 identities on LLaVA-OneVision-7B. Region labels give the numbers of identities in the corresponding exclusive or shared sets. The central region contains heads whose three paths all lie in their respective low-importance sets.}
    \label{fig:appendix_attn_venn}
\end{figure}

\paragraph{Overlap of Low-Importance Paths.}
Figure~\ref{fig:appendix_attn_venn} extends the overlap analysis in Figure~\ref{fig:path-overview} to all four task categories. The three-way intersection accounts for only 32.1\%, 49.5\%, 28.8\%, and 36.0\% of each path's low-importance set for General, Reasoning, OCR, and Grounding. Thus, low importance on one path does not imply that the entire head is redundant.

Across all four categories, the \VtwoV{} and \TtwoV{} intersection is larger than the \TtwoV{} and \TtwoT{} intersection. This indicates that the low-importance patterns of the two visually related paths overlap more strongly, whereas text-side redundancy is less consistently aligned with cross-modal redundancy. These differences motivate selecting each path according to its own importance, rather than assigning a shared pruning decision to all three paths within a head.

\begin{figure}[htbp]
    \centering
    \includegraphics[width=\linewidth]{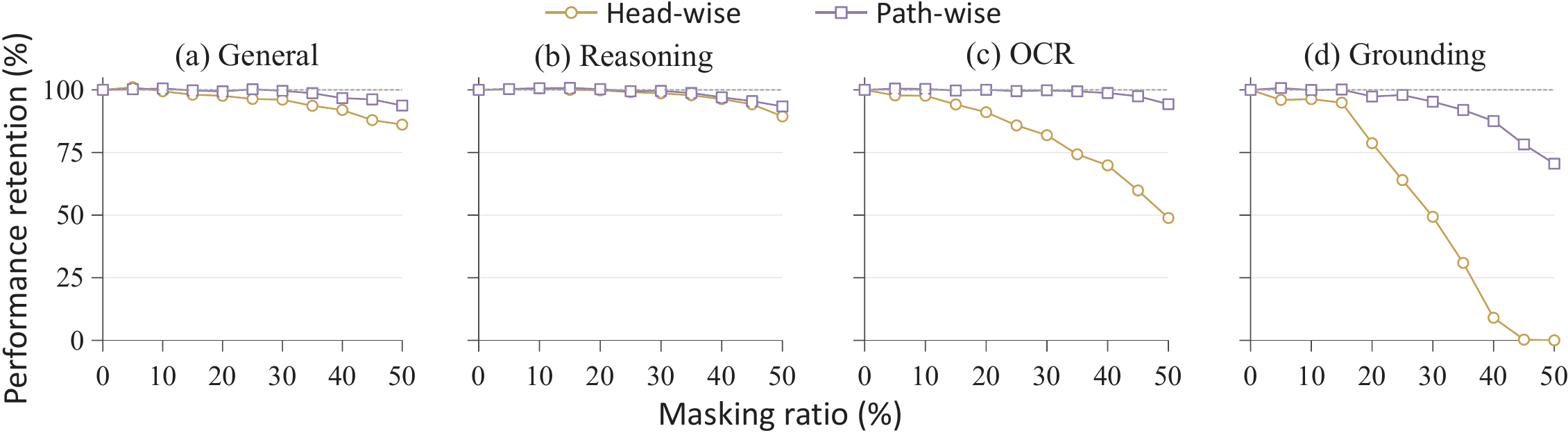}
    \caption{\small \textbf{Whole-head versus path-wise attention masking on LLaVA-OneVision-7B.} Both use Taylor importance. The x-axis gives the fraction of masked heads or the independently masked fraction of each path type, matching the total number of removed path blocks up to rounding. Panels show training-free performance retention on the four task categories relative to the dense reference. The dashed line denotes 100\% retention.}
    \label{fig:appendix_head_vs_path}
\end{figure}

\paragraph{Whole-Head versus Path-Wise Masking.}
Figure~\ref{fig:appendix_head_vs_path} compares Taylor-guided masking at the two granularities. At ratio $q$, head-wise masking removes a fraction $q$ of heads and all three of their paths, whereas path-wise masking selects a fraction $q$ independently within each path type. The advantage of independent path selection is especially pronounced on OCR and Grounding. At 30\% masking, Grounding retains 95.3\% performance under path-wise selection, compared with 49.3\% under whole-head selection. General and Reasoning exhibit smaller gaps over the same range.

\begin{figure}[htbp]
    \centering
    \includegraphics[width=\linewidth]{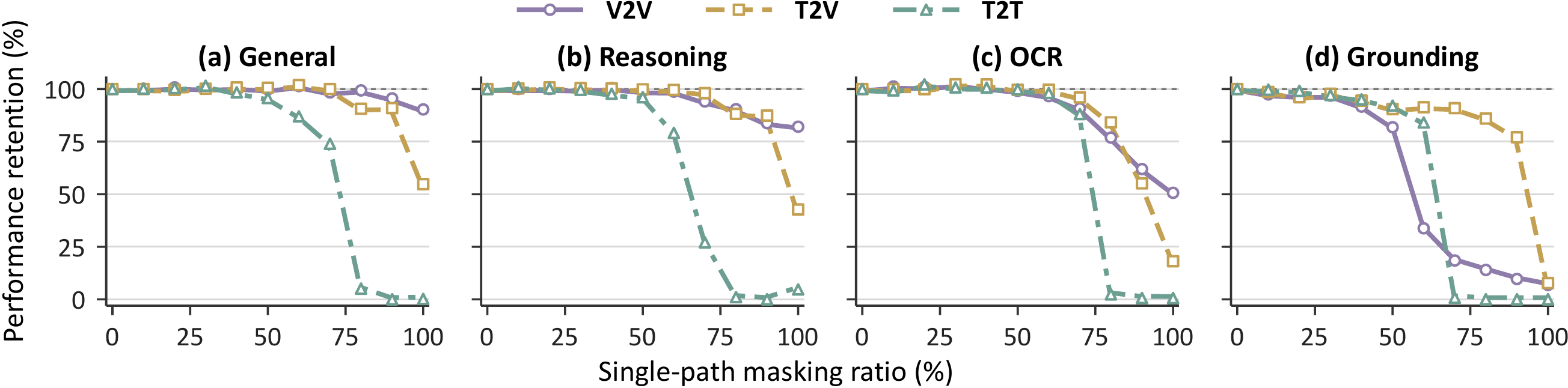}
    \caption{\small \textbf{Training-free sensitivity to single-path attention masking on LLaVA-OneVision-7B.} In each curve, Taylor-guided masking is applied only to the indicated path, with the other two paths kept dense. The horizontal axis gives the masking ratio of that path. The vertical axis gives performance retention relative to the dense reference. Panels cover the four task categories, and the dashed line denotes 100\% retention.}
    \label{fig:appendix_path_sweep_n1000}
\end{figure}

\paragraph{Sensitivity to Individual Paths.}
Figure~\ref{fig:appendix_path_sweep_n1000} isolates each path by masking it while keeping the other two dense. The sensitivity profiles differ substantially across tasks. General tolerates extensive \VtwoV{} masking, whereas Grounding deteriorates much earlier when the same path is removed. In contrast, Grounding retains over 90\% performance at 70\% \TtwoV{} masking. Aggressive \TtwoT{} masking produces a sharp collapse across categories, although the onset differs by task. These single-path interventions support using task-dependent pruning ratios rather than treating all three paths as equally redundant.

\subsection{Cross-Task Similarity of Attention Importance}
\label{app:attn_cross_task}

\begin{figure}[t]
    \centering
    \includegraphics[width=\linewidth]{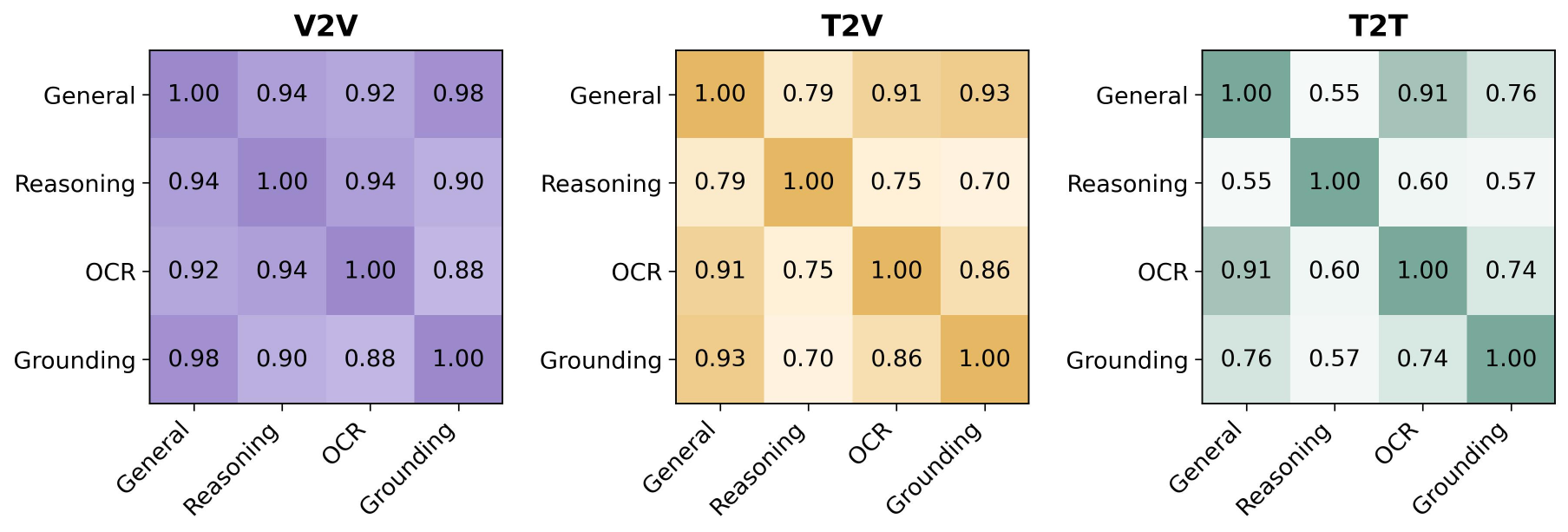}
    \caption{\small \textbf{Cross-task similarity of attention-path Taylor importance on LLaVA-OneVision-7B.} Each matrix compares the layer-head importance profiles of General, Reasoning, OCR, and Grounding for one path. Cell values report pairwise Spearman rank correlations, with larger values indicating more similar rankings.}
    \label{fig:appendix_attn_category_taylor_heatmaps}
\end{figure}

Figure~\ref{fig:appendix_attn_category_taylor_heatmaps} reports Spearman rank correlations between task-specific Taylor importance profiles, computed over the 784 layer-head identities separately for each path. \VtwoV{} has consistently high cross-task similarity, with off-diagonal values ranging from 0.88 to 0.98. The corresponding ranges are 0.70 to 0.93 for \TtwoV{} and 0.55 to 0.91 for \TtwoT{}. Reasoning is particularly distinct on \TtwoT{}, with similarities of 0.55, 0.60, and 0.57 to General, OCR, and Grounding, respectively.

These results indicate that visual-side importance is more consistent across the evaluated task categories, whereas text-side importance is more task-dependent. We further compare calibration tasks to evaluate how these ranking similarities translate into transferable pruning masks.

\subsection{Effect of the Calibration Task}
\label{app:attn_calibration}

We compare calibration tasks under fixed $(\VtwoV,\TtwoV,\TtwoT)$ masking ratios of $(40\%,60\%,10\%)$ and the same recovery protocol. Each single-category setting uses 256 calibration samples, while \textit{All Tasks} combines four category-specific rankings estimated from 256 samples each. The aggregate rank is $r(u)=\min_c r_c(u)$, where $r_c(u)$ is the rank of unit $u$ in category $c$ and rank 1 denotes highest importance, favoring units important to at least one category.

Table~\ref{tab:task-specific-config} shows that OCR-guided ranking achieves the highest single-category ATR (99.9\%), with category-wise retention ranging from 98.5\% to 100.8\%. In contrast, Reasoning-guided ranking yields 96.6\% ATR and only 90.3\% OCR retention, suggesting that OCR-based calibration produces a more transferable importance ranking in this comparison. With its larger calibration budget, \textit{All Tasks} achieves 100.0\% ATR and a narrower category-wise retention range of 99.5\% to 100.3\%.

\begin{table}[htbp]
\centering
\caption{\small \textbf{Effect of the calibration task on attention-path pruning.} Performance retention (\%) is measured relative to the unpruned, post-trained LLaVA-OneVision-7B.}
\label{tab:task-specific-config}

\begin{tabular}{l|ccc|cccc}
\toprule
\multirow{2}{*}{\textbf{Setting}}
& \multicolumn{3}{c|}{\textbf{Aggregate (\%)}}
& \multicolumn{4}{c}{\textbf{Category (\%)}} \\
\cmidrule(lr){2-4}
\cmidrule(lr){5-8}

& \textbf{ATR}
& \textbf{GRR}
& \textbf{OGR}
& \textbf{Gen.}
& \textbf{Reas.}
& \textbf{OCR}
& \textbf{Grd.} \\

\midrule
Dense
& 100.0 & 100.0 & 100.0 & 100.0 & 100.0 & 100.0 & 100.0 \\
\midrule
All Tasks
& \textbf{100.0} & \textbf{99.9} & \underline{100.1} & \textbf{100.3} & \textbf{99.5} & \underline{99.9} & \underline{100.3} \\
General
& 99.5 & 99.4 & 99.6 & \underline{100.2} & 98.6 & 98.9 & 100.2 \\
Reasoning
& 96.6 & 98.6 & 94.5 & 97.9 & \underline{99.3} & 90.3 & 98.8 \\
OCR
& \underline{99.9} & 99.1 & \textbf{100.7} & 99.8 & 98.5 & \textbf{100.6} & \textbf{100.8} \\
Grounding
& 99.3 & \underline{99.5} & 99.2 & 100.1 & 98.9 & 98.6 & 99.7 \\
\bottomrule
\end{tabular}
\end{table}

\section{More FFN Analysis}
\label{app:ffn_analysis}
We extend the analysis in Section~\ref{sec:ffn} with a detailed investigation of modality-conditioned FFN importance and redundancy. We first compare channel importance criteria (Section~\ref{app:ffn_3_metric}) and examine ranking changes and the effect of rescoring after attention pruning (Section~\ref{app:ffn_change_after_attn}). We then compare shared and modality-specific channel masks and analyze visual-side versus text-side sensitivity across General, Reasoning, OCR, and Grounding (Section~\ref{app:ffn_4tasks}). Together, these analyses inform FFN importance estimation and the allocation of pruning budgets between visual and text tokens.

\subsection{Importance of Modality-Conditioned FFN Channels}
\label{app:ffn_3_metric}
An FFN pruning unit is an intermediate channel on visual or text tokens. Let $h_{\ell,i,c}(x)$ be its gated activation immediately before the down projection. FFN weights are shared across modalities, but channel importance and the resulting masks are modality-specific. We compare three criteria.

\paragraph{Activation.}
For channel $c$ at layer $\ell$ and modality $\mu$, the mean absolute activation is
\begin{equation*}
S^{\mathrm{act}}_{\ell,c,\mu}
=\frac{\sum_{x\in\mathcal D_{\mathrm{cal}}}\sum_{i\in\mathcal I_\mu(x)}
|h_{\ell,i,c}(x)|}{\sum_{x\in\mathcal D_{\mathrm{cal}}}|\mathcal I_\mu(x)|},
\end{equation*}
where $\mathcal I_\mu(x)$ denotes the token positions of modality $\mu$. This score measures channel activity.

\begin{figure}[!b]
    \centering
    \includegraphics[width=\linewidth]{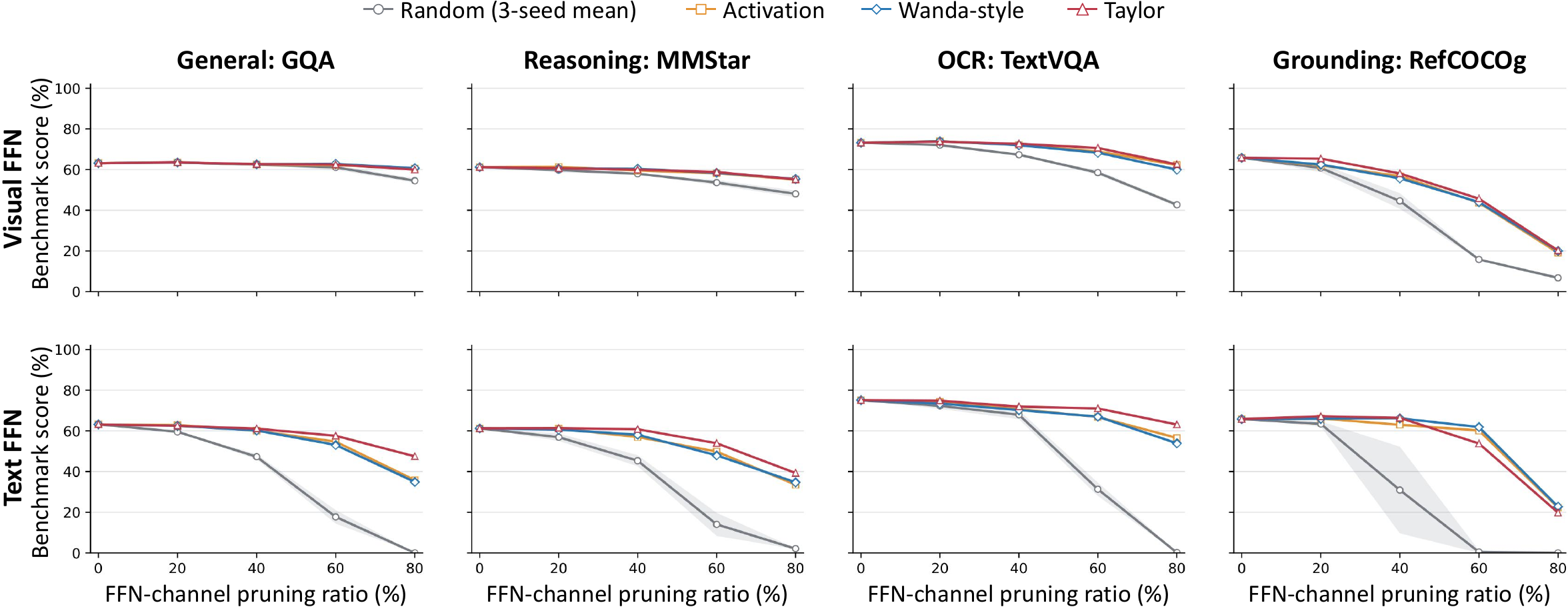}
    \caption{\small \textbf{Comparison of FFN-channel importance criteria on LLaVA-OneVision-7B.} Columns show GQA, MMStar, TextVQA, and RefCOCOg for General, Reasoning, OCR, and Grounding. The top and bottom rows respectively mask visual-side and text-side channels. Random curves average three seeds.}
    \label{fig:appendix_ffn_metric}
\end{figure}

\paragraph{Wanda.}
We adapt Wanda~\citep{sun2024simple} to structured channels by weighting activation with the Euclidean norm of the corresponding down-projection column:
\begin{equation*}
S^{\mathrm{Wanda}}_{\ell,c,\mu}
=\|W_{\mathrm{down},\ell}[:,c]\|_2 S^{\mathrm{act}}_{\ell,c,\mu}.
\end{equation*}

\paragraph{Taylor.}
For a modality-specific channel gate, the Taylor score is
\begin{equation*}
S^{\mathrm{gate}}_{\ell,c,\mu}
=\frac{1}{|\mathcal D_{\mathrm{cal}}|}\sum_{x\in\mathcal D_{\mathrm{cal}}}
\left|\sum_{i\in\mathcal I_\mu(x)}h_{\ell,i,c}(x)
\frac{\partial\mathcal L_x}{\partial h_{\ell,i,c}(x)}\right|.
\end{equation*}
We sum activation-gradient products over tokens within each sample before taking the absolute value, then average across samples. This score is used in Figure~\ref{fig:appendix_ffn_metric}. For joint pruning, FFN scores are recomputed on the initial model with the selected attention mask applied.

\paragraph{Comparison of Importance Criteria.}
For a controlled comparison, the three scores use the same calibration samples. For each criterion, we rank channels within each layer and task, average the ranks equally across the four tasks, and apply matched per-layer channel budgets.

Figure~\ref{fig:appendix_ffn_metric} shows that importance-guided masking generally outperforms random selection. The three criteria perform similarly for visual-side masking. Under aggressive text-side masking, Taylor has a clearer advantage on GQA, MMStar, and TextVQA, although it does not lead at every ratio on RefCOCOg. Thus, the benefit of loss-based ranking depends on both task and token modality.

\subsection{FFN Scoring after Attention Pruning}
\label{app:ffn_change_after_attn}

\begin{figure}[htbp]
    \centering
    \includegraphics[width=\linewidth]{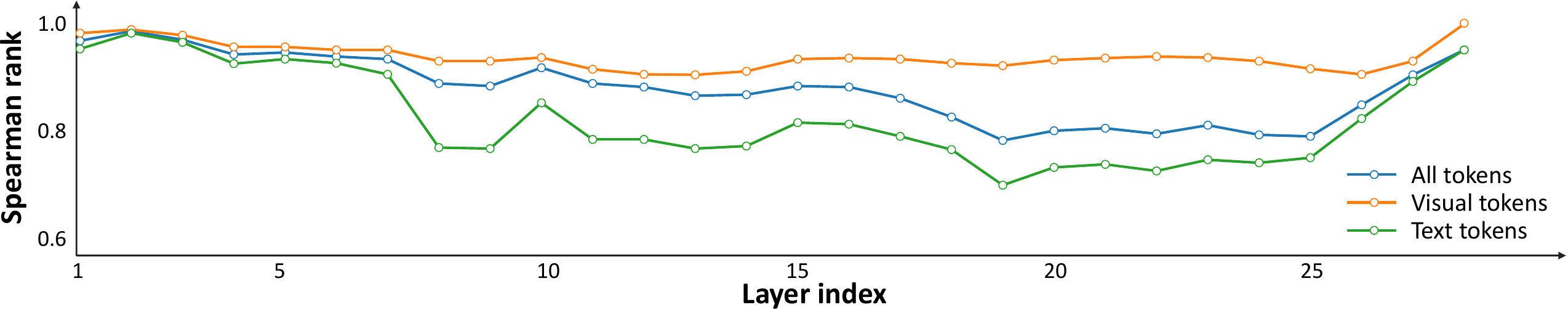}
    \caption{\small \textbf{Layer-wise agreement of FFN Taylor rankings across model states.} Spearman rank correlation compares channel rankings from the same initial checkpoint before and after applying the selected attention mask, with scores estimated from all tokens, visual tokens, or text tokens.}
    \label{fig:appendix_ffn_taylor_ranks_change}
\end{figure}

Figure~\ref{fig:appendix_ffn_taylor_ranks_change} compares FFN Taylor rankings from the same initial checkpoint before and after applying the selected attention mask. Visual token rankings remain relatively stable across layers, whereas text token rankings change more through much of the middle and later decoder. Rankings computed over all tokens show an intermediate pattern. These differences motivate estimating FFN importance in the model state in which the selected FFN masks will subsequently be applied.

In Table~\ref{tab:old-new-full-comparison}, \textit{Dense} is the post-trained LLaVA-OneVision-7B reference without attention or FFN masking, used to normalize retention metrics. \textit{Old} computes FFN Taylor rankings on the initial unpruned checkpoint, while \textit{New} uses the same weights with attention masking ratios of $(40\%,60\%,10\%)$ for $(\VtwoV,\TtwoV,\TtwoT)$. Both settings apply the attention and visual-side FFN masks to the initial checkpoint and perform one epoch of recovery, with text-side FFNs kept dense.

\begin{table}[htbp]
\centering
\caption{\small \textbf{Effect of recomputing FFN rankings under attention masking.} Dense is the unmasked post-trained reference used to normalize ATR, GRR, and OGR. Bold marks the better Old/New result at each masking ratio. Arrows indicate changes from Old to New in percentage points.}
\label{tab:old-new-full-comparison}
\setlength{\tabcolsep}{2pt}
\resizebox{\textwidth}{!}{
\begin{tabular}{cc|ccc|ccc|ccc|ccc|ccc}
\toprule
\multirow{2}{*}{\textbf{Ratio}} &
\multirow{2}{*}{\textbf{Setting}} &
\multicolumn{3}{c|}{\textbf{General}} &
\multicolumn{3}{c|}{\textbf{Reasoning}} &
\multicolumn{3}{c|}{\textbf{OCR}} &
\multicolumn{3}{c|}{\textbf{Grounding}} &
\multirow{2}{*}{\makecell{\textbf{ATR}\\\textbf{(\%)}}} &
\multirow{2}{*}{\makecell{\textbf{GRR}\\\textbf{(\%)}}} &
\multirow{2}{*}{\makecell{\textbf{OGR}\\\textbf{(\%)}}} \\
&
& \textbf{GQA} & \textbf{VQA$^{\text{v2}}$} & \textbf{VQA$^{\text{ok}}$}
& \textbf{SQA} & \textbf{AI2D} & \textbf{MMStar}
& \textbf{VQA$^{\text{T}}$} & \textbf{VQA$^{\text{D}}$} & \textbf{OCRB}
& \textbf{Ref} & \textbf{Ref+} & \textbf{RefG}
& & & \\
\midrule
0\% & Dense
& 63.1 & 82.2 & 59.9 & 96.0 & 81.9 & 61.4 & 75.3 & 87.4 & 63.5 & 77.9 & 70.5 & 76.9 & 100.0 & 100.0 & 100.0 \\
\midrule

\multirow{2}{*}{10\%}
& Old & 63.2 & 81.8 & \textbf{59.5} & 95.4 & 81.0 & 60.1 & 73.6 & 85.1 & \textbf{66.6} & \textbf{77.8} & \textbf{69.9} & \textbf{76.4} & 99.4 & 99.2 & \textbf{99.7} \\
& New & \textbf{64.4} & \textbf{83.6} & 56.9 & \textbf{96.9} & \textbf{82.3} & \textbf{63.6} & \textbf{74.8} & \textbf{85.3} & 65.7 & 77.1 & 69.8 & 76.2 & \textbf{100.1}~\textcolor{red!60!black}{\((\uparrow0.7)\)} & \textbf{100.6}~\textcolor{red!60!black}{\((\uparrow1.5)\)} & 99.6~\textcolor{green!60!black}{\((\downarrow0.1)\)} \\
\midrule

\multirow{2}{*}{20\%}
& Old & 63.0 & 81.6 & \textbf{58.2} & 94.1 & 79.8 & 59.1 & 72.4 & 84.0 & 64.9 & 76.7 & 69.0 & \textbf{75.5} & 98.1 & 98.0 & 98.2 \\
& New & \textbf{64.0} & \textbf{82.8} & 55.7 & \textbf{97.0} & \textbf{82.3} & \textbf{63.8} & \textbf{73.6} & \textbf{85.8} & \textbf{65.7} & \textbf{77.2} & \textbf{69.9} & 75.2 & \textbf{99.7}~\textcolor{red!60!black}{\((\uparrow1.6)\)} & \textbf{100.1}~\textcolor{red!60!black}{\((\uparrow2.1)\)} & \textbf{99.2}~\textcolor{red!60!black}{\((\uparrow1.1)\)} \\
\midrule

\multirow{2}{*}{30\%}
& Old & 62.9 & 80.9 & \textbf{56.4} & 89.9 & 78.8 & 58.0 & 70.6 & 82.8 & 64.5 & 76.0 & 68.4 & 75.0 & 96.5 & 96.1 & 97.0 \\
& New & \textbf{64.4} & \textbf{83.0} & 56.3 & \textbf{97.0} & \textbf{81.9} & \textbf{64.0} & \textbf{75.0} & \textbf{86.0} & \textbf{65.1} & \textbf{78.0} & \textbf{70.5} & \textbf{76.6} & \textbf{100.2}~\textcolor{red!60!black}{\((\uparrow3.7)\)} & \textbf{100.4}~\textcolor{red!60!black}{\((\uparrow4.3)\)} & \textbf{100.0}~\textcolor{red!60!black}{\((\uparrow3.0)\)} \\
\midrule

\multirow{2}{*}{40\%}
& Old & 62.2 & 80.3 & 52.7 & 83.9 & 76.1 & 56.3 & 69.3 & 81.4 & 64.5 & 74.3 & 66.9 & 73.4 & 94.1 & 92.7 & 95.4 \\
& New & \textbf{63.4} & \textbf{82.0} & \textbf{60.0} & \textbf{96.2} & \textbf{81.3} & \textbf{61.3} & \textbf{74.1} & \textbf{85.2} & \textbf{65.6} & \textbf{77.8} & \textbf{70.8} & \textbf{77.3} & \textbf{100.0}~\textcolor{red!60!black}{\((\uparrow5.9)\)} & \textbf{99.9}~\textcolor{red!60!black}{\((\uparrow7.2)\)} & \textbf{100.0}~\textcolor{red!60!black}{\((\uparrow4.6)\)} \\

\bottomrule
\end{tabular}
}
\end{table}

Recomputing FFN Taylor importance rankings yields larger aggregate gains as the visual-side masking ratio increases. At 40\% masking, pruning based on the recomputed rankings achieves 100.0\% ATR, compared with 94.1\% using the original rankings, with GRR and OGR gains of 7.2 and 4.6 percentage points, respectively. These results support recomputing FFN importance with the selected attention mask applied before choosing the modality-specific channel masks for recovery.

\subsection{Modality-Conditioned Redundancy across Task Categories}
\label{app:ffn_4tasks}

\begin{figure}[htbp]
    \centering
    \includegraphics[width=\linewidth]{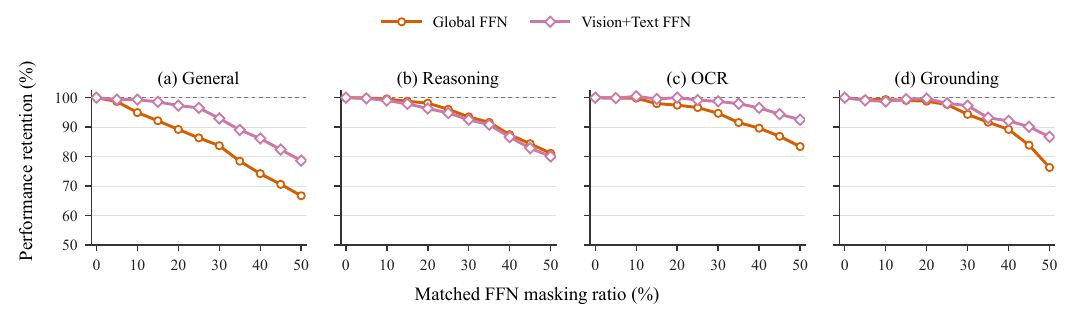}
    \caption{\small \textbf{Shared versus modality-specific FFN masks across four task categories.} All curves use Taylor-guided selection on LLaVA-OneVision-7B under the fixed $(40\%,60\%,10\%)$ attention-path mask and report training-free performance retention relative to the attention-only reference.}
    \label{fig:appendix_ffn_global_VS_joint_vision_text}
\end{figure}

\paragraph{Shared versus Modality-Specific Channel Masks.}
At masking ratio $r$, \textit{Global FFN} uses one low-importance channel set for both modalities, whereas \textit{Vision+Text FFN} selects a separate bottom-$r$ set for visual and text tokens. Both settings suppress the same fraction of channel executions for each modality, so their comparison isolates the effect of allowing modality-specific channel identities. Figure~\ref{fig:appendix_ffn_global_VS_joint_vision_text} shows that the advantage of Vision+Text grows with the masking ratio on General, OCR, and Grounding, while the curves remain close on Reasoning. Thus, modality conditioning improves the trade-off between performance and compression for several task categories.

\begin{figure}[b]
    \centering
    \includegraphics[width=\linewidth]{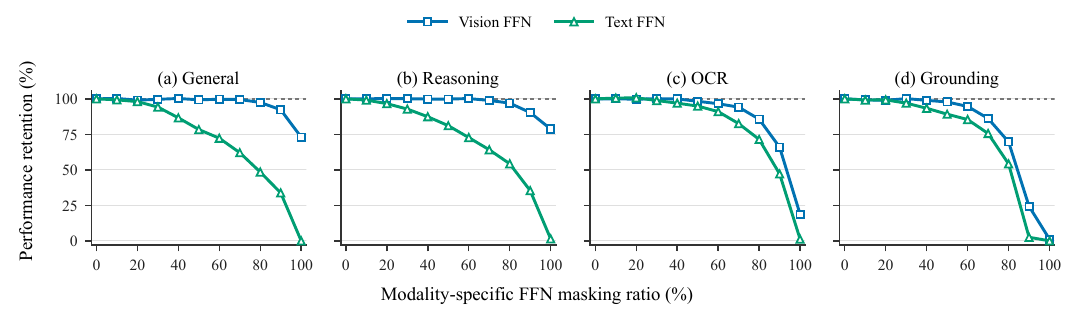}
    \caption{\small \textbf{Sensitivity to modality-specific FFN masking.} Vision FFN and Text FFN mask visual and text token channel executions respectively, while keeping the other modality dense. Masks are selected by Taylor importance on LLaVA-OneVision-7B with the fixed $(40\%,60\%,10\%)$ attention-path mask.}
    \label{fig:appendix_ffn_vision_text_ffn_}
\end{figure}

\paragraph{Visual-Side versus Text-Side Sensitivity.}
Figure~\ref{fig:appendix_ffn_vision_text_ffn_} shows that visual-side FFN masking preserves General and Reasoning performance over a broad range, whereas text-side masking causes an earlier decline. At moderate text-side masking ratios, General and Reasoning degrade more than OCR and Grounding. Under very aggressive visual-side masking, however, OCR and Grounding also deteriorate sharply. The relative sensitivity of the task categories varies with token modality and masking ratio. These results also support concentrating the FFN pruning budget on visual-side executions while conservatively retaining text-side computation.

\section{Additional Results}
\label{app:add_exp}
We present detailed pruning results and a complementary analysis of input-length effects. We first report per-benchmark attention-path and modality-conditioned FFN masking results on LLaVA-OneVision-7B (Section~\ref{app:detail_result_ov}), followed by corresponding results and comparisons with compression methods on Qwen2.5-VL-7B (Section~\ref{app:detail_result_qwen}). We then discuss how input sequence length affects computational costs and the interpretation of speedup (Section~\ref{app:sequence-length}). Together, these results characterize performance retention and computational efficiency across tasks, backbones, and input profiles.

For single-image evaluation, category retention averages the three benchmark score ratios to their post-trained dense references (in percent), and ATR, GRR, and OGR average category retentions over all four categories, General and Reasoning, and OCR and Grounding, respectively.

\subsection{Detailed Results on LLaVA-OneVision-7B}
\label{app:detail_result_ov}

\paragraph{Attention-Path Masking.}
Table~\ref{tab:path-mask-detailed} details the benchmark-level changes underlying Table~\ref{tab:path-mask-ratio}. Increasing \VtwoV{} masking from 30\% to 60\% lowers DocVQA from 87.1 to 83.6, while GQA and VQAv2 remain nearly unchanged. Similarly, with \VtwoV{} fixed at 40\%, increasing \TtwoV{} masking from 50\% to 80\% reduces DocVQA from 85.7 to 81.9 and TextVQA from 74.7 to 72.0. Document and scene-text understanding expose losses that can be obscured by stable general VQA scores.

The \TtwoT{} sweep exhibits a different pattern. Increasing its masking ratio from 10\% to 40\% lowers MMStar from 60.5 to 58.1, while the three grounding scores remain stable or improve. Thus, near-dense aggregate retention need not imply uniform preservation across benchmarks. Selection also matters within a fixed budget: at 40\% \VtwoV{} masking, Taylor exceeds random selection by 6.5 points on DocVQA and 6.9 points on RefCOCO+, despite identical masking ratios and recovery settings.

\begin{table}[htbp]
\centering
\caption{\small \textbf{Per-benchmark attention-path masking results on LLaVA-OneVision-7B after recovery.} Setting identifies the path varied at each stage, and the three mask columns specify the complete configuration.}
\label{tab:path-mask-detailed}
\setlength{\tabcolsep}{2pt}
\resizebox{\textwidth}{!}{
\begin{tabular}{c|ccc|ccc|ccc|ccc|ccc|ccc}
\toprule
\multirow{2}{*}{\textbf{Setting}} &
\multicolumn{3}{c|}{\textbf{Path Mask (\%)}} &
\multicolumn{3}{c|}{\textbf{General}} &
\multicolumn{3}{c|}{\textbf{Reasoning}} &
\multicolumn{3}{c|}{\textbf{OCR}} &
\multicolumn{3}{c|}{\textbf{Grounding}} &
\multirow{2}{*}{\makecell{\textbf{ATR}\\\textbf{(\%)}}} &
\multirow{2}{*}{\makecell{\textbf{GRR}\\\textbf{(\%)}}} &
\multirow{2}{*}{\makecell{\textbf{OGR}\\\textbf{(\%)}}} \\
& \textbf{\VtwoV} & \textbf{\TtwoV} & \textbf{\TtwoT}
& \textbf{GQA} & \textbf{VQA$^{\text{v2}}$} & \textbf{VQA$^{\text{ok}}$}
& \textbf{SQA} & \textbf{AI2D} & \textbf{MMStar}
& \textbf{VQA$^{\text{T}}$} & \textbf{VQA$^{\text{D}}$} & \textbf{OCRB}
& \textbf{Ref} & \textbf{Ref+} & \textbf{RefG}
& & & \\
\midrule
Dense Model
& 0 & 0 & 0
& 63.1 & 82.2 & 59.9 & 96.0 & 81.9 & 61.4 & 75.3 & 87.4 & 63.5 & 77.9 & 70.5 & 76.9 & 100.0 & 100.0 & 100.0 \\
\midrule
\multirow{5}{*}{\VtwoV}
& 30 & 0 & 0
& \underline{63.2} & \textbf{82.2} & 60.4 & \textbf{96.4} & \textbf{81.6} & 61.2 & \textbf{75.2} & \textbf{87.1} & \textbf{64.0} & \textbf{78.5} & \underline{70.2} & \textbf{77.0} & \textbf{100.1} & \underline{100.1} & \textbf{100.1} \\
& \graycell{40} & \graycell{0} & \graycell{0}
& \graycell{\underline{63.2}} & \graycell{\textbf{82.2}} & \graycell{\textbf{60.9}} & \graycell{\underline{96.2}} & \graycell{\textbf{81.6}} & \graycell{\textbf{61.7}} & \graycell{\underline{74.5}} & \graycell{\underline{86.7}} & \graycell{\underline{63.7}} & \graycell{\underline{78.0}} & \graycell{\textbf{70.4}} & \graycell{\underline{76.8}} & \graycell{\underline{100.0}} & \graycell{\textbf{100.3}} & \graycell{\underline{99.7}} \\
& $40^{\mathrm{R}}$ & 0 & 0
& 62.8 & 81.6 & 60.3 & 94.7 & 79.9 & 59.3 & 73.1 & 80.2 & 61.7 & 71.9 & 63.5 & 71.0 & 96.0 & 98.7 & 93.4 \\
& 50 & 0 & 0
& \underline{63.2} & \textbf{82.2} & \underline{60.8} & \underline{96.2} & \underline{81.5} & \underline{61.6} & \underline{74.5} & 85.6 & 63.4 & 77.6 & 70.0 & 76.7 & 99.7 & \textbf{100.3} & 99.2 \\
& 60 & 0 & 0
& \textbf{63.4} & \underline{82.1} & 60.1 & 95.9 & 81.4 & 60.9 & 74.1 & 83.6 & 61.8 & 77.0 & 69.1 & 75.8 & 98.8 & 99.8 & 97.8 \\
\midrule

\multirow{6}{*}{\TtwoV}
& 40 & 50 & 0
& \textbf{63.3} & \textbf{82.1} & \underline{60.0} & \textbf{96.6} & \underline{81.5} & 61.1 & \underline{74.7} & \textbf{85.7} & \textbf{66.1} & \textbf{78.5} & \underline{71.5} & \textbf{78.0} & \textbf{100.4} & \underline{100.0} & \textbf{100.8} \\
& \graycell{40} & \graycell{60} & \graycell{0}
& \graycell{\textbf{63.3}} & \graycell{\underline{82.0}} & \graycell{\textbf{60.3}} & \graycell{\underline{96.4}} & \graycell{\textbf{81.8}} & \graycell{\underline{61.3}} & \graycell{74.6} & \graycell{\underline{85.3}} & \graycell{\underline{65.5}} & \graycell{\underline{78.4}} & \graycell{\textbf{71.8}} & \graycell{\underline{77.8}} & \graycell{\underline{100.3}} & \graycell{\textbf{100.1}} & \graycell{\underline{100.6}} \\
& 40 & $60^{\mathrm{R}}$ & 0
& 62.4 & 80.8 & 57.6 & 91.5 & 78.4 & 54.9 & 70.4 & 81.3 & 64.5 & 73.1 & 65.2 & 71.0 & 95.0 & 95.6 & 94.5 \\
& $40^{\mathrm{R}}$ & $60^{\mathrm{R}}$ & 0
& 61.7 & 80.1 & 58.5 & 68.5 & 74.0 & \textbf{61.9} & \textbf{88.1} & 77.1 & 52.7 & 67.1 & 58.2 & 63.9 & 91.3 & 92.6 & 90.0 \\
& 40 & 70 & 0
& \underline{63.2} & 81.9 & 59.2 & \underline{96.4} & 81.4 & 60.9 & 73.6 & 84.7 & 65.0 & 77.5 & 70.4 & 76.9 & 99.5 & 99.6 & 99.4 \\
& 40 & 80 & 0
& 62.8 & 81.4 & 58.0 & 95.7 & \underline{81.5} & 59.9 & 72.0 & 81.9 & 65.3 & 77.3 & 69.7 & 76.6 & 98.5 & 98.7 & 98.3 \\
\midrule

\multirow{6}{*}{\TtwoT}
& \graycell{40} & \graycell{60} & \graycell{10}
& \graycell{\textbf{63.3}} & \graycell{\textbf{82.0}} & \graycell{\textbf{60.4}} & \graycell{\textbf{96.5}} & \graycell{\underline{81.5}} & \graycell{\underline{60.5}} & \graycell{74.0} & \graycell{85.3} & \graycell{\textbf{65.9}} & \graycell{78.1} & \graycell{71.0} & \graycell{77.0} & \graycell{\textbf{100.0}} & \graycell{\textbf{99.9}} & \graycell{100.1} \\
& 40 & 60 & $10^{\mathrm{R}}$
& \textbf{63.3} & \underline{81.9} & 58.8 & \underline{96.2} & 81.3 & 59.3 & \underline{74.5} & 85.4 & \underline{65.8} & \underline{78.2} & 70.5 & \underline{77.8} & 99.7 & 99.0 & \underline{100.3} \\
& $40^{\mathrm{R}}$ & $60^{\mathrm{R}}$ & $10^{\mathrm{R}}$
& 61.4 & 80.0 & 56.0 & 86.6 & 75.4 & 52.6 & 69.5 & 73.6 & 60.3 & 66.1 & 56.6 & 61.7 & 89.4 & 92.7 & 86.1 \\
& 40 & 60 & 20
& \underline{63.1} & \underline{81.9} & \underline{60.0} & 96.0 & \textbf{81.7} & \textbf{61.2} & \textbf{74.9} & \underline{85.5} & 65.5 & 77.4 & 70.8 & 76.3 & \underline{99.9} & \textbf{99.9} & 99.9 \\
& 40 & 60 & 30
& \underline{63.1} & \underline{81.9} & \underline{60.0} & 95.8 & 81.0 & 59.2 & 74.1 & \textbf{85.7} & 64.7 & \textbf{78.7} & \underline{71.4} & 77.7 & 99.7 & \underline{99.1} & 100.2 \\
& 40 & 60 & 40
& 63.0 & \underline{81.9} & 59.9 & 94.7 & 80.3 & 58.1 & 74.0 & \textbf{85.7} & 64.5 & \textbf{78.7} & \textbf{71.9} & \textbf{77.9} & 99.4 & 98.5 & \textbf{100.4} \\
\bottomrule
\end{tabular}
}
\end{table}

\paragraph{Modality-Conditioned FFN Masking.}
Table~\ref{tab:ffn-mask-detailed} decomposes the aggregate FFN results in Table~\ref{tab:ffn-progressive-search}. With text-side FFNs kept dense, raising visual-side masking from 60\% to 70\% reduces OK-VQA from 60.1 to 50.0, whereas GQA changes from 63.1 to 63.3 and VQAv2 from 81.9 to 81.7. The decline in General performance at this setting is therefore uneven across its constituent benchmarks, with a particularly sharp loss on the knowledge-oriented OK-VQA benchmark.

The text-side sweep reveals a complementary concentration of errors. With visual-side masking fixed at 40\%, increasing text-side masking from 0\% to 30\% lowers ScienceQA from 96.2 to 86.7 and OK-VQA from 60.0 to 54.3. Over the same sweep, RefCOCO+ changes only from 70.8 to 70.4 and RefCOCOg from 77.3 to 76.6. These differences motivate checking individual benchmarks alongside overall retention when choosing modality-specific FFN budgets.
\begin{table}[t]
\centering
\caption{\small \textbf{Per-benchmark FFN masking results on LLaVA-OneVision-7B after recovery.} Visual and Text specify modality-conditioned channel masking ratios under the fixed $(40\%,60\%,10\%)$ attention-path mask. MWOP\textsubscript{Attn} is the attention-only reference with dense FFNs.}
\label{tab:ffn-mask-detailed}
\setlength{\tabcolsep}{2pt}
\resizebox{\textwidth}{!}{
\begin{tabular}{c|cc|ccc|ccc|ccc|ccc|ccc}
\toprule
\multirow{2}{*}{\textbf{Setting}} &
\multicolumn{2}{c|}{\textbf{Channel Mask (\%)}} &
\multicolumn{3}{c|}{\textbf{General}} &
\multicolumn{3}{c|}{\textbf{Reasoning}} &
\multicolumn{3}{c|}{\textbf{OCR}} &
\multicolumn{3}{c|}{\textbf{Grounding}} &
\multirow{2}{*}{\makecell{\textbf{ATR}\\\textbf{(\%)}}} &
\multirow{2}{*}{\makecell{\textbf{GRR}\\\textbf{(\%)}}} &
\multirow{2}{*}{\makecell{\textbf{OGR}\\\textbf{(\%)}}} \\
& \textbf{Visual} & \textbf{Text}
& \textbf{GQA} & \textbf{VQA$^{\text{v2}}$} & \textbf{VQA$^{\text{ok}}$}
& \textbf{SQA} & \textbf{AI2D} & \textbf{MMStar}
& \textbf{VQA$^{\text{T}}$} & \textbf{VQA$^{\text{D}}$} & \textbf{OCRB}
& \textbf{Ref} & \textbf{Ref+} & \textbf{RefG}
& & & \\
\midrule

MWOP\textsubscript{Attn}
& 0 & 0
& 63.3 & 82.0 & 60.4 & 96.5 & 81.5 & 60.5 & 74.0 & 85.3 & 65.9 & 78.1 & 71.0 & 77.0 & 100.0 & 99.9 & 100.1 \\
\midrule

\multirow{6}{*}{\shortstack{Visual\\Neuron}}
& 40 & 0
& \textbf{63.4} & \textbf{82.0} & 60.0 & \textbf{96.2} & \underline{81.3} & \textbf{61.3} & \textbf{74.1} & \textbf{85.2} & \textbf{65.6} & \underline{77.8} & \textbf{70.8} & \textbf{77.3} & \textbf{100.0} & \textbf{99.9} & \textbf{100.0} \\
& $40^{\mathrm{R}}$ & 0
& 63.1 & 81.0 & 58.9 & 92.3 & 79.4 & 57.5 & 70.9 & 80.4 & 61.5 & 76.0 & 67.7 & 74.7 & 96.4 & 97.3 & 95.6 \\
& \graycell{50} & \graycell{0}
& \graycell{\textbf{63.4}} & \graycell{\textbf{82.0}} & \graycell{\textbf{60.5}} & \graycell{\textbf{96.2}} & \graycell{\underline{81.3}} & \graycell{60.5} & \graycell{\underline{73.8}} & \graycell{\underline{85.0}} & \graycell{\underline{64.5}} & \graycell{\textbf{77.9}} & \graycell{\underline{70.7}} & \graycell{\underline{77.1}} & \graycell{\underline{99.7}} & \graycell{\textbf{99.9}} & \graycell{\underline{99.6}} \\
& $50^{\mathrm{R}}$ & 0
& 62.9 & 80.5 & 58.6 & 91.2 & 78.7 & 56.2 & 70.3 & 77.4 & 61.2 & 74.4 & 66.0 & 72.5 & 94.9 & 96.3 & 93.6 \\
& 60 & 0
& 63.1 & \underline{81.9} & \underline{60.1} & \underline{96.0} & \textbf{81.4} & \underline{60.7} & 72.8 & 84.4 & \underline{64.5} & 77.5 & 70.2 & 76.8 & 99.3 & \underline{99.7} & 98.9 \\
& 70 & 0
& \underline{63.3} & 81.7 & 50.0 & 94.9 & 80.8 & 59.7 & 72.3 & 82.5 & 64.2 & 76.8 & 69.3 & 75.7 & 97.0 & 96.3 & 97.8 \\
\midrule

\multirow{5}{*}{\shortstack{Text\\Neuron}}
& \graycell{40} & \graycell{10}
& \graycell{\textbf{63.0}} & \graycell{\textbf{81.8}} & \graycell{\textbf{60.0}} & \graycell{\textbf{95.4}} & \graycell{\textbf{81.4}} & \graycell{\textbf{60.4}} & \graycell{\underline{73.5}} & \graycell{\textbf{85.4}} & \graycell{\underline{64.5}} & \graycell{\underline{77.9}} & \graycell{\textbf{71.1}} & \graycell{76.7} & \graycell{\textbf{99.5}} & \graycell{\textbf{99.4}} & \graycell{\textbf{99.6}} \\
& 40 & $10^{\mathrm{R}}$
& \underline{62.9} & \underline{81.4} & 56.4 & \underline{93.7} & 80.0 & 59.0 & \textbf{73.6} & \underline{85.1} & 62.0 & \textbf{78.1} & \underline{70.8} & \underline{76.8} & 98.1 & 97.3 & 98.9 \\
& $40^{\mathrm{R}}$ & $10^{\mathrm{R}}$
& 62.7 & 80.4 & 55.2 & 89.6 & 77.1 & 55.5 & 70.7 & 79.7 & 60.8 & 74.2 & 65.5 & 72.5 & 94.2 & 94.5 & 93.9 \\
& 40 & 20
& 62.7 & \underline{81.4} & \underline{57.8} & 90.8 & \underline{80.2} & \underline{59.6} & 73.0 & 85.0 & \textbf{65.0} & 77.5 & 70.7 & \textbf{76.9} & \underline{98.4} & \underline{97.4} & \underline{99.4} \\
& 40 & 30
& 62.4 & 80.7 & 54.3 & 86.7 & 77.9 & 59.5 & 73.1 & 84.7 & 64.0 & \textbf{78.1} & 70.4 & 76.6 & 97.0 & 95.0 & 99.1 \\
\bottomrule
\end{tabular}
 }
\end{table}

\subsection{Detailed Results on Qwen2.5-VL-7B}
\label{app:detail_result_qwen}
\paragraph{Attention-Path Masking.}
Table~\ref{tab:qwen-path-mask-detailed} repeats the path-ratio sweep on Qwen2.5-VL-7B. The $(60\%,70\%,0\%)$ configuration retains 99.8\% ATR, and adding 10\% \TtwoT{} masking gives 99.5\%. Larger \TtwoT{} ratios increasingly affect General and Reasoning: at 30\%, GRR decreases to 98.1\%, compared with 99.9\% when \TtwoT{} remains dense. Thus, this backbone also supports substantial compression of visually related paths while benefiting from a conservative text-side allocation.

\begin{table}[htbp]
\centering
\caption{\small \textbf{Per-benchmark attention-path masking results on Qwen2.5-VL-7B after recovery.} All path masks use Taylor importance. \VtwoV{} is swept first, followed by \TtwoV{} with \VtwoV{} fixed at 60\%, and then \TtwoT{} with the first two ratios fixed at 60\% and 70\%, respectively.}
\label{tab:qwen-path-mask-detailed}
\resizebox{\textwidth}{!}{
\begin{tabular}{c|ccc|ccc|ccc|ccc|ccc|ccc}
\toprule
\multirow{2}{*}{\textbf{Setting}} &
\multicolumn{3}{c|}{\textbf{Path Mask (\%)}} &
\multicolumn{3}{c|}{\textbf{General}} &
\multicolumn{3}{c|}{\textbf{Reasoning}} &
\multicolumn{3}{c|}{\textbf{OCR}} &
\multicolumn{3}{c|}{\textbf{Grounding}} &
\multirow{2}{*}{\makecell{\textbf{ATR}\\\textbf{(\%)}}} &
\multirow{2}{*}{\makecell{\textbf{GRR}\\\textbf{(\%)}}} &
\multirow{2}{*}{\makecell{\textbf{OGR}\\\textbf{(\%)}}} \\
& \textbf{\VtwoV} & \textbf{\TtwoV} & \textbf{\TtwoT}
& \textbf{GQA} & \textbf{VQA$^{\text{v2}}$} & \textbf{VQA$^{\text{ok}}$}
& \textbf{SQA} & \textbf{AI2D} & \textbf{MMStar}
& \textbf{VQA$^{\text{T}}$} & \textbf{VQA$^{\text{D}}$} & \textbf{OCRB}
& \textbf{Ref} & \textbf{Ref+} & \textbf{RefG}
& & & \\
\midrule

Dense Model
& 0 & 0 & 0
& 62.9 & 82.5 & 59.3 & 87.8 & 83.0 & 63.6 & 83.4 & 90.9 & 84.8 & 88.9 & 82.2 & 86.1 & 100.0 & 100.0 & 100.0 \\
\midrule

\multirow{4}{*}{\VtwoV}
& 40 & 0 & 0
& \underline{63.0} & \textbf{82.5} & 58.9 & \underline{87.8} & \textbf{83.1} & 63.0 & \textbf{83.7} & \textbf{91.1} & \textbf{85.0} & 88.7 & 82.0 & 85.9 & \underline{99.9} & \underline{99.8} & \textbf{100.0} \\
& 50 & 0 & 0
& \textbf{63.2} & \textbf{82.5} & 58.9 & 87.7 & \underline{82.9} & \textbf{63.6} & \underline{83.6} & \underline{90.7} & 84.7 & \textbf{89.1} & \underline{82.2} & \textbf{86.1} & \textbf{100.0} & \textbf{99.9} & \textbf{100.0} \\
& \graycell{60} & \graycell{0} & \graycell{0}
& \graycell{62.9} & \graycell{\textbf{82.5}} & \graycell{\textbf{59.3}} & \graycell{87.0} & \graycell{\underline{82.9}} & \graycell{\underline{63.5}} & \graycell{83.5} & \graycell{90.4} & \graycell{84.7} & \graycell{\underline{88.9}} & \graycell{\textbf{82.4}} & \graycell{\underline{86.0}} & \graycell{\underline{99.9}} & \graycell{\underline{99.8}} & \graycell{\underline{99.9}} \\
& 70 & 0 & 0
& 62.9 & \underline{82.3} & \underline{59.0} & \textbf{88.0} & 82.6 & 62.9 & 83.0 & 89.5 & \underline{84.9} & 88.4 & 81.5 & 85.7 & 99.5 & 99.7 & 99.4 \\
\midrule

\multirow{4}{*}{\TtwoV}
& 60 & 50 & 0
& \underline{63.2} & \textbf{82.3} & \underline{59.0} & 87.4 & 83.2 & \textbf{63.2} & \textbf{83.6} & \textbf{90.0} & 83.9 & 88.8 & \underline{82.0} & 85.7 & \underline{99.7} & \underline{99.8} & \underline{99.6} \\
& 60 & 60 & 0
& \underline{63.2} & \textbf{82.3} & \textbf{59.1} & \textbf{87.7} & \underline{83.4} & 62.5 & \underline{83.4} & \underline{89.9} & \underline{84.3} & \textbf{89.0} & \textbf{82.1} & \underline{85.8} & \underline{99.7} & \underline{99.8} & \textbf{99.7} \\
& \graycell{60} & \graycell{70} & \graycell{0}
& \graycell{\textbf{63.4}} & \graycell{\underline{82.2}} & \graycell{58.7} & \graycell{\underline{87.5}} & \graycell{\textbf{83.8}} & \graycell{\underline{63.0}} & \graycell{83.0} & \graycell{89.3} & \graycell{\textbf{84.9}} & \graycell{\underline{88.9}} & \graycell{\underline{82.0}} & \graycell{\textbf{86.1}} & \graycell{\textbf{99.8}} & \graycell{\textbf{99.9}} & \graycell{\underline{99.6}} \\
& 60 & 80 & 0
& \textbf{63.4} & 82.1 & 58.0 & 87.4 & 83.3 & 62.0 & 81.4 & 87.4 & \underline{84.3} & 88.3 & 81.1 & 85.4 & 98.8 & 99.2 & 98.4 \\
\midrule

\multirow{3}{*}{\TtwoT}
& \graycell{60} & \graycell{70} & \graycell{10}
& \graycell{\textbf{63.4}} & \graycell{\textbf{82.1}} & \graycell{\textbf{58.1}} & \graycell{\underline{87.9}} & \graycell{\textbf{83.7}} & \graycell{\textbf{62.7}} & \graycell{\textbf{82.7}} & \graycell{\textbf{89.0}} & \graycell{84.6} & \graycell{\textbf{88.7}} & \graycell{\textbf{82.0}} & \graycell{\underline{85.6}} & \graycell{\textbf{99.5}} & \graycell{\textbf{99.6}} & \graycell{\underline{99.3}} \\
& 60 & 70 & 20
& \underline{63.1} & \underline{82.0} & \underline{56.2} & \textbf{88.3} & \underline{83.5} & 61.4 & \textbf{82.7} & \underline{88.9} & \underline{84.8} & \underline{88.6} & \textbf{82.0} & \textbf{85.9} & \underline{99.0} & \underline{98.7} & \textbf{99.4} \\
& 60 & 70 & 30
& \underline{63.1} & \underline{82.0} & 54.9 & 86.4 & 82.8 & \underline{62.4} & \underline{82.5} & 88.6 & \textbf{85.7} & 88.1 & \underline{81.7} & 85.2 & 98.6 & 98.1 & 99.2 \\

\bottomrule
\end{tabular}
}
\end{table}

\paragraph{Modality-Conditioned FFN Masking.}
For the FFN experiments in Table~\ref{tab:qwen-ffn-mask-detailed}, we fix the attention mask at $(60\%,70\%,0\%)$. This attention mask differs from the highlighted 10\% \TtwoT{} row in the attention-only sweep. Masking 40\% of visual-side FFN executions while retaining all text-side executions gives 99.5\% ATR, 99.7\% GRR, and 99.2\% OGR. Adding 20\% text-side masking reduces ATR to 97.6\%, with a larger decline in GRR than OGR.

The adopted configurations therefore share the same modality-aware structure across backbones, but not the same ratios: Qwen2.5-VL-7B uses more \VtwoV{} and \TtwoV{} masking and less visual-side FFN masking than LLaVA-OneVision-7B. This supports selecting the operating point for each backbone rather than transferring a single numerical mask configuration unchanged.

\begin{table}[t]
\centering
\caption{\small \textbf{Per-benchmark FFN masking results on Qwen2.5-VL-7B after recovery.} Visual and Text give modality-conditioned channel masking ratios under the fixed $(60\%,70\%,0\%)$ attention-path mask.}
\label{tab:qwen-ffn-mask-detailed}
\setlength{\tabcolsep}{2pt}
\resizebox{\textwidth}{!}{
\begin{tabular}{c|cc|ccc|ccc|ccc|ccc|ccc}
\toprule
\multirow{2}{*}{\textbf{Setting}} &
\multicolumn{2}{c|}{\textbf{Channel Mask (\%)}} &
\multicolumn{3}{c|}{\textbf{General}} &
\multicolumn{3}{c|}{\textbf{Reasoning}} &
\multicolumn{3}{c|}{\textbf{OCR}} &
\multicolumn{3}{c|}{\textbf{Grounding}} &
\multirow{2}{*}{\makecell{\textbf{ATR}\\\textbf{(\%)}}} &
\multirow{2}{*}{\makecell{\textbf{GRR}\\\textbf{(\%)}}} &
\multirow{2}{*}{\makecell{\textbf{OGR}\\\textbf{(\%)}}} \\
& \textbf{Visual} & \textbf{Text}
& \textbf{GQA} & \textbf{VQA$^{\text{v2}}$} & \textbf{VQA$^{\text{ok}}$}
& \textbf{SQA} & \textbf{AI2D} & \textbf{MMStar}
& \textbf{VQA$^{\text{T}}$} & \textbf{VQA$^{\text{D}}$} & \textbf{OCRB}
& \textbf{Ref} & \textbf{Ref+} & \textbf{RefG}
& & & \\
\midrule
MWOP\textsubscript{Attn}
& 0 & 0
& 63.4 & 82.2 & 58.7 & 87.5 & 83.8 & 63.0 & 83.0 & 89.3 & 84.9 & 88.9 & 82.0 & 86.1 & 99.8 & 99.9 & 99.6 \\
\midrule

\multirow{4}{*}{\shortstack{Visual\\Neuron}}
& 30 & 0
& \textbf{63.6} & \textbf{82.3} & \textbf{58.7} & 87.9 & \textbf{83.5} & 62.8 & \textbf{82.6} & \textbf{89.1} & \underline{84.3} & 88.6 & 81.9 & \underline{86.0} & \textbf{99.6} & \textbf{99.9} & \textbf{99.3} \\
& \graycell{40} & \graycell{0}
& \graycell{\underline{63.4}} & \graycell{\textbf{82.3}} & \graycell{\underline{58.5}} & \graycell{87.5} & \graycell{83.0} & \graycell{\textbf{63.3}} & \graycell{\underline{81.9}} & \graycell{\underline{88.7}} & \graycell{\textbf{85.0}} & \graycell{88.6} & \graycell{81.8} & \graycell{85.9} & \graycell{\underline{99.5}} & \graycell{\underline{99.7}} & \graycell{\underline{99.2}} \\
& 50 & 0
& \underline{63.4} & \underline{82.1} & 57.7 & \textbf{88.3} & \underline{83.3} & \underline{62.9} & 81.0 & 88.5 & 83.4 & \textbf{89.0} & \textbf{82.2} & \textbf{86.1} & 99.2 & 99.6 & 98.8 \\
& 60 & 0
& 63.3 & 82.0 & 57.3 & \underline{88.0} & 82.8 & 62.5 & 80.0 & 88.4 & 83.9 & \underline{88.8} & \underline{82.1} & 85.9 & 98.9 & 99.2 & 98.6 \\
\midrule

\multirow{2}{*}{\shortstack{Text\\Neuron}}
& \graycell{40} & \graycell{10}
& \graycell{\textbf{63.2}} & \graycell{\textbf{81.8}} & \graycell{\textbf{55.4}} & \graycell{\textbf{88.0}} & \graycell{\textbf{83.2}} & \graycell{\textbf{62.4}} & \graycell{\textbf{81.8}} & \graycell{\textbf{88.6}} & \graycell{\textbf{85.0}} & \graycell{\textbf{88.7}} & \graycell{\textbf{81.8}} & \graycell{\textbf{85.6}} & \graycell{\textbf{98.9}} & \graycell{\textbf{98.6}} & \graycell{\textbf{99.1}} \\
& 40 & 20
& \textbf{63.2} & \underline{81.0} & \underline{50.3} & \underline{87.0} & \underline{82.8} & \textbf{62.4} & \underline{81.0} & \underline{88.5} & \underline{84.0} & \underline{88.2} & \underline{81.3} & \textbf{85.6} & \underline{97.6} & \underline{96.7} & \underline{98.5} \\

\bottomrule
\end{tabular}
}
\end{table}

\paragraph{Comparison with Compression Methods.}
Table~\ref{tab:qwen-main-comparison} reports per-benchmark comparisons on Qwen2.5-VL-7B. MWOP retains 99.5\% ATR with operation compression alone. Its token-compressed variants, MWOP\textsubscript{Z} and MWOP\textsubscript{P}, retain 96.2\% and 95.6\% ATR, respectively. Compared with the corresponding token-only methods with post-training recovery in this table, MWOP\textsubscript{Z} increases ATR from 94.9\% to 96.2\% and OGR from 92.4\% to 95.1\%. MWOP\textsubscript{P} increases ATR from 88.5\% to 95.6\% and OGR from 83.0\% to 93.8\%. These results extend the benefit of combining operation and token compression to the second backbone.

\begin{table}[htbp]
\centering
\caption{\small \textbf{Comparison of compression methods on Qwen2.5-VL-7B at approximately 40\% retained LLM-side FLOPs.} Standalone MWOP at 67.4\% retained FLOPs is included as a reference.}
\label{tab:qwen-main-comparison}
\setlength{\tabcolsep}{2pt}
\resizebox{\textwidth}{!}{
\begin{tabular}{c|ccc|ccc|ccc|ccc|ccc|ccc}
\toprule
\multirow{2}{*}{\textbf{Method}} &
\multicolumn{3}{c|}{\textbf{Sparsity}} &
\multicolumn{3}{c|}{\textbf{General}} &
\multicolumn{3}{c|}{\textbf{Reasoning}} &
\multicolumn{3}{c|}{\textbf{OCR}} &
\multicolumn{3}{c|}{\textbf{Grounding}} &
\multirow{2}{*}{\makecell{\textbf{ATR}\\\textbf{(\%)}}} &
\multirow{2}{*}{\makecell{\textbf{GRR}\\\textbf{(\%)}}} &
\multirow{2}{*}{\makecell{\textbf{OGR}\\\textbf{(\%)}}} \\
& \textbf{Dep.} & \textbf{Wid.} & \textbf{Tok.}
& \textbf{GQA} & \textbf{VQA$^{\text{v2}}$} & \textbf{VQA$^{\text{ok}}$}
& \textbf{SQA} & \textbf{AI2D} & \textbf{MMStar}
& \textbf{VQA$^{\text{T}}$} & \textbf{VQA$^{\text{D}}$} & \textbf{OCRB}
& \textbf{Ref} & \textbf{Ref+} & \textbf{RefG}
& & & \\
\midrule
Qwen2.5-VL-7B
& - & - & -
& 62.9 & 82.5 & 59.3 & 87.8 & 83.0 & 63.6 & 83.4 & 90.9 & 84.8 & 88.9 & 82.2 & 86.1 & 100.0 & 100.0 & 100.0 \\
\midrule
PyramidDrop\textsuperscript{$\ddagger$}
& & & \cmark
& 53.9 & 76.4 & 35.3 & 82.4 & 72.1 & 44.4 & 78.7 & 65.4 & 23.5 & 31.7 & 26.3 & 27.2 & 65.1 & 81.4 & 48.9 \\
ZOO-Prune
& & & \cmark
& 59.7 & 79.6 & 34.3 & 84.9 & 76.8 & 55.8 & 57.2 & 64.0 & 57.0 & 81.3 & 73.7 & 76.4 & 83.5 & 87.7 & 79.4 \\
ShortV
& \cmark & &
& 54.0 & 71.1 & 31.6 & 81.5 & 68.1 & 46.7 & 26.1 & 57.7 & 47.1 & 55.1 & 47.9 & 48.7 & 66.7 & 78.9 & 54.5 \\
RedundancyLens
& & \cmark &
& 57.0 & 78.1 & 34.4 & 83.2 & 75.9 & 55.8 & 72.8 & 80.2 & 69.5 & 78.0 & 70.1 & 72.2 & 86.0 & 86.2 & 85.7 \\
\rowcolor{red!8}
PyramidDrop
& & & \cmark
& 60.2 & 79.4 & \underline{57.2} & 84.1 & 77.1 & 55.6 & 80.6 & 81.1 & 67.8 & 70.4 & 61.4 & 67.1 & 88.5 & 94.1 & 83.0 \\
\rowcolor{red!8}
ZOO-Prune\textsuperscript{$\dagger$}
& & & \cmark
& 62.7 & 80.8 & 56.8 & 85.5 & 81.0 & \underline{61.0} & 75.7 & 81.1 & 71.7 & 86.2 & 79.0 & 83.3 & 94.9 & \underline{97.4} & 92.4 \\
\rowcolor{red!8}
ShortV\textsuperscript{$\dagger$}
& \cmark & &
& 62.2 & 80.3 & 54.9 & 82.2 & 78.3 & 55.8 & 77.4 & 80.7 & 75.4 & 84.2 & 77.1 & 81.0 & 93.1 & 94.1 & 92.2 \\
\rowcolor{red!8}
RedundancyLens\textsuperscript{$\dagger$}
& & \cmark &
& 62.4 & 81.2 & 56.4 & 84.0 & 80.5 & 58.4 & 78.9 & \underline{86.5} & \underline{81.0} & 86.8 & 79.8 & 83.6 & \underline{96.2} & 96.2 & \underline{96.2} \\
\rowcolor{lightblue}
MWOP (ours)
& & \cmark &
& \textbf{63.4} & \textbf{82.3} & \textbf{58.5} & \underline{87.5} & \textbf{83.0} & \textbf{63.3} & \textbf{81.9} & \textbf{88.7} & \textbf{85.0} & \textbf{88.6} & \textbf{81.8} & \textbf{85.9} & \textbf{99.5} & \textbf{99.7} & \textbf{99.2} \\
\rowcolor{lightblue}
MWOP\textsubscript{Z} (ours)
& & \cmark & \cmark
& \underline{62.9} & \underline{81.3} & 56.3 & 85.5 & \underline{81.7} & 60.1 & 77.4 & 83.4 & 77.8 & \underline{87.3} & \underline{80.3} & \underline{84.4} & \underline{96.2} & 97.3 & 95.1 \\
\rowcolor{lightblue}
MWOP\textsubscript{P} (ours)
& & \cmark & \cmark
& 62.5 & \underline{81.3} & 56.6 & \textbf{88.0} & 80.5 & 59.6 & \underline{81.2} & 85.7 & 75.6 & 84.0 & 76.1 & 81.9 & 95.6 & \underline{97.4} & 93.8 \\
\bottomrule
\end{tabular}
}
\end{table}

\begin{figure}[htbp]
    \centering
    \includegraphics[width=0.9\linewidth]{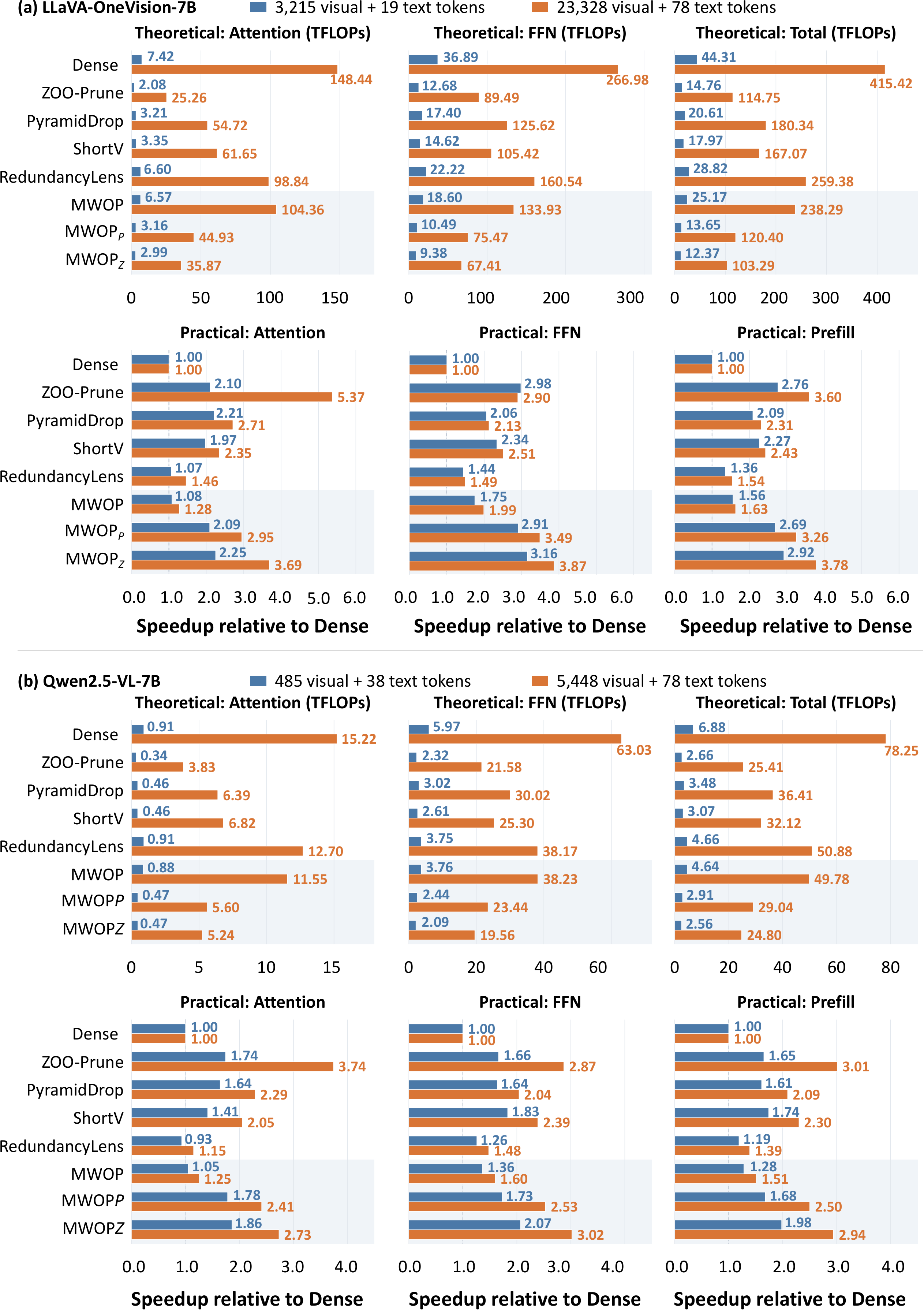}
    \caption{\textbf{Input-length dependence of computational cost and acceleration.} Blue and orange bars represent single-image and video inputs. Input lengths are averaged over 12 single-image benchmarks and three video benchmarks (MVBench, Video-MME, and NExT-QA). Within each model block, the \textbf{upper row} reports theoretical attention, FFN, and total decoder FLOPs in TFLOPs, while the \textbf{lower row} reports measured attention, FFN, and prefill speedups relative to Dense. Lower FLOPs and higher speedups indicate greater efficiency under the corresponding input-length setting.}
    \label{fig:appendix_img_video_speedup}
\end{figure}

\subsection{Effect of Input Sequence Length}
\label{app:sequence-length}

We evaluate decoder prefill on an NVIDIA A100-SXM4-40GB GPU with batch size 1 and BF16 precision for both the single-image and video input profiles. Figures~\ref{fig:appendix_img_video_speedup} compares computational cost and measured prefill acceleration at two representative input profiles for LLaVA-OneVision-7B and Qwen2.5-VL-7B, respectively. The blue bars use the mean visual and text token counts from the 12 single-image benchmarks. The orange bars use the corresponding means from MVBench, Video-MME, and NExT-QA. These profiles represent shorter image inputs and longer video inputs. System tokens are excluded from the FLOPs inputs for both single-image profiles and the Qwen2.5-VL video profile (Section~\ref{app:flops}). Latency measurements include system tokens following Section~\ref{app:efficient-inference}. Each method and its dense reference use matching input profiles.

Longer visual sequences increase the attention share of the computation. For dense LLaVA-OneVision-7B, attention accounts for approximately 16.7\% of total FLOPs at the image profile and 35.7\% at the video profile. The corresponding shares for Qwen2.5-VL-7B are 13.2\% and 19.5\%. This shift is consistent with the quadratic growth of causal visual self-attention, compared with the linear token dependence of projections and FFNs (Section~\ref{app:flops}). FFNs nevertheless remain the larger component at both profiles, leaving room for complementary savings by reducing both the number of tokens and the operations performed on each retained token.

The measured results show larger prefill speedups at the video profile. MWOP\textsubscript{Z} increases prefill speedup from $2.92\times$ to $3.78\times$ on LLaVA-OneVision-7B and from $1.98\times$ to $2.94\times$ on Qwen2.5-VL-7B. The gains for MWOP alone are more modest, increasing from $1.56\times$ to $1.63\times$ and from $1.28\times$ to $1.51\times$ on LLaVA-OneVision-7B and Qwen2.5-VL-7B, respectively.

\section{Implementation Details}
\label{app:implement_details}
We describe the implementation and experimental settings of MWOP. We first introduce the evaluated LMMs (Section~\ref{app:lmm}), specify the adopted operation and token pruning configurations (Section~\ref{app:method_config}), and present the sequential search and recovery procedure (Section~\ref{app:algorithm_impl}). We then detail the FLOPs calculation (Section~\ref{app:flops}) and the prefill implementation and latency evaluation protocol (Section~\ref{app:efficient-inference}). Finally, we summarize the post-training data composition (Section~\ref{app:post-train-data}) and training configuration (Section~\ref{app:train_config}). Together, these details specify the setup for pruning, recovery, and efficiency evaluation across the two backbones and their combinations with token pruning.

\subsection{Details of LMMs}
\label{app:lmm}

\textbf{LLaVA-OneVision-7B} connects a SigLIP vision encoder to Qwen2-7B-Instruct through a two-layer MLP projector that maps visual features into the language embedding space. Its AnyRes strategy represents high-resolution images using a global view and local crops.

\textbf{Qwen2.5-VL-7B} integrates a vision transformer supporting native dynamic resolution with Qwen2.5-7B. Its vision encoder uses window attention, while an MLP merger combines each spatial group of $2\times2$ patch features into a single visual token aligned with the language embedding space.

Both models process visual and textual tokens within an autoregressive language decoder. Table~\ref{tab:model_config} summarizes their language-decoder configurations. The attention and FFN masks remain fixed throughout post-training and are applied during multimodal prefill at inference.

\begin{table}[htbp]
\centering
\caption{\small Language-model backbone configurations of LLaVA-OneVision-7B and Qwen2.5-VL-7B.}
\begin{tabular}{lcc}
\toprule
Model & LLaVA-OneVision-7B & Qwen2.5-VL-7B \\
\midrule
LLM Backbone & Qwen2-7B-Instruct & Qwen2.5-7B \\
Blocks       & 28                & 28          \\
Query Heads  & 28                & 28          \\
KV Heads     & 4                 & 4           \\
Head Dim     & 128               & 128         \\
Hidden Dim   & 3584              & 3584        \\
FFN Dim      & 18944             & 18944       \\
\bottomrule
\end{tabular}
\label{tab:model_config}
\end{table}

\subsection{Configuration Details of Our Method}
\label{app:method_config}

\begin{figure}[htbp]
    \centering
    \includegraphics[width=\linewidth]{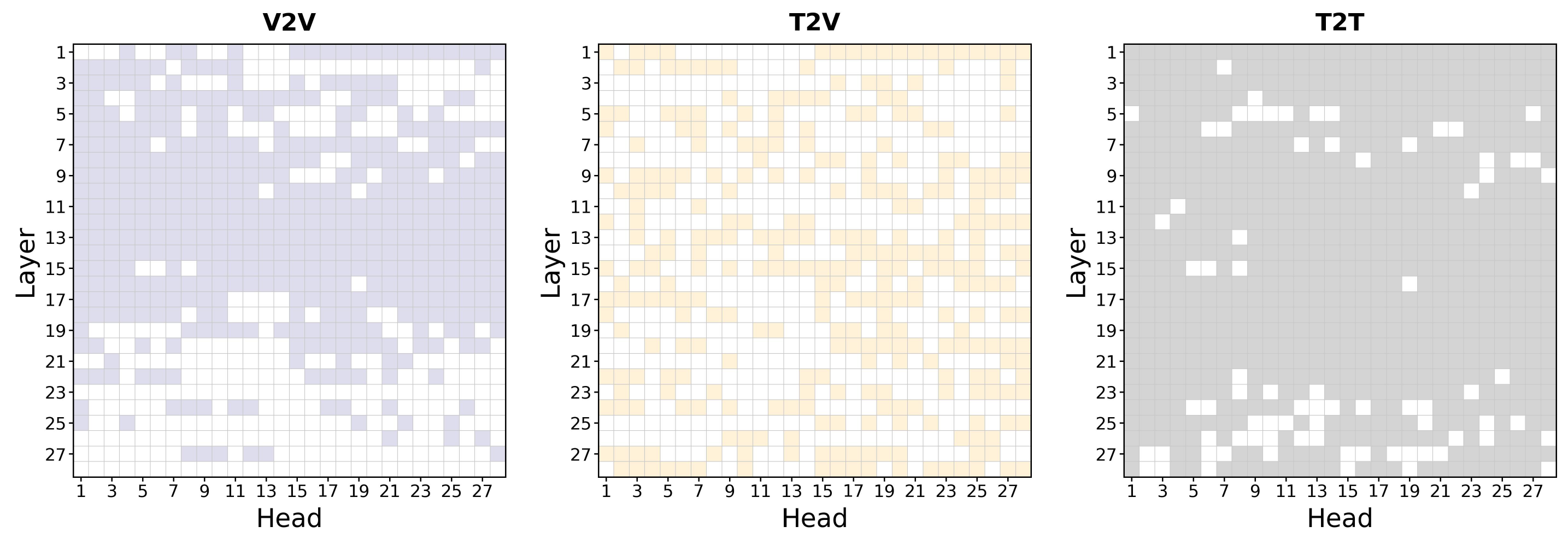}
    \caption{\small \textbf{Layer-head attention-path masks for LLaVA-OneVision-7B.} Rows and columns denote decoder layers and query heads, respectively, using one-based labels. Colored cells are retained, while white cells are masked. The masks remove 314/784 \VtwoV{} paths (40.1\%), 470/784 \TtwoV{} paths (59.9\%), and 78/784 \TtwoT{} paths (9.9\%). Each path is selected independently within a head.}
    \label{fig:attn_path_masks}
\end{figure}

\paragraph{Attention-Path Pruning.}
Figure~\ref{fig:attn_path_masks} visualizes the layer-head masks for V2V, T2V, and T2T attention paths, with target pruning ratios of 40\%, 60\%, and 10\%, respectively. Each mask specifies the retained modality-specific interactions across 28 decoder layers and 28 query heads. Pruning is applied to individual attention paths rather than entire attention heads.

\paragraph{Modality-Conditioned FFN Pruning.}
Table~\ref{tab:mwop-ffn-layerwise} reports the layer-wise FFN pruning configuration. We prune 50\% of the visual token intermediate channels globally across all decoder layers while retaining all text token channels. Under the L27-first configuration, all visual token FFN channels in the final decoder layer (layer 27, zero-indexed) are masked first. The remaining pruning budget is allocated across the other layers using the global Taylor ranking. The resulting allocation is nonuniform across layers, with substantially heavier pruning in the early and late layers than in the middle layers. Qwen2.5-VL-7B instead uses attention-path ratios of $(60\%,70\%,0\%)$ and FFN masking ratios of $(40\%,0\%)$ for visual and text tokens, respectively. The layer-wise table below describes the LLaVA-OneVision configuration, with separate entries for visual and text tokens.

\begin{table}[htbp]
\centering
\caption{\small \textbf{Layer-wise FFN masking for the adopted LLaVA-OneVision-7B configuration.} Layers are indexed from 0 to 27. Counts and percentages refer to masked modality-conditioned channel executions out of 18,944 channels per layer. The allocation masks 50\% of visual-side channels globally across layers while retaining all text-side channels for the entire input sequence in every decoder layer.}
\label{tab:mwop-ffn-layerwise}
\begin{tabular}{c cc | c cc}
\toprule
\multirow{2}{*}{\textbf{Layer}} & \multicolumn{2}{c|}{\textbf{Pruned FFN Channels}} &
\multirow{2}{*}{\textbf{Layer}} & \multicolumn{2}{c}{\textbf{Pruned FFN Channels}} \\
\cmidrule(lr){2-3}\cmidrule(lr){5-6}
& \textbf{Vision} & \textbf{Text} & & \textbf{Vision} & \textbf{Text} \\
\midrule
0 & 16,473 (87.0\%) & 0 (0.0\%) & 14 & 980 (5.2\%) & 0 (0.0\%) \\
1 & 14,690 (77.5\%) & 0 (0.0\%) & 15 & 2,533 (13.4\%) & 0 (0.0\%) \\
2 & 14,722 (77.7\%) & 0 (0.0\%) & 16 & 7,231 (38.2\%) & 0 (0.0\%) \\
3 & 12,337 (65.1\%) & 0 (0.0\%) & 17 & 9,414 (49.7\%) & 0 (0.0\%) \\
4 & 10,083 (53.2\%) & 0 (0.0\%) & 18 & 12,317 (65.0\%) & 0 (0.0\%) \\
5 & 9,088 (48.0\%) & 0 (0.0\%) & 19 & 15,706 (82.9\%) & 0 (0.0\%) \\
6 & 126 (0.7\%) & 0 (0.0\%) & 20 & 15,423 (81.4\%) & 0 (0.0\%) \\
7 & 9 (0.0\%) & 0 (0.0\%) & 21 & 16,450 (86.8\%) & 0 (0.0\%) \\
8 & 18 (0.1\%) & 0 (0.0\%) & 22 & 16,345 (86.3\%) & 0 (0.0\%) \\
9 & 637 (3.4\%) & 0 (0.0\%) & 23 & 17,375 (91.7\%) & 0 (0.0\%) \\
10 & 2 (0.0\%) & 0 (0.0\%) & 24 & 18,010 (95.1\%) & 0 (0.0\%) \\
11 & 1 (0.0\%) & 0 (0.0\%) & 25 & 17,873 (94.3\%) & 0 (0.0\%) \\
12 & 11 (0.1\%) & 0 (0.0\%) & 26 & 18,346 (96.8\%) & 0 (0.0\%) \\
13 & 72 (0.4\%) & 0 (0.0\%) & 27 & 18,944 (100.0\%) & 0 (0.0\%) \\
\bottomrule
\end{tabular}
\end{table}

\paragraph{Combination with Token Pruning.}
For MWOP combined with PDrop, pruning occurs at entry to layers 7, 14, and 21 (zero-based), after 7, 14, and 21 blocks have been executed. Cumulative visual token retention ratios are approximately 54.4\%, 29.6\%, and 16.1\%, respectively, measured relative to the original visual token count. Both MWOP\textsubscript{P} and MWOP\textsubscript{Z} retain approximately 50\% of the original visual tokens on average across all decoder layers. In both combinations, the original position IDs of the retained tokens are preserved throughout inference.

\subsection{Algorithm Implementation}
\label{app:algorithm_impl}
\begin{algorithm}[htbp]
\caption{Sequential Redundancy Search with Post-Training Recovery}
\label{alg:mwop_algorithm}
\small
\begin{algorithmic}[1]

\Statex \textbf{Input:}
Dense model $\mathcal{M}_0$,
calibration data $\mathcal{D}_{\mathrm{cal}}$,
recovery data $\mathcal{D}_{\mathrm{rec}}$,
selection set $\mathcal{D}_{\mathrm{sel}}$ (1,000 samples),
full benchmark suite $\mathcal{B}_{\mathrm{full}}$,
ratio grid $\mathcal{R}=\{0,0.1,\ldots,1.0\}$.
\Statex \textbf{Order:}
$\mathcal{U}=[\mathrm{V2V},\mathrm{T2V},\mathrm{T2T},
\mathrm{FFN}_{V},\mathrm{FFN}_{T}]$.

\State $\Omega\gets\varnothing,\quad \mathcal{H}\gets[]$
\State $\mathcal{I}\gets
\Call{TaylorImportance}{\mathcal{M}_0,\mathcal{D}_{\mathrm{cal}},\mathrm{Attn}}$

\For{$u\in\mathcal{U}$}
    \If{$u=\mathrm{FFN}_{V}$}
        \State $\mathcal{I}\gets
        \Call{TaylorImportance}
        {\operatorname{Mask}(\mathcal M_0,\Omega),\mathcal{D}_{\mathrm{cal}},\mathrm{FFN}}$
        \Comment{Both modalities, once}
    \EndIf

    \For{$\rho\in\mathcal{R}$}
        \State $\Omega_{\rho}\gets
        \Call{AddMask}{\Omega,u,\mathcal{I}_u,\rho}$
        \State $A_{\mathrm{TF}}(\rho)\gets
        \Call{EvaluateATR}
        {\operatorname{Mask}(\mathcal M_0,\Omega_{\rho}),
        \mathcal{D}_{\mathrm{sel}}}$
    \EndFor

    \State $\mathcal{C}\gets
    \Call{DropCandidates}{\mathcal{R},A_{\mathrm{TF}}}$
    \While{$\mathcal{C}$ has untested candidates}
        \State Select an untested $\rho\in\mathcal{C}$
        \State $\mathcal{M}^{\mathrm{REC}}_{\rho}\gets
        \Call{PostTrain}
        {\operatorname{Mask}(\mathcal M_0,\Omega_{\rho}),
        \mathcal{D}_{\mathrm{rec}}}$
        \Comment{One epoch, fixed masks}
        \State $A_{\mathrm{full}}(\mathcal M^{\mathrm{REC}}_{\rho})\gets
        \Call{EvaluateATR}
        {\mathcal{M}^{\mathrm{REC}}_{\rho},\mathcal{B}_{\mathrm{full}}}$
        \State Expand $\mathcal C\subseteq\mathcal R$ using this stage's performance changes
    \EndWhile

    \State Append this stage's recovered candidates to $\mathcal H$ in search order
    \State $\rho^{\mathrm{anc}}\gets
    \arg\min_{\rho\in\mathcal{C}}|A_{\mathrm{full}}(\mathcal M^{\mathrm{REC}}_{\rho})-100|$
    \State $\Omega\gets\Omega_{\rho^{\mathrm{anc}}},\quad \mathcal M_{\mathrm{anc}}\gets\mathcal M^{\mathrm{REC}}_{\rho^{\mathrm{anc}}}$
\EndFor

\State $\mathcal M^*\gets\mathcal M_{\mathrm{anc}}$ \Comment{Final-stage anchor}
\For{$\mathcal M\in\mathcal H$ in search order}
    \If{$A_{\mathrm{full}}(\mathcal M)>A_{\mathrm{full}}(\mathcal M^*)$ and $F(\mathcal M)<F(\mathcal M^*)$}
        \State $\mathcal M^*\gets\mathcal M$
    \EndIf
\EndFor
\Statex \textbf{Output:}
Compressed model $\mathcal{M}^{*}$ and
$\Call{GetAllMasks}{\mathcal{M}^{*}}$.

\end{algorithmic}
\end{algorithm}

Algorithm~\ref{alg:mwop_algorithm} searches V2V, T2V, and T2T attention paths, followed by visual and text FFN channels. Attention importance is estimated once on the initial model $\mathcal M_0$. FFN importance for both modalities is estimated once on $\mathcal M_0$ with the attention masks selected through the T2T stage applied. At each stage, a training-free ratio sweep evaluates masked models on the selection set $\mathcal D_{\mathrm{sel}}$, which contains 1,000 samples, to identify candidates for post-training recovery. The candidate set is expanded according to the observed performance changes. Every candidate is evaluated with its cumulative masks applied to $\mathcal M_0$ and independently recovered for one epoch on $\mathcal D_{\mathrm{rec}}$. The masks remain fixed throughout recovery, using the settings in Section~\ref{app:train_config}.

Each recovered candidate is evaluated on all samples in the full benchmark suite $\mathcal B_{\mathrm{full}}$. The candidate whose ATR is closest to 100\% provides the anchor mask for the next stage. Stages pass the selected masks forward, while all candidate runs start from the same initial weights $\mathcal M_0$. We store all recovered candidates in $\mathcal H$ in search order and complete every stage through $\mathrm{FFN}_T$. The recorded full-benchmark ATR scores are reused for final selection. Starting from the final-stage anchor, we revisit candidates in search order and replace the current selection only when a candidate has both higher full-benchmark ATR and lower theoretical FLOPs. Here, $F(\mathcal M)$ denotes the theoretical decoder FLOPs of candidate $\mathcal M$, computed using the counting convention in Section~\ref{app:flops}.

\subsection{FLOPs Calculation Details}
\label{app:flops}
\paragraph{Input Profiles.}
For theoretical FLOPs, we use representative input profiles with $V_0$ visual tokens and $T$ text tokens, excluding system tokens. The single-image profiles are $(V_0,T)=(3215,19)$ for LLaVA-OneVision-7B and $(485,38)$ for Qwen2.5-VL-7B. The Qwen2.5-VL video profile uses $(V_0,T)=(5448,78)$ under the same FLOPs counting convention.

\paragraph{Counting Convention.}
We count two FLOPs per multiply-accumulate and report matrix-multiplication costs in TFLOPs ($10^{12}$ FLOPs). At decoder layer $\ell$, the counted sequence length is $N_\ell=V_\ell+T$, where $V_\ell$ denotes the remaining visual tokens. System tokens are excluded from both projection costs and attention-pair counts. MWOP preserves token counts, whereas token pruning reduces $V_\ell$ at the layer boundaries specified by each token-pruning method.

\paragraph{Attention Computation.}
Both decoders have $L=28$ layers, $H=28$ query heads, $H_{\mathrm{KV}}=4$ key/value heads, head dimension $d_h=128$, hidden width $d=3584$, and key/value width $d_{\mathrm{KV}}=512$. MWOP retains all Q/K/V/O projection dimensions and the native grouped-query attention mapping. Within the counted input sequence, let $\mathcal A_{\ell,h}(i)$ denote the causal keys retained for query $i$ in head $h$, including unpruned interactions with any non-system text preceding the visual span. The mean retained pair count per head and the attention FLOPs at layer $\ell$ are
\begin{equation*}
a_\ell=\frac{1}{H}\sum_{h=1}^{H}\sum_{i=1}^{N_\ell}
\bigl|\mathcal A_{\ell,h}(i)\bigr|,
\qquad
F_{\mathrm{attn},\ell}=4N_\ell d(d+d_{\mathrm{KV}})+4d a_\ell.
\end{equation*}
The first term accounts for the four projections. The second accounts for QK products and attention-value multiplication, each costing $2d a_\ell$. Dense causal attention gives $a_\ell=N_\ell(N_\ell+1)/2$, including the diagonal. For MWOP, the pair count follows the actual layer-head masks, which need not distribute the global pruning ratio uniformly across layers.

\paragraph{FFN Computation and Total FLOPs.}
Let $m=18944$ be the full intermediate width and $k_{V,\ell}$ the retained visual channel count. With alignment $b$, the executed visual width and FFN cost are
\begin{equation*}
\widetilde m_{V,\ell}
=\min\!\left(m,b\left\lceil\frac{k_{V,\ell}}{b}\right\rceil\right),
\qquad
F_{\mathrm{FFN},\ell}=6d\bigl(V_\ell\widetilde m_{V,\ell}+Tm\bigr).
\end{equation*}
The factor six counts the gate, up, and down projections. Text tokens retain the full FFN width in the evaluated configurations. The standalone MWOP implementations and the Qwen2.5-VL combinations use $b=64$, while the LLaVA-OneVision combinations use $b=1$. Zero retained channels therefore incur no visual FFN projection cost. The decoder total includes all attention and FFN projections, retained attention interactions, and the last-token language-model head:
\begin{equation*}
F_{\mathrm{total}}=
\sum_{\ell=0}^{L-1}\left(F_{\mathrm{attn},\ell}+F_{\mathrm{FFN},\ell}\right)+2dU,
\end{equation*}
where $U$ is the vocabulary size, 152128 for LLaVA-OneVision and 152064 for Qwen2.5-VL. The attention and FFN columns exclude the language-model head. The single-image profiles above yield dense totals of 44.31 and 6.88 TFLOPs for LLaVA-OneVision and Qwen2.5-VL, respectively.

These counts include FFN alignment padding but exclude normalization, RoPE, softmax, elementwise operations, tensor movement, and excess work in boundary tiles. Comparison methods are counted using their actual token lengths, projection shapes, attention patterns, and FFN widths. In particular, ShortV layers that bypass visual queries still compute the visual K/V projections required by text queries. Online token scoring and ranking are excluded from both FLOPs and latency comparisons under the decoder-only evaluation protocol. This choice is conservative for MWOP, whose pruning masks are static and incur no online selection overhead at inference, while some token-pruning baselines require additional runtime scoring or ranking.

\subsection{Efficient Inference Implementation}
\label{app:efficient-inference}

MWOP reduces decoder-prefill computation through path-sparse attention and compact visual FFNs.

\paragraph{Path-Sparse Attention.}
For the retained causal key set $\mathcal A_{\ell,h}(i)$ in the actual input sequence, including system tokens, the attention output for query $i$ is computed as
\begin{equation*}
o_i=\frac{\sum_{j\in\mathcal A_{\ell,h}(i)}\exp(s_{ij})v_j}
{\sum_{j\in\mathcal A_{\ell,h}(i)}\exp(s_{ij})},
\qquad s_{ij}=q_i^\top k_j/\sqrt{d_h}.
\end{equation*}
The Triton kernel traverses only the retained key intervals and maintains a single online softmax state across them. This implements pre-softmax path masking while skipping pruned QK and attention-value products, without materializing the attention matrix. Causal and interval-end masks handle boundary tiles. Operation pruning preserves GQA mapping, projection widths, and KV-cache dimensions.

\paragraph{Modality-Conditioned FFNs.}
The retained rows of the gate and up projections, along with the corresponding columns of the down projection, are packed into compact visual weights. Visual tokens use these weights, while text tokens use the full FFN. Empty visual channel sets yield a zero FFN update. The packed weights require additional storage alongside the full text weights. To reduce intermediate data movement, the optimized Qwen implementation fuses normalization and modality splitting, gated activation, and output merging with residual addition.

\paragraph{Latency Evaluation.}
We measure batch-one BF16 prefill latency on an NVIDIA A100-SXM4-40GB using model weights and synthetic embedding-table inputs. Timing inputs include system tokens. For single-image measurements, these add 18 tokens for LLaVA-OneVision-7B and 11 for Qwen2.5-VL-7B to the profiles in Section~\ref{app:flops}. Dense and pruned models use matching input lengths within each timing setting. The dense baseline uses FlashAttention-2. Qwen2.5-VL uses its native multimodal rotary operator with identical linear temporal and spatial coordinates.
We complete compilation, weight packing, kernel tuning, and graph capture before timing, with kernel configurations fixed for each measured shape. After warmup, CUDA events measure CUDA Graph replay latency. The reported latency is the median of five window means, each computed from ten replay batches. Attention and FFN latencies are obtained by summing separately measured layer-module latencies, while total prefill latency is measured independently. Speedup is the ratio of dense to pruned total prefill latency for each input setting on the same hardware.

\subsection{Post-training data composition}
\label{app:post-train-data}

For post-training recovery, we use a mixture of 282,293 samples developed as part of a separate work. This specific training mixture has not yet been publicly released. Table~\ref{tab:posttrain_data} summarizes its source datasets and sample counts, covering general visual understanding, grounding, reasoning, OCR, and language-only instruction following. The recovery training data do not overlap with any of the evaluation sets used in this paper for single-image, multi-image, or video understanding.

\begin{table}[htbp]
\centering
\caption{Composition of the post-training data mixture.}
\label{tab:posttrain_data}
\begin{tabular}{lrc}
\toprule
\textbf{Dataset} & \textbf{Count} & \textbf{Category} \\
\midrule
COCO~\citep{lin2014microsoft}           & 103{,}909 & General \\
VG~\citep{krishna2017visual}             &  55{,}504 & Grounding \\
GQA~\citep{hudson2019gqa}            &  42{,}777 & Reasoning \\
OCRVQA~\citep{mishra2019ocr}         &  27{,}346 & General + OCR \\
ChartQA~\citep{masry2022chartqa}        &  10{,}000 & Reasoning + OCR \\
DVQA~\citep{kafle2018dvqa}           &  10{,}000 & Reasoning + OCR \\
DocVQA~\citep{mathew2021docvqa}         &  10{,}000 & OCR \\
AI2D~\citep{kembhavi2016diagram}           &   7{,}508 & Reasoning \\
LLaVAR~\citep{zhang2023llavar}         &   6{,}785 & OCR \\
TQA~\citep{kembhavi2017you}            &   6{,}501 & Reasoning \\
ShareGPT~\citep{chiang2023vicuna}       &   1{,}000 & Language-only \\
TextVQA~\citep{singh2019towards}        &      963 & General + OCR \\
\midrule
\textbf{Total} & \textbf{282{,}293} & \textbf{N/A} \\
\bottomrule
\end{tabular}
\end{table}

\subsection{Training Configuration}
\label{app:train_config}
For each LLaVA-OneVision-7B candidate, we apply its cumulative pruning masks to $\mathcal M_0$ and independently finetune for one epoch on the mixture in Table~\ref{tab:posttrain_data}. The selected configuration is given in Section~\ref{app:method_config}. Only the multimodal projector and language-model LoRA adapters are updated. The vision encoder and base language-model weights remain frozen. LoRA uses rank 128, $\alpha=256$, and dropout 0.05. We use AdamW with a learning rate of $1 \times 10^{-5}$ for both trainable components, zero weight decay, cosine scheduling, and a warmup ratio of 0.03. Training runs on eight A100s, with a per-device batch size of 1 and 32 gradient accumulation steps. The effective batch size is 256. Training uses BF16, gradient checkpointing, SDPA, and DeepSpeed ZeRO-2.

\section{Details of Comparison Methods}
\label{app:methods_details}

\paragraph{FastV~\citep{chen2024image}.}
FastV prunes visual tokens after shallow language-model layers using text-to-vision attention scores. It discards low-scoring tokens from subsequent layers, reducing sequence length and the associated attention and FFN computation.

\paragraph{PyramidDrop~\citep{xing2024pyramiddrop}.}
PyramidDrop progressively prunes visual tokens at multiple depths of the language model. Instruction-guided attention determines which tokens to retain, allocating larger token budgets to shallow layers and smaller budgets to deeper layers.

\paragraph{VisionZip~\citep{yang2025visionzip}.}
VisionZip compresses visual representations before language-model processing by retaining dominant tokens and constructing contextual tokens. It combines attention-based selection with similarity-based aggregation to preserve salient visual content and complementary contextual information across the image.

\paragraph{DivPrune~\citep{alvar2025divprune}.}
DivPrune formulates visual token selection as a Max-Min Diversity Problem. It greedily maximizes the minimum pairwise distance among retained tokens, reducing redundant representations while maintaining broad coverage of diverse visual content.

\paragraph{DART~\citep{wen2025stop}.}
DART performs duplication-aware visual token pruning using visual and textual pivot tokens. It retains tokens with low similarity to these pivots, emphasizing complementary visual information rather than relying solely on attention-based importance scores.

\paragraph{VisPruner~\citep{zhang2025beyond}.}
VisPruner combines visual importance and diversity for token pruning without text-guided attention. It first selects important tokens using visual attention, then supplements them with diverse tokens through similarity-based duplicate removal.

\paragraph{FEATHER~\citep{endo2025feather}.}
FEATHER combines token selection based on RoPE-free attention scores with uniform spatial sampling. Its two-stage pruning strategy preserves spatial coverage during early selection and further reduces visual tokens in later layers.

\paragraph{MMTok~\citep{dong2026mmtok}.}
MMTok formulates visual token selection as multimodal coverage maximization. It jointly considers text-visual relevance and visual-visual coverage, greedily selecting tokens that represent the image while preserving information relevant to the input query.

\paragraph{PruMerge+~\citep{shang2025llava}.}
PruMerge+ combines attention-guided token selection, similarity-based merging, and spatial sampling. Selected tokens aggregate similar visual features, while additional spatially sampled tokens preserve broader image coverage beyond regions highlighted by attention scores in the initial token selection step.

\paragraph{ZOO-Prune~\citep{kim2026zoo}.}
ZOO-Prune estimates visual token sensitivity at the projector stage using zeroth-order perturbations. It combines sensitivity with diversity to retain informative tokens before language-model processing, without requiring gradient backpropagation.

\paragraph{ShortV~\citep{yuan2025shortv}.}
ShortV identifies low-contribution layers through output divergence and freezes visual hidden states in those layers. It skips visual attention updates and FFN computation while preserving visual keys and values for text queries.

\paragraph{RedundancyLens~\citep{li2025redundancylens}.}
RedundancyLens combines Hollow Attention with Probe-Activated Dynamic FFN to reduce visual computation. It restricts visual self-attention to local neighborhoods and uses sampled visual tokens to select active FFN channels in designated layers.

\paragraph{HalfV~\citep{shi2026inheritance}.}
HalfV distinguishes intrinsic visual redundancy from architecture-dependent saturation. It combines token pruning with backbone-aware reductions in later visual processing, adapting compression to how visual representations evolve across the language model.

\paragraph{DOP$_{\mathrm{V}}$~\citep{liu2025fine}.}
DOP$_{\mathrm{V}}$ combines VisPruner-based visual token reduction with depth-wise operation pruning across attention and FFNs. Under a computational budget, it uses output divergence to select different visual processing depths for attention and FFNs.

\paragraph{YOPO~\citep{zhang2024treat}.}
YOPO reduces visual computation through head pruning, local visual attention, late-layer operation removal, and FFN channel pruning. These reductions establish a computation pattern shared across inputs rather than selected dynamically for each sample.

\paragraph{Skip-Vision~\citep{zeng2025skip}.}
Skip-Vision allows selected visual tokens to bypass FFN transformations while remaining available to attention. It combines token aggregation with learned summarization and removes skipped-token cache entries after prefilling to reduce decoding overhead.

\section{Details of Benchmarks}
\label{app:benchmarks}

\subsection{Single-image benchmarks}
\label{app:single-img_benchmarks}

\paragraph{GQA~\citep{hudson2019gqa}.}
GQA evaluates compositional visual reasoning using scene-graph annotations from Visual Genome. It requires models to combine information about object attributes and relationships to answer questions grounded in the visual scene.

\paragraph{VQA-v2~\citep{goyal2017making}.}
VQA-v2 is a balanced, open-ended visual question answering benchmark designed to reduce language priors. It encourages models to examine visual evidence and produce image-grounded answers rather than rely primarily on question patterns.

\paragraph{OK-VQA~\citep{marino2019ok}.}
OK-VQA evaluates knowledge-based visual question answering, where image content is often insufficient. Models must connect visual observations with external or commonsense knowledge to answer questions that extend beyond directly observable information.

\paragraph{ScienceQA~\citep{lu2022learn}.}
ScienceQA evaluates multimodal scientific reasoning across natural, language, and social sciences. Its image-based subset requires models to combine visual evidence with question context and relevant scientific knowledge to identify the correct answer.

\paragraph{AI2D~\citep{kembhavi2016diagram}.}
AI2D evaluates diagram understanding through questions involving labels, symbols, visual structures, and spatial relations. Models must connect these elements and interpret their relationships to reason about the information conveyed by a diagram.

\paragraph{MMStar~\citep{chen2024we}.}
MMStar evaluates multimodal understanding with curated questions that require visual evidence. Its samples span multiple visual and reasoning capabilities and are selected to reduce language-only shortcuts and potential data leakage. We include it in the Reasoning category when reporting category-level performance retention in our experiments.

\paragraph{TextVQA~\citep{singh2019towards}.}
TextVQA evaluates scene-text understanding by requiring models to recognize and reason over text in natural images. Answering its questions involves integrating textual content with the surrounding visual context rather than performing text recognition alone.

\paragraph{DocVQA~\citep{mathew2021docvqa}.}
DocVQA evaluates question answering over document images, emphasizing OCR, layout understanding, and information localization. Models must locate relevant document regions and interpret their textual content in the context of the question.

\paragraph{OCRBench~\citep{liu2024ocrbench}.}
OCRBench evaluates OCR-related capabilities of MLLMs across text recognition, document understanding, information extraction, and related tasks. It assesses the accuracy of text perception and the ability to use textual information for downstream understanding.

\paragraph{RefCOCO~\citep{kazemzadeh2014referitgame,yu2016modeling}.}
RefCOCO evaluates referring expression comprehension by requiring models to localize objects described in natural language. Models must match object attributes and relationships to the expression to distinguish the target.

\paragraph{RefCOCO+~\citep{kazemzadeh2014referitgame,yu2016modeling}.}
RefCOCO+ disallows location words during annotation, emphasizing appearance-based referring expressions. Models must identify the target from visual attributes such as color, shape, and clothing.

\paragraph{RefCOCOg~\citep{mao2016generation,yu2016modeling}.}
RefCOCOg contains longer and more descriptive referring expressions, testing fine-grained and context-rich visual grounding. Models must integrate multiple cues, including object attributes and relationships, to identify the intended target.

\subsection{Multi-image benchmarks}
\label{app:multi-img_benchmarks}

\paragraph{Q-Bench2~\citep{zhang2024q}.}
Q-Bench2 evaluates low-level visual perception through questions about image pairs. Models must jointly inspect and compare visual attributes, such as clarity, lighting, and distortions, to identify subtle differences and similarities between images.

\paragraph{Mantis-Eval~\citep{jiang2024mantis}.}
Mantis-Eval evaluates multi-image reasoning and co-reference through manually curated multiple-choice and short-answer questions. Models must compare visual content and combine evidence across images to answer questions requiring joint interpretation.

\paragraph{BLINK~\citep{fu2024blink}.}
BLINK evaluates core visual perception through tasks such as relative depth estimation, visual correspondence, and multi-view reasoning. It requires models to interpret fine-grained visual and spatial cues from single or multiple images.

\subsection{Video benchmarks}
\label{app:vid_benchmarks}

\paragraph{MVBench~\citep{li2024mvbench}.}
MVBench evaluates temporal video understanding through 20 tasks spanning perception and reasoning. Models must combine evidence across frames to recognize actions, track state changes, and understand event sequences rather than rely on a single frame.

\paragraph{Video-MME~\citep{fu2025video}.}
Video-MME evaluates video understanding across diverse domains and durations, from short clips to hour-long videos. Tasks assess visual perception, temporal reasoning, contextual understanding, and information integration across time.

\paragraph{NExT-QA~\citep{xiao2021next}.}
NExT-QA evaluates video question answering with an emphasis on causal and temporal action reasoning. Models must understand event order, object interactions, and the reasons behind actions, moving beyond recognition and description of visible content.

\end{document}